\documentclass[10pt,twocolumn,letterpaper]{article}

\usepackage[pagenumbers]{iccv}      % To produce the REVIEW version
\usepackage{graphicx}
\usepackage{amsthm}
\usepackage[most]{tcolorbox}
\usepackage[dvipsnames]{xcolor}
\usepackage{amssymb}
\usepackage{amsmath}
\usepackage{booktabs}
\usepackage{multirow}
\usepackage{textcomp}
\usepackage{float}
\usepackage{relsize}
\usepackage{wrapfig,lipsum}
\usepackage{enumitem}
\usepackage[group-separator={,}]{siunitx}
\usepackage{dsfont}
\usepackage{tocloft}

\newcommand{\listappendixname}{Content List}
\newlistof{appendices}{app}{\listappendixname}

\usepackage{amsmath,amsfonts,bm}

\def\eqref#1{equation~\ref{#1}}
\def\1{\bm{1}}

\def\vs{{\bm{s}}}

\DeclareMathAlphabet{\mathsfit}{\encodingdefault}{\sfdefault}{m}{sl}
\SetMathAlphabet{\mathsfit}{bold}{\encodingdefault}{\sfdefault}{bx}{n}

\usepackage{url}
\usepackage{subcaption}

\theoremstyle{definition}
\newtheorem{definition}{Definition}

\newtcolorbox{promptbox}[1][]{%
  enhanced,
  colback=gray!10,       % Background color
  colframe=gray!70,      % Frame color
  coltitle=black,        % Title color (black by default)
  coltext=black,         % Text color (black by default)
  sharp corners,         % No round corners
  boxrule=0.5mm,         % Line thickness
  top=10pt,              % Top margin
  bottom=10pt,           % Bottom margin
  left=10pt,             % Left margin
  right=10pt,            % Right margin
  before skip=10pt,      % Space before
  after skip=10pt,       % Space after
  breakable=true,
  overlay={%
    \ifcase\tcbsegmentstate
    \or%
    \else%
    \fi%
}
}

\renewcommand{\vs}{\text{vs.\,}}

\definecolor{iccvblue}{rgb}{0.21,0.49,0.74}
\usepackage[pagebackref,breaklinks,colorlinks,allcolors=iccvblue]{hyperref}

\def\paperID{9046} % *** Enter the Paper ID here
\def\confName{ICCV}
\def\confYear{2025}

\title{VACT: A Video Automatic Causal Testing System and a Benchmark}

\author{
Haotong Yang$^{1,2,*}$,
Qingyuan Zheng$^{4,*}$,
Yunjian Gao$^{4,*}$,
Yongkun Yang$^{5}$,\\
Yangbo He$^{4,6,\dag}$,
Zhouchen Lin$^{1,2,3,\dag}$,
Muhan Zhang$^{2,3,\dag}$\\[0.2cm]
$^1$School of Intelligence Science and Technology, Peking University\\
$^2$Institution for Artificial Intelligence, Peking University\\
$^3$State Key Lab of General AI, Peking University\hspace{10pt}
$^4$School of Mathematical Science, Peking University\\
$^5$Yuanpei College, Peking University\hspace{10pt}
$^6$Center for Statistical Science, Peking University\\[0.1cm]
$^*$Equal contribution, \hspace{10pt}$^\dag$Corresponding authors\\[0.1cm]
\tt \normalsize \{haotongyang, qyzheng, zlin, muhan\}@pku.edu.cn, \\
\tt \normalsize heyb@math.pku.edu.cn,
\tt \normalsize \{gyj20010915, yang0316\}@stu.pku.edu.cn
}

\begin{document}
\maketitle
\begin{abstract}
With the rapid advancement of text-conditioned Video Generation Models (VGMs), the quality of generated videos has significantly improved, bringing these models closer to functioning as ``\textit{world simulators}'' and making real-world-level video generation more accessible and cost-effective. However, the generated videos often contain factual inaccuracies and lack understanding of fundamental physical laws. While some previous studies have highlighted this issue in limited domains through manual analysis, a comprehensive solution has not yet been established, primarily due to the absence of a generalized, automated approach for modeling and assessing the causal reasoning of these models across diverse scenarios.
To address this gap, we propose VACT: an \textit{automated} framework for modeling, evaluating, and measuring the causal understanding of VGMs in real-world scenarios. By combining causal analysis techniques with a carefully designed large language model assistant, our system can assess the causal behavior of models in various contexts without human annotation, which offers strong generalization and scalability. Additionally, we introduce multi-level causal evaluation metrics to provide a detailed analysis of the causal performance of VGMs.
As a demonstration, we use our framework to benchmark several prevailing VGMs, offering insight into their causal reasoning capabilities. Our work lays the foundation for systematically addressing the causal understanding deficiencies in VGMs and contributes to advancing their reliability and real-world applicability.
\end{abstract}    
\section{Introduction}
\label{sec:intro}
% \begin{figure*}[ht]
%     \centering
%     \begin{subfigure}{1\linewidth}
%     \centering
%     \includegraphics[width=0.8\linewidth]{figs/stone_pool_openai.jpg}
%     \includegraphics[width=0.8\linewidth]{figs/feather_pool_openai.jpg}
%     \caption{OpenAI Sora Generation}
%     \end{subfigure}
    
%     \begin{subfigure}{1\linewidth}
%     \centering
%     \includegraphics[width=0.8\linewidth]{figs/stone_pool_cogvideox2.jpg}
%     \includegraphics[width=0.8\linewidth]{figs/feather_pool_cogvideox2.jpg}
%     \caption{CogVideoX-2 Generation}
%     \end{subfigure}

%     \begin{subfigure}{1\linewidth}
%     \centering
%     \includegraphics[width=0.8\linewidth]{figs/stone_pool_runway.jpg}
%     \includegraphics[width=0.8\linewidth]{figs/feather_pool_runway.jpg}
%     \caption{Gen-3 Alpha Generation}
%     \end{subfigure}
    
%     \caption{Videos generated by (a) OpenAI Sora and (b) CogVideoX-2 (c) Gen-3 Alpha,shown as frames. For each model, the text prompt of the \textbf{Above} is: \textit{a stone is thrown into a swimming pool}; \textbf{Below} is: \textit{a feather is thrown into a swimming pool}. Both generation show \textit{noticeable splashes}, which is correct for the above (stone) scene but \textbf{incorrect} for the \textbf{below} (feather) scene.}
%     \label{fig:bad_case}
% \end{figure*}

% A shorter version
\begin{figure*}[ht]
    \centering
    \includegraphics[width=1\linewidth]{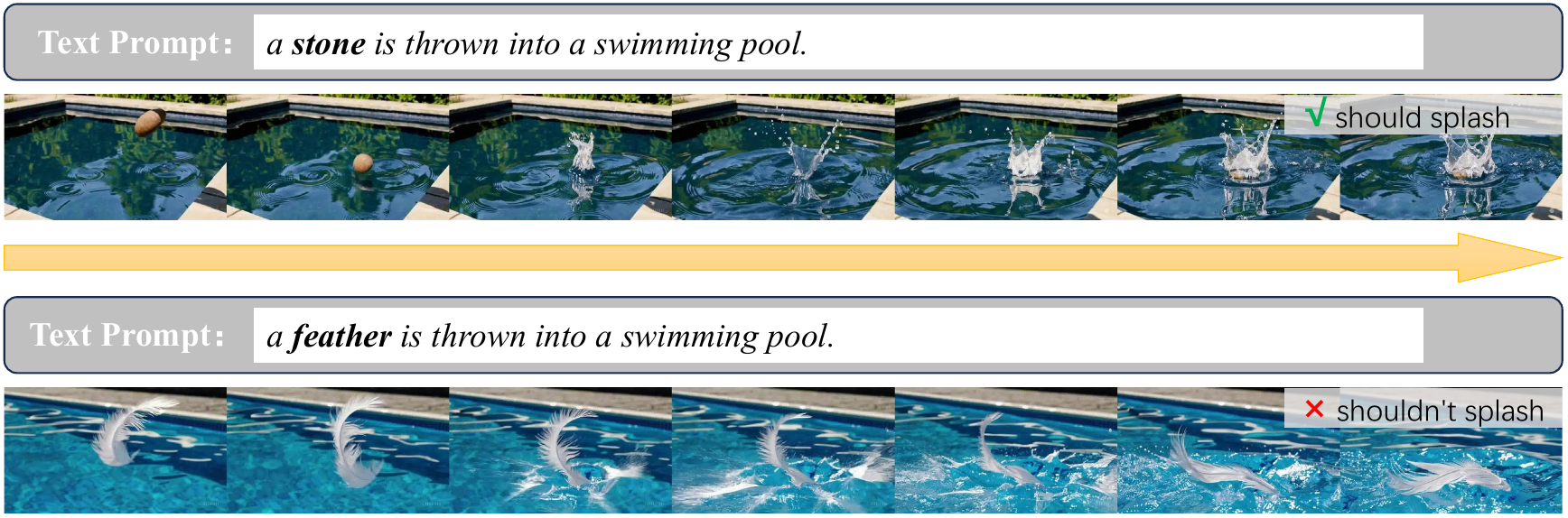}
    \caption{Videos generated by OpenAI Sora, shown as frames. The text prompt of the \textbf{Above} is: \textit{a stone is thrown into a swimming pool}; \textbf{Below} is: \textit{a feather is thrown into a swimming pool}. Both the generations show \textit{noticeable splashes}, which is correct for the above (stone) scene but \textbf{incorrect} for the \textbf{below} (feather) scene.}
    \label{fig:bad_case}
\end{figure*}
With the rapid development of Video Generation Models (VGMs), generated videos are becoming increasingly indistinguishable from real recordings. VGMs, particularly text-to-video (T2V) models\footnote{In this paper, the term VGM is referred specifically to T2V models. Text-conditioned generation is the most versatile and user-friendly method in world simulation, whereas image-conditioned models can enable T2V generation by combining with a text-to-image model.}, are expected to serve as ``world models'' or ``world simulators'', allowing users to generate scenes from text descriptions of real-world events or environments. This approach is cheaper, faster, and more scalable than arranging and recording real-world scenes and is expected to benefit fields like robotics, autonomous driving, and video understanding.

However, the ``\textit{hallucination}'' problem hinders the progress, which refers to a generation that seems correct but contains factual errors or fabrications. While VGMs have made significant strides in video quality---such as clarity, dynamic range, and continuity---they still struggle with issues like cause-and-effect confusion, detail errors, and incorrect object relationships, making the videos appear misleading upon closer inspection.
% However, VGMs also face the ``\textit{hallucination}'' problem%, similar to that in language models, which hinders the progress. Hallucination refers to the generation of content that seems correct but contains factual errors or fabrications. While VGMs have made significant strides in video quality -- such as clarity, dynamic range, and continuity -- they still struggle with issues like cause-and-effect confusion, detail errors, and incorrect object relationships, making the videos appear misleading upon closer inspection.

In Figure~\ref{fig:bad_case}, OpenAI Sora~\citep{sora_is_here} is 
% some VGMs are 
required to generate videos for two scenarios: ``\textit{a stone is thrown into a swimming pool}'' and ``\textit{a feather is thrown into a swimming pool}''. In both cases, an obvious splash and ripples occur around the object. While in the stone scenario the splash is accurate, the feather scenario fails to follow the correct physics principles, as the feather is too light to create a noticeable splash or ripples in reality. Here, the model seems to learn a \textbf{spurious correlation} (or ``\textbf{shortcut}'') between ``\textit{object hitting water}'' and ``\textit{splash}'', without understanding the actual causal factors, such as \textit{mass} and \textit{velocity}, making it difficult for the VGMs to generalize to less typical situations or being utilized as ``world simulator''. We provide similar results with other VGMs and more analysis in Appendix~\ref{app:more_examples}.

Although some work has acknowledged the hallucination problem in VGMs and proposed preliminary benchmarks to identify commonsense violations~\citep{bansal2024videophyevaluatingphysicalcommonsense,meng2024worldsimulatorcraftingphysical}, most of them rely on manual design of physical rules and test cases and focus on limited fields. However, real-world causal relationships are highly complex, with different scenarios involving different physical laws. Furthermore, even a simple scenario can involve various causal relationships. For instance, in the case of ``\textit{two objects collision}'', dynamics might focus on mass, velocity, and elasticity to determine the object motion after collision, while material properties like hardness or brittleness might determine whether the objects deform or break. More complex relationships, like sparks from a flint or splashes from wet objects, further highlight this complexity,  making it difficult to systematically address the hallucination problem through manual labeling.

% Although some work has recognized the hallucination problem and proposed some preliminary benchmarks to remind researchers of the components that violate commonsense in VGM generation, most of these benchmarks rely on manual design and are limited to a few individual fields, making it difficult to form a systematic solution. In fact, there is an essential difficulty, that is \textit{the complexity of causal relationships and rules} in the real world. Different scenarios may involve completely different laws, and even in a simple scenario, the ``possible'' causal relationships involved can be very diverse. For example, when we consider ``\textit{the collision of two objects}'', from a dynamics perspective, we may be concerned about the mass and velocity (i.e. momentum) and elasticity of the objects, which corresponds to the motion state of the objects after the collision in the video; from a materials perspective, we may be concerned about the hardness, brittleness, and so on of the object's material to determine whether the object has deformed or broken during the collision. If we take into account more complex situations, such as ``arcs and sparks may be generated when flints collide'' and ``liquid may splash when wet objects collide'', we will realize that the complexity of causal relationships in real scenarios makes it very difficult to systematically solve the ``hullucination'' problem through manual labeling.

To address this challenge, we propose an \textbf{automatic} method for identifying causal rules in specific scenarios and evaluating models' \textbf{causal understanding}. %For each scenario, we automatically identify relevant factors and their relationships, constructing a \textbf{causal graph} to represent direct causal links and a \textbf{causal system} that characterizes the functional relationships between these variables through a carefully designed large language model (LLM) multi-round reasoning process. We demonstrate that the automatic generation can effectively accomplish this step by comparing them with human annotations.
Our process, utilizing an LLM assistant, identifies possibly involved causal factors and rules between them for a given scenario, and then describes them as a causal system. The automatic generation proves effective by comparing them with human annotations. \textbf{Intervention experiments}~\citep{causality2009pearl}, one of the effective methods in causal inference, are then used to assess causal behaviors in VGMs by varying the text prompts with different factor values.
For example, as shown in Figure~\ref{fig:bad_case}, replacing a heavy stone with a light feather revealed that the VGM had not correctly learned the causal rules related to density.

To analyze causal learning in VGMs, we define three levels of consistency: \textbf{text consistency} (to follow explicit cause conditions),  \textbf{generation consistency} (to maintain consistent generation under the same conditions) and \textbf{rule consistency} (to learn correct causal rules). %These three metrics correspond to ``the ability to follow explicitly given causes (and results)'', ``the consistency of generation under the same conditions'', and ``whether model learn correct rules''. These metrics assess the causal understanding and generative capabilities of the model, with progressively higher requirements.
Each of these three metrics forms the prerequisite for the next, creating a multi-level evaluation for the ``world simulator'' with progressively increasing difficulty.

% In this way, we propose an Video Automatic Causal Testing (VACT) system that is not limited to specific scenarios and does not involve human annotation, scoring and intervention. This makes our process highly generalizable and scalable, and can be applied to video content in various fields without additional manual annotation and intervention. 
In summary, we introduce the \textbf{V}ideo \textbf{A}utomatic \textbf{C}ausal \textbf{T}esting (\textbf{VACT}) system, which requires no human annotation, scoring, or intervention. To our knowledge, this is \textbf{the first approach to automatically apply causal analysis tools for testing causal understanding in VGMs}. It is scalable, generalizable, and can be applied across various fields without additional manual effort, while also providing a detailed causal analysis of model behavior. To validate its effectiveness and generalizability, we conducted crowd experiments, where 60 causal systems under 20 different scenarios by our system (involving various scenarios such as motion, force, light, heat, fluid, material, etc.) are compared to human annotation, showing that automatic annotations achieve comparable (and even better) performance with human annotation. We also use these 60 systems to construct a benchmark to assess current video generation models, revealing that no existing model achieves satisfactory causal learning. This system offers a powerful tool to enhance our understanding of VGM reliability and lays the groundwork for a systematic solution to the hallucination problem, like dataset supplementation or alignment by reinforcement learning.

\section{Related work}

\paragraph{Text-to-Video (T2V) generation models}
T2V models %~\citep{esser2023structure,videoworldsimulators2024,wang2023modelscopetexttovideotechnicalreport,khachatryan2023text2video} 
generate videos from textual descriptions. Early methods using generative adversarial networks (GANs)~\citep{wang2020imaginator} and variational autoencoders (VAEs)~\citep{li2018video, pan2017create} faced limitations like low resolution and diversity.
%However, these methods were constrained by issues such as low resolution, limited diversity, and an emphasis on single, isolated movements~\citep{xing2024survey}.
Starting with Video Diffusion Models~\citep{ho2022video}, recent advances in diffusion models have significantly improved T2V generation. %, with Video Diffusion Models (VDM) extending diffusion models to the video domain. % Diffusion Transformers (DiT)~\citep{peebles2023scalable} improved temporal consistency by leveraging a transformer backbone that models complex temporal dynamics and long-range dependencies through attention mechanisms~\citep{hong2022cogvideolargescalepretrainingtexttovideo}.
CogVideo~\citep{hong2022cogvideolargescalepretrainingtexttovideo} combines a pre-trained text-to-image model with a T2V framework, facilitating effective learning.  LaVie~\citep{wang2024lavie} enhances video quality with interpolation and super-resolution techniques. VideoCrafter2~\citep{chen2024videocrafter2} leverages Diffusion Transformers(DiT)~\citep{peebles2023scalable} to synthesize high-quality videos by refining generated sequences with high-resolution images. Models like Gen-3 Alpha~\citep{esser2023structure}, %, developed by Runway, leverages a new infrastructure designed specifically for large-scale multimodal training. Another leading model,
HunyuanVideo~\citep{kong2025hunyuanvideosystematicframeworklarge}, % employs a Transformer architecture with a Full Attention design for video generation. In addition, 
Haoluo~\citep{Hailuo}, pika~\citep{Pika}, Kling~\citep{Kling},
and Sora~\citep{sora_is_here}
further push the boundaries with advanced architectures and techniques. %(About the technique, we leave a discussion in Appendix~\ref{app:prompt_enhance}.)
%, adopts \textit{enhanced language processing} techniques to improve video continuity and detail quality. 
% A clear trend in this field is the increasing involvement of both open-source and closed-source models, driving rapid progress in text-to-video generation. 
%The advancements in this field have been comprehensively reviewed in~\citet{xing2024survey} and~\citet{sun2024soraseesurveytexttovideo}, which offer valuable insights into the evolution of text-to-video generation, including key model architectures and the historical progression of the domain.
Comprehensive reviews on the developments are available in~\citet{xing2024survey} and~\citet{sun2024soraseesurveytexttovideo}.
% However, even the latest T2V models face limitations in causal inference, struggling to align generated videos with real-world causal rules effectively.

\paragraph{Evaluation for video generation models}
%The rapid advancement of video generation models has highlighted the need for accurate video quality evaluation. Traditional methods, such as IS~\citep{salimans2016improved} and FVD~\citep{unterthiner2019accurategenerativemodelsvideo}, CLIP~\citep{hessel2022clipscorereferencefreeevaluationmetric, liu2023fetvbenchmarkfinegrainedevaluation} focus on very limited aspects such as frame quality or frame and fail to make evaluations that are consistent with human values.
%A growing need for more comprehensive and reliable metrics prompts benchmarks like V-Bench\citep{huang2024vbench} and EvalCrafter~\citep{liu2024evalcrafterbenchmarkingevaluatinglarge}, offering comprehensive evaluations across multiple dimensions such as subject consistency, spatial relationship, dynamic degree, human action continuity, and so on. However, these indicators still evaluate the ``visual quality'', ignoring the logic about the events and scenes in the video. 
The rapid advancement of VGMs has underscored the need for accurate quality evaluation. Traditional metrics like IS~\citep{salimans2016improved}, FVD~\citep{unterthiner2019accurategenerativemodelsvideo}, and CLIP~\citep{hessel2022clipscorereferencefreeevaluationmetric, liu2023fetvbenchmarkfinegrainedevaluation} assess only limited aspects like frame quality, and often fail to align with human judgment. To address this, benchmarks like V-Bench~\citep{huang2024vbench} and EvalCrafter~\citep{liu2024evalcrafterbenchmarkingevaluatinglarge} provide more comprehensive evaluations, considering factors like subject consistency, spatial relationships, dynamic degree, and action continuity. However, these metrics still focus on visual quality while overlooking the logical coherence of events and scenes in videos.

\paragraph{Evaluation for world simulators}
As video quality further improves and the concept of a ``\textit{world simulator}'' becomes an expectation, the focus has shifted from \textit{aesthetics} to \textit{authenticity} --- ensuring generated content follows real-world physics rules. Recent benchmarks including VideoPhy~\citep{bansal2024videophyevaluatingphysicalcommonsense} and PhyGenBench~\citep{meng2024worldsimulatorcraftingphysical} have made initial attempts to address this.
%As the quality of videos continues to improve and the concept ``world simulator'' starts to be an expectation, the ``authenticity'' of video content instead of ``aesthetics'' has been required, that is, the video generated content should conform to the basic physics and natural laws of the real world. Recently, some preliminary work has proposed benchmarks that include real physical scene. 
VideoPhy uses human collection and evaluation to verify commonsense violations within three classes of physics scene: solid-solid, solid-fluid and fluid-fluid. Their benchmark heavily depends on human efforts and is hard to generalize to new field. Their attempts to fine-tune a vision-text model for automatic ranking have yet to align well with human assessments, limiting its scalability.
%VideoPhy~\citep{bansal2024videophyevaluatingphysicalcommonsense} chooses some text prompts within three classes of physics scene: solid-solid, solid-fluid and fluid-fluid, with LLM generation and human evaluation. After the videos have been generated, human annotators are required to check whether the required content and rule violation has been observed in videos, making their benchmark heavily depends on human efforts and is hard to generalize to new field. They also try to finetune a vision-text model to auto-rank the videos. However, this model is far from being aligned with human annotations, making this technology still not suitable for large-scale use on benchmarks. 
PhyGenBench~\citep{meng2024worldsimulatorcraftingphysical} tests on 27 \textit{human-designed} physics laws, using LLM-generated questions to check rule fidelity in videos by a video language model. However, it remains human-dependent in rule design and does not distinguish between true causal understanding and shortcut-based pattern recognition. For instance, a model generating ``stone splashing water'' doesn't prove it understands the physics of the splash, as testing with a feather (density) or other setting is necessary.
%PhyGenBench~\citep{meng2024worldsimulatorcraftingphysical} extends to generate text prompts in 4 different aspects and explicit 27 physics laws, where each text prompt is used to test on a certain laws. With the explicit labeling of the target laws, they use an LLM to generate some questions and then use a VLM to read the videos and the question to check whether the target rules is obeyed. Though the automatic tools have been utilized, because the candidate rules and prompts are prepared by human. At the same time, the indicator they care about is ``whether the expected physical phenomenon occurs''. They cannot distinguish whether ``\textit{the model has learned the causal relationship}'', or ``\textit{simply simulated common physical results} (some shortcuts)''. Specifically, when we observe ``stone splashing water'', this phenomenon does not mean that the model understands the relevant physical laws. Only through intervention experiments like replacing the stone with a feather, replacing the rapid throw with a gentle placement, or replacing the water surface with a thicker liquid, etc., can we evaluate the model's understanding of various complex relationships in the rule and avoid shortcuts.

As concurrent works, \citet{motamed2025generative} evaluate whether VGMs learn physical principles by predicting video continuations. They first film 396 videos, each divided into two segments: a conditioning segment and a subsequent ground truth segment. Predictions generated by the models based on the conditioning segment are compared against the ground truth to assess accuracy. \citet{li2025worldmodelbenchjudgingvideogeneration} collect image and text conditions about 5 common physics laws such as Newton's first law and fluid mechanism. They finetune a VLLM using extensive human annotations to identify the potential errors in videos.

Our work further expands this series of work in two aspects: 
%In short, our article further expands this series of work in two aspects: 
1) \textbf{Full automation}: our approach eliminates manual rule design, allowing physical rules to be automatically inferred from a short textual descriptions, enhancing scalability.
2) \textbf{Causal evaluation}: We introduce \textit{intervention experiments} to test whether models truly understand physics rather than relying on shortcuts, ensuring a more robust assessment.
%First, we further improve the automation of the entire process, so that human annotators no longer need to design implicit or explicit physical rules. Only by entering a text description can the automatic construction of physical rules be completed, so as to have better scalability; second, we introduced causal analysis tools to truly evaluate the model's understanding of the law through "intervention experiments" to avoid the model from achieving false high scores through shortcuts.

Additionally, other works like \citep{kang2024farvideogenerationworld} explore 2D physics simulation in VGMs, while WorldSimBench~\citep{qin2024worldsimbenchvideogenerationmodels} assesses world simulators from an embodied perspective. These works, along with ours, collectively contribute to a multi-faceted understanding of world simulators,
%There are also some other works focus on the understanding of real world in generated vidoes. \citet{kang2024farvideogenerationworld} focuses on the 2D physical simulation capabilities of VGMs. WorldSimBench\citep{qin2024worldsimbenchvideogenerationmodels} assesses World Simulators from an embodied perspective.

%Another study~ VideoPhy\citep{bansal2024videophyevaluatingphysicalcommonsense} and PhyGenBench\citep{meng2024worldsimulatorcraftingphysical} provide insights into how T2V generation models capture physical commonsense. To bring models closer to real-world physics, \\
%However, these metrics fall short when it comes to evaluating a model’s ability to perform causal reasoning based on fundamental rules, which is crucial for generating rational videos, particularly when the prompt scenarios are out of distribution. Unlike specific physical rules, causal inference requires models to track and adapt to changes in multiple influential factors across several rounds, comparing the resulting videos to ensure that the model is not simply memorizing fixed patterns. To fill this gap, we introduce a benchmark that features an automated pipeline for constructing testbeds and offers comprehensive methods for measurement, specifically aimed at assessing causal inference capabilities.
%\subsection{Benchmark and evaluation of physics in videos}

\section{Pipeline of automatic causal rule testing}

\subsection{Scenario-based causal rule testing}

% Our tests are started from some \textbf{scenario}s. A \textbf{scenario} is defined as a short text description of an event, like ``something is thrown into a swimming pool'' in Figure~\ref{fig:bad_case}. Given a scenario, the properties of some objects or events involved can be regarded as variables, and there are causal relationships between these variables. A causal graph and a causal system can be used to model these relationships. Here, we call the functions to define the relationships as \textbf{causal rules}. At the same time, the rules consist of a \textbf{test case}, which is the unit of our tests. As mentioned in the introduction, a scenario may have different causal choices and therefore can correspond to multiple different test cases, if we choose different factors (therefore, different rules). Here, we provide a mathematical definition of the causal graph and causal system.
Our tests begin with \textbf{scenario}s, short text descriptions of an event, such as ``something is thrown into a swimming pool'' (in Figure~\ref{fig:bad_case}). Each scenario involves variables representing object or event properties, linked by causal relationships modeled as a causal graph and a causal system.

\begin{definition}[Causal graph and system~\citep{causality2009pearl}] 
A deterministic \textbf{causal system} over a set of variables $\mathbf{V}$ is a directed acyclic graph $G$ with the node set $\mathbf{V}$ and edge set $\mathbf{E}$, and a series of structural equations $V_j=f_j(pa(V_j))$ for every $V_j\in \mathbf{V}$, where $pa(V_j)=\{V_k\in \mathbf{V}:V_k\to V_j\in \mathbf{E}\}$ are the parents of the node $V_j$. The graph $G$ is called the \textbf{causal graph}. Furthermore, let $\mathbf{X}=\{V_j\in\mathbf{V}: pa(V_j)=\emptyset\}$ be the \textbf{root} (\textbf{cause}) variables and $\mathbf{Y}=\mathbf{V} \setminus \mathbf{X}$ be the \textbf{non-root} (\textbf{outcome}) variables.

\end{definition}

\begin{figure}[t!]
    \centering
    \includegraphics[width=0.8\linewidth]{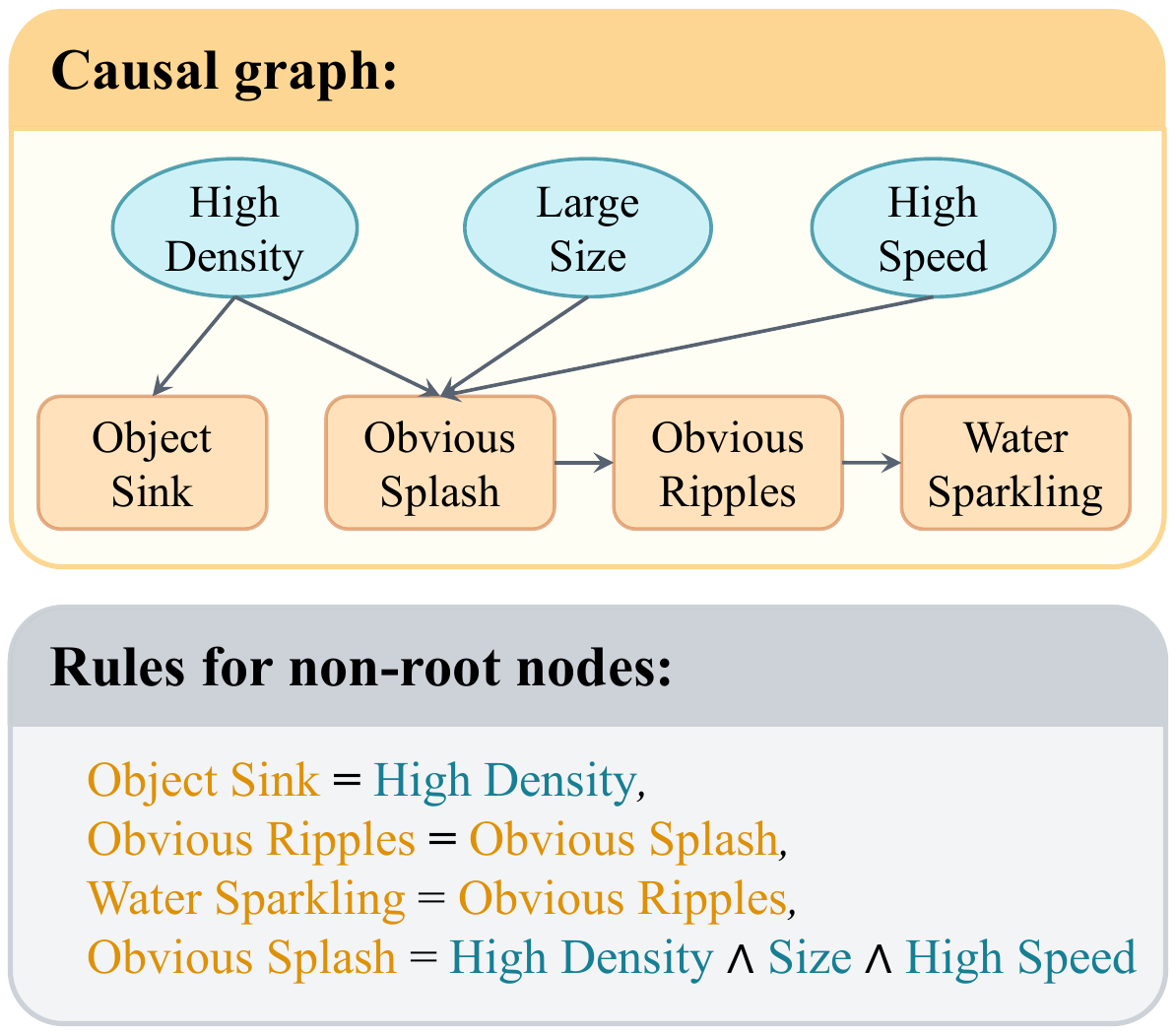}

    \caption{An example causal graph and system: ``\textit{throwing something into a swimming pool}''. \textcolor{cyan}{Blue} denotes root nodes and \textcolor{orange}{orange} denotes non-root nodes. Physical explanation can be found in~\citep{dropimpact} Figure 6.}
    \label{fig:causal_graph_pool}
\end{figure}

\begin{figure*}[t]
    \centering
    \includegraphics[width=0.85\linewidth]{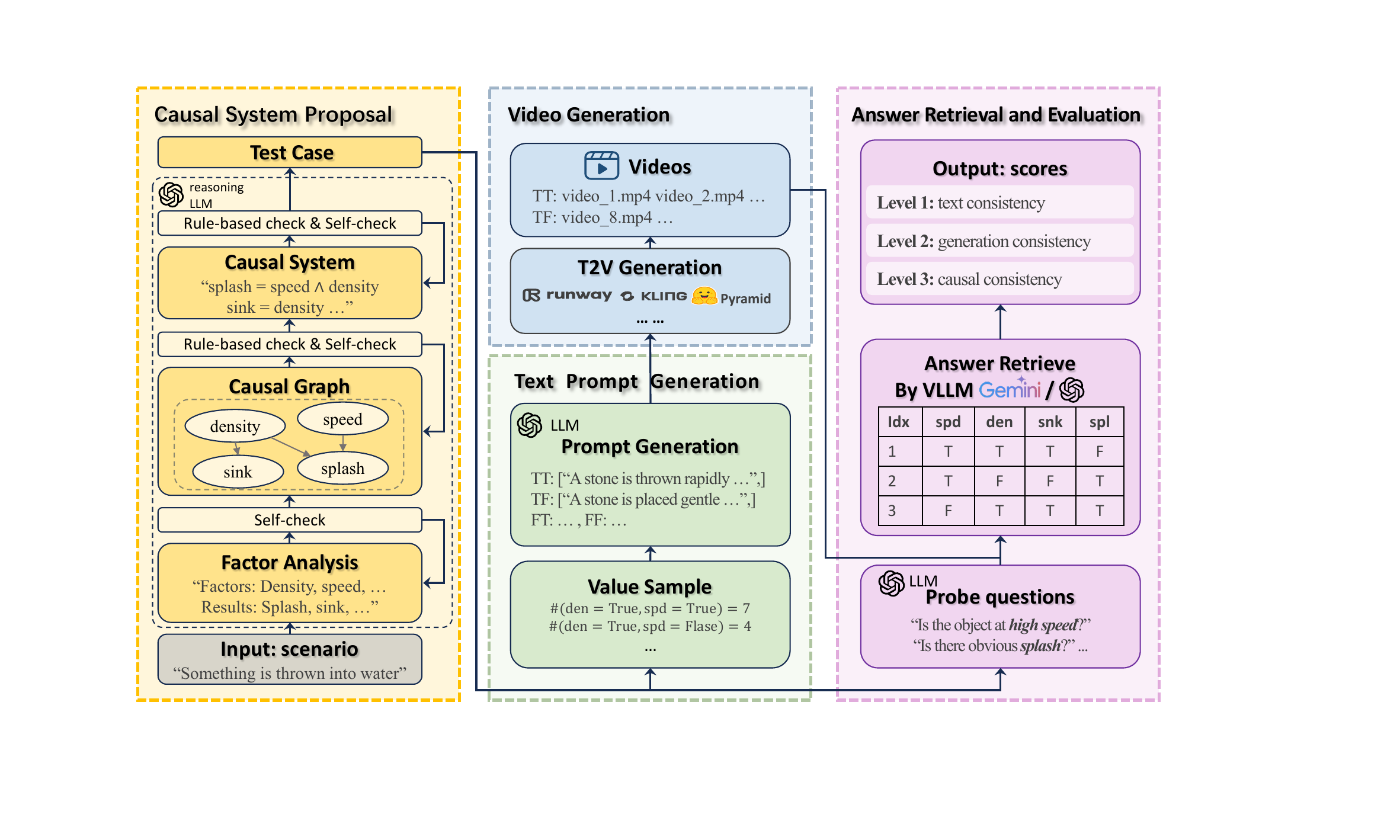}
    
    \caption{Pipeline of VACT. The pipeline mainly consists of four parts: causal system (\ie test case) proposal (\textcolor{Goldenrod}{yellow}), text prompt generation, (\textcolor{LimeGreen}{green}), video generation (\textcolor{Cyan}{blue}), answer retrieval and evaluation (\textcolor{pink}{pink}). The pipeline receive a sentence describe a scenario as input and automatically evaluate video generation models \textit{without} any human supervision or annotation.}
    
    \label{fig:pipeline}
\end{figure*}

We illustrate this with an example in Figure~\ref{fig:causal_graph_pool}, where the system captures commonsense physical knowledge—for instance, density determines whether an object will sink, while speed, size, and density collectively influence the splash.
The directed edges in the graph represent causal relationships between variables, such as the edge ``high density'' $\to$ ``object sink'' indicating causation, while there is no causal effect between ``large size'' and ``object sink'', since a dense object will sink regardless of its size.
Root variables (\textcolor{cyan}{blue}) are basic conditions of a scenario that can be directly manipulated. Non-root variables (\textcolor{orange}{orange}) are outcomes of other variables that cannot be directly adjusted but may influence other factors. For example, the non-root variable ``splash'' causes ``ripples'', which in turn causes ``sparkling''~\citep{dropimpact}.
The basic unit of our VACT is a causal system, consisting of these rules (\ie the equations). One scenario may generate different test cases depending on the selected factors.
%and $\mathbf{Y}=\mathbf{V} \setminus \mathbf{X}$. Then $\mathbf{X}$ is called the set of root (or cause) variables, and $\mathbf{Y}$ is called the set of non-root (or outcome) variables. 
%From the definition we can conclude that $\mathbf{Y}=f(\mathbf{X})$ for some function $f$. We call $f$ the \textbf{rule} in this scenario.

% In this article, for the sake of clarity, we require all variables to take Boolean values. Therefore, a causal system is a series of Boolean functions. These Boolean functions are a reasonable simplification of various physical relationships in nature. For those physical properties that take continuous values, we can actually avoid complex calculations and highlight the ``essence'' of their causal relationships by binarizing them --- for example, simplifying the continuous value of speed to ``fast'' and ``slow'', and simplifying the mass of an object to ``light'' and ``heavy''. In fact, this simplification is also in line with human commonsense, because in daily life we generally use these binary rather than continuous values to make commonsense judgments and think.

For the sake of clarity, all variables in our system are \textit{Boolean}, meaning the rules are Boolean functions. This simplification reasonably abstracts physical relationships, avoiding complex calculations while preserving essential causal structures. Continuous properties, such as speed or mass, can be binarized (\eg ``fast'' \vs ``slow'' or ``light'' \vs ``heavy''), as we often make judgments using such discrete categories in daily life. Additionally, the variables must be visually discernible\footnote{The visualization here is a relative requirement. For example, although the density is essentially invisible, we can infer the density of an object through its visible material.} (\textit{visibility}) to ensure suitability for video evaluation, and all root nodes can be set \textit{independently}.
If a video generation model learns the physical laws of a scenario, it should have learned the rule $f$. Thus, by analyzing variable states in generated videos under different conditions, we can assess whether the model understands the underlying law.

\subsection{LLM-aided automatic generation of test cases}
\label{subsec:auto_label}

As discussed in Section~\ref{sec:intro}, extracting key causal rules from scenarios is challenging due to their inherent complexity and diversity. This task requires both creativity (to imagine alternative scenarios) and commonsense reasoning (to recognize common causal patterns). Fortunately, the advanced commonsense reasoning capabilities of Large Language Models (LLMs) enable automation of this process. We designed a multi-step annotation method that leverages LLMs, incorporating task splitting, rule checking, and self-correction to ensure accuracy and reliability.

As illustrated in Figure~\ref{fig:pipeline} (\textcolor{Goldenrod}{yellow} part), our system accepts a scenario (a brief textual description) and prompts the LLM to sequentially perform the following steps: 1) identify key causal factors and outcomes within the scenario, 2) discover relationships between these elements to construct a causal graph, and 3) derive Boolean expressions representing the discovered relationships.

After each step, we implement self-checking and rule-checking mechanisms. If errors are identified, such as formatting issues, unusual relationships, isolated nodes in the causal graph, or mismatched Boolean expressions, we provide explicit error feedback to the LLM, prompting it to self-correct. If no errors are found, the model is prompted to perform an additional self-check. The step-splitting and self-correction strategy ensure that errors are detected and corrected, significantly improving the overall quality of generated outputs. The final test case comprises both the Boolean expression and the original scenario text. Detailed generation requirements, inspection indicators, and a comprehensive description of the process can be found in Appendix~\ref{app:automatic_generation_detail}.

\paragraph{Crowd experiments} 
We evaluate the effectiveness of the automatic generation by crowd experiments. We collected 20 diverse scenarios (listed in Appendix~\ref{app:scenario_list}) and generated three causal systems for each, 60 in total. For comparison, three undergraduates manually annotated the same scenarios using identical instructions given to the LLM, resulting in another 60 \textit{human-annotated} causal systems. Then, another five undergraduates blindly scored both human and LLM annotations based on three criteria: \textit{requirement} (adherence to visibility, binarity and root independence), \textit{rationality} (reasonableness of factor selection), and \textit{soundness} (accuracy of causal rules). For each criteria, the scores range from 1 which means there are essential error to 4 which means the annotation perfectly meets the requirement.  The average score from 5 scorer and 60 samples are reported in Table~\ref{tab:crowd_average}. LLM-generated annotations surprisingly \textbf{outperformed} those from human, demonstrating the effectiveness of the LLM-driven process and its strong alignment with human reasoning. 
We also provide the details of the crowd experiments, the score distribution of each scorer and further analysis in Appendix~\ref{app:crowd_exp}.

\addtolength{\tabcolsep}{-3pt} 
\begin{table}[htpb]

    \caption{Average scores from crowd experiments}

    \label{tab:crowd_average}
    \centering
    \begin{tabular}{lcccc}
    \toprule
    Source& Requirement & Rationality & Soundness & Average \\ \midrule
    \textbf{LLM}   &  $\textbf{3.91}\mathsmaller{\pm 
 0.02}$ & $\textbf{3.49}\mathsmaller{\pm 0.04}$ & $\textbf{3.78}\mathsmaller{\pm 0.03}$ & $\textbf{3.73}\mathsmaller{\pm 0.02}$\\
    \textbf{Human} &  $3.80\mathsmaller{\pm 0.03}$ & $\textbf{3.51}\mathsmaller{\pm 0.04}$ & $3.63\mathsmaller{\pm 0.04}$ & $3.65\mathsmaller{\pm 0.03}$ \\ \bottomrule
    \end{tabular}
\end{table}

\subsection{Automatic intervention experiment pipeline}
\label{ssec:automatic}

Given a causal system, our testing as intervention experiments contains five parts: sampling, prompt generation, video generation, answer retrieval, and evaluation, as shown in Figure~\ref{fig:pipeline}. Details and the prompt used in these steps are in Appendix~\ref{app:test_pipeline}.

\textbf{Sampling}.
In our intervention experiments, observations are videos characterized by varying condition values $\mathbf{X}$.
We sample diverse combinations of root values $\mathbf{X}$ for these intervention. The sample size for each $\mathbf{X}$ value combination is determined by metrics discussed in Section~\ref{sec: three levels}. Given the high cost of video generation, it is crucial to minimize the number of videos needed while ensuring accurate and stable metrics measurement. Therefore, we (1) conduct preliminary experiments to determine the minimum number of samples that can ensure the relative stability and validity of each metric and (2) strategically share generated videos across different metrics whenever feasible. Finally, the number of sample for each $\mathbf{X}$ value combination is set to the maximum requirement across the three metrics. We provide the details of samples in Appendix~\ref{app:sample_strategy} and the preliminary experiments used to determine the minimum number in Appendix~\ref{app:sample_size_exp}. Through our experiments, we conduct that approximately 30 to 45 videos per causal system are sufficient to achieve stable evaluations.
%We sample some combinations of root values $\mathbf{X}$ as intervention experiments. The number of samples for each $\mathbf{X}$ value is determined by the metrics we use (in Section~\ref{sec: three levels}) and described in Appendix~\ref{app:sample_strategy}. In our experiments, we took approximately 20 - 40 (\textcolor{red}{need to check}) samples for each causal system. 

%specify the value of root nodes $\mathbf{X}$, that is, we manipulate the state of cause variables in the scenario. The distribution we set for $\mathbf{X}$ is related to the metric we use, described in Section~\ref{sec: three levels}.

\textbf{Prompt generation}. 
Given $\mathbf{X}$ values, we use an LLM to generate sentences to constrain restrict the variables in the scenario as meeting the given values. For example, in the scenario ``throwing something into a swimming pool'', the prompt ``\textit{a large rock was thrown quickly into the pool}'' sets three root variables: high density, large size, and high speed to true, while another prompt ``\textit{a tiny stone is gently placed on the water}'' alert large size and high speed to false. These sentences serve as text prompts for video generation.

%Given the values of $\mathbf{X}$, we can generate a sentence to restrict the variables in the scenario as meeting the given values. For example, for the scenario ``throwing something into the swimming pool'', a prompt like `` \textit{a large rock was thrown quickly into the pool}'' can restrict the values of factors ``\textit{high density}'', ``\textit{large size}'' and ``\textit{high speed}'' as all true; while `\textit{`a tiny stone is gently placed on the water}'' corresponding the ``\textit{large size}'' and ``\textit{high speed}'' as false.
% These sentences are used as text prompts for the generation of videos. %We consider two types of prompts: coarse and detailed. For the coarse prompt, we only specify the value $\mathbf{X} = \mathbf{x}$; for the detailed prompt, we specify both $\mathbf{X} = \mathbf{x}$ and $\mathbf{Y} = f(\mathbf{x})$. The formal one is only used for % In either case, if VGMs have sufficient text understanding and conditioned generation capabilities, they can generate videos that match the given $X$ and corresponding $Y$ values.

\textbf{Video generation}. 
The prompts generated are provided to the tested VGM. These models are treated as black boxes, requiring no constraints on their structure.

\textbf{Answer retrieval}.
Each generated video serves as an observation of the intervention experiments for the causal system. We check (1) whether it follows the text description of variable values $\mathbf{X}$ and (2) whether the generated values $\mathbf{Y}$ align with the causal rules. 
Following \citet{meng2024worldsimulatorcraftingphysical}, we use a vision-LLM (VLLM) to retrieve the observed values $\hat{\mathbf{X}}$ and $\hat{\mathbf{Y}}$ in each video, by prompting the VLLM with the video and some ``\textit{yes-no}'' questions (\ie \textit{probes}). These probes are also generated by an LLM given the variable list of a causal system. Notice that our variable has been required to be binary when generating test cases, so the generation of these questions is a very simple tasks.

%Each video generated is an observation of the intervention experiments about this causal system. To measure the causal understanding of VGMs, we need to know (1) whether they faithfully follow the text's description of the choice of variable values $\mathbf{X}$ and (2) whether they generate values $\mathbf{Y}$ that match the causal rules? Like~\citet{meng2024worldsimulatorcraftingphysical}, we use an multi-modal LLM (VLLM) with video understanding ability to retrieve the value of both $\mathbf{X}$ and $\mathbf{Y}$ in the videos. Specifically, we prompt a generated video with some ``yes-no'' questions (which is generated simultaneously when the graph is generated) to the VLLM and collect the answer as the ``\textit{observed value}'' (denoted by $\hat{\mathbf{X}}$ and $\hat{\mathbf{Y}}$).
% use a large language model that can accept both video and text as input to query the content of the video, thereby obtaining the observed states of the variables in the scenario within the generated video, denoted by $\hat{\mathbf{X}}$ and $\hat{\mathbf{Y}}$.

%\textbf{Evaluation}. For the video generation model, the ideal scenario is when $\mathbf{X} = \hat{\mathbf{X}}$ and $\mathbf{Y} = \hat{\mathbf{Y}}$. However, when they are not equal, it may indicate errors in the generation process or the model's understanding of the objective world. In Section~\ref{sec: three levels}, we will propose several metrics to measure these deviations.
% With the observation, we can evaluate the causal behavior of the tested model. In Section~\ref{sec: three levels}, we will propose several metrics.

We adopt an LLM and a VLLM to automate the steps \textit{prompt generation}, \textit{probe generation} and \textit{Answer retrieval}. To ensure the correctness, we performed random manual checks. We found that the vast majority of the results are reliable. The check results are shown in Appendix~\ref{app:manual_check}.

\section{Three levels of causal ability and the corresponding metrics}

\label{sec: three levels}
% In the main part of the section, we introcude the scenario-based scores.
To assess the deviation of the model's understanding of the objective world, we propose a three-level framework of causal capabilities with corresponding evaluation metrics. The mathematical definitions are provided in Appendix~\ref{app:detailed_def_for_metrics}. Here, we focus on an intuitive description.

\paragraph{Text consistency}
The first level assesses whether the model accurately reflects the state of every variable described in the prompt. By generating a video from a detailed prompt specifying variable values, the resulting video should correctly reflect those values. It is a fundamental requirement for not only the general usage of video generation models but also our intervention experiments because we need to control video variables through text.
The similar metrics have proposed in some previous benchmarks, such as video-text consistency in VBench~\citep{huang2024vbench}, but it is still different. In our test, a video generation requires multiple important but complex attributes simultaneously, and there are some uncommon combinations. Therefore, our test is more difficult than similar evaluations.

We use two types of prompts: the standard prompts (described in Section~\ref{ssec:automatic} and also used in the next two metrics) specifies \textit{all root} variables $\mathbf{X}$. In addition, we generate prompts constraining all variable values (including roots $\mathbf{X}$ and non-roots $\mathbf{Y}$), describing not only the condition but also expected results in the scenario. These two types correspond to the ``\textit{root}'' score and ``\textit{all}'' score, respectively. 
The first type aligns with common real-world application, where the results are not stated before generating. The second serves as a complementary test about whether explicitly stating expected outcomes helps the model generate correct physical behaviors. For each setting, we measure text consistency using the \textit{average accuracy} of whether the observed values match the described ones.

% The metric $s_1^{\mathrm{root}}\propto \sum_{i,V\in\mathbf{V}}1(X_i=\hat{X}_i)$ measures the accuracy across all variables. Additionally, we also introduce $s_1^{\mathrm{all}}\propto \sum_{i,X\in\mathbf{X}}1(X_i=\hat{X}_i)$, which focuses on the consistency of cause variables. These metrics evaluate whether the model has learned to maintain consistent representations of the input variables, reflecting its ability to adhere to the provided descriptions.

\paragraph{Generation consistency}
The second level evaluates whether the model stably produces the stable and predictable outcomes given identical causes $\mathbf{X}$, or if its outcomes vary arbitrarily due to unrelated factors like random seed or wording differences.
%The second level evaluates whether the model consistently produces the same outcomes given identical causes $\mathbf{X}$. This indicates whether the model has internalized stable input-output rules, or if its outcomes vary arbitrarily due to unrelated factors like random seed or wording differences.
To measure this, we group the samples by identical $\mathbf{X}$ values, and calculate the mean variance of outcomes $Y_j$ within each group. Since grouping is based on $\mathbf{X}$, errors arise if text consistency (level one) is imperfect. To address this, we use two scoring criteria: ``\textit{truth}'' score groups samples according to expected $\mathbf{X}$ in text prompt and evaluates end-to-end consistency in practical usage, while ``\textit{observe}'' score groups samples according to actually observed $\hat{\mathbf{X}}$ which is retrieved from videos and ignores condition generation errors. Notice that as text consistency improves, both scores should converge to the same. %$\sum_{\mathbf{S}}\sum_{i=1,\dots,|\mathbf{Y}_\mathbf{S}|}\operatorname{Var}(Y_i)/(|\mathbf{Y}_\mathbf{S}||\mathbf{S}|)$ as the score. 

\paragraph{Rule consistency}
The third level, our main and long-term goal, tests the model’s ability to learn and apply causal rules consistent with the real world. For one test case, we first calculate the accuracy for each rule (\ie for each outcome $Y_j\in\mathbf{Y}$).
For each outcome $Y_j=f(pa(Y_j))$, we sample prompts to make sure the groundtruth value of $Y_j$ is 50\% True and 50\% False. 
Having sampled videos $\mathbf{S}=\{s^{(1)},\dots,s^{(n)}\}$, the rule consistency is calculated as (1) the average accuracy $\sum_{i=1}^n \mathds{1}(Y^{(i)}_j=\hat{Y}^{(i)}_j)$, or (2) a threshold-based 0-1 score $\mathds{1}\{\operatorname{mean}[\mathds{1}(Y_j=\hat{Y_j})]\geq t\}$. 
Then the score of the test case take the average value among variables $\mathbf{Y}$. Here, we also distinguish two scores, ``\textit{truth}'' score using the groundtruth $pa(Y_j)$ and ``\textit{observe}'' score using the observed $\hat{pa}(Y_j)$ to get the expected $Y_j=f_j(pa(Y_j))$ where the latter isolates the direct cause-effect relationship, 
excluding errors from unexpected causes.

%For each outcome variable, we generate samples with balanced ground truth values (true or false) and measure accuracy using $s_3^{\mathrm{truth}}\propto \sum_{i,Y\in\mathbf{Y}}1(Y_i=\hat{Y}_i)$. Higher accuracy indicates that the model has learned consistent rules for inferring outcomes under varying conditions. 
%To mitigate text inconsistency, we also use observed root variables $\hat{\mathbf{X}}_i$ to compute $\tilde{Y}_i$ and measure $s_3^{\mathrm{observe}}\propto \sum_{i,Y\in\mathbf{Y}}1(\tilde{Y}_i=\hat{Y}_i)$. These metrics are the most critical, as they directly evaluate whether the model's learned rules align with the ground truth, reflecting its ability to capture and apply causal relationships.

\paragraph{Scores on samples} Though our experiments need comparison and evaluation on the scenario level where tens of videos could be involved, these metrics can be also applied to individual videos by redistributing the scores of each video according to its proportion in the overall score. It convenient to identify specific instances where the model's performance deviates, provide insights into its learning mechanisms or provide reward in a reinforcement learning stage. See Appendix~\ref{app:detailed_def_for_metrics} for detailed definitions and some example analysis in Appendix~\ref{app:example_score_based_analysis}.

% \section{An automatic generation of testbed}
% \subsection{The automatic generation of ground-truth causal systems}
% \subsection{Crowd experiments}
% \label{subsec:crowd_exp}

% \subsection{The automatic generation of test prompts}

% \subsection{Automatic testing}

\section{A benchmark of causal rule testing}
\begin{table*}[htbp]
\centering

\caption{VACT benchmark on prevailing VGMs. The rule consistency is calculated by the ``average accuracy''.}

\label{tab:main_benchmark}
\begin{tabular}{lccccccc} 
\toprule
\multirow{2}{*}{\textbf{Model Names}} & \multirow{2}{*}{\textbf{N/A ratio}} & \multicolumn{2}{c}{\textbf{Text Consistency} $\uparrow$} & \multicolumn{2}{c}{\textbf{Generation Consistency} $\downarrow$} & \multicolumn{2}{c}{\textbf{Rule Consistency} $\uparrow$} \\
\cmidrule(lr){3-4} \cmidrule(lr){5-6} \cmidrule(lr){7-8}  
& & all & root & truth & observe & truth & observe \\
 \midrule
% Random Generator& 1.0& -&- &- & -& -& - \\
% Stubborn Generator&0.0& 0.5& 0.5 & 0.0 & - &0.5 & -  \\
% Text-loyal Generator&0.0& 1.0 &1.0 &0.25 &0.25 & 0.5&0.5 \\
% Consistent Generator&0.0& 1.0 & 1.0 & 0.0 & 0.0 & $\sim$&$\sim$ \\
% Causal Perfect Generator&0.0& 1.0 & 1.0 & 0.0 & 0.0& 1.0& 1.0\\
 
% \midrule
CogVideoX1.5-5B & .07 & $.56\mathsmaller{\pm .01}$ & $.61\mathsmaller{\pm .01}$ & $.10\mathsmaller{\pm .00}$ & $.09\mathsmaller{\pm 
.01}$ & $.55\mathsmaller{\pm .01}$ & $.72\mathsmaller{\pm .02}$\\
CogVideoX-5B & .07 & $.58\mathsmaller{\pm .01}$ & $.64\mathsmaller{\pm .02}$ & $.09\mathsmaller{\pm .00}$ & $.09\mathsmaller{\pm .01}$ & $.56\mathsmaller{\pm .01}$ & $.71\mathsmaller{\pm .03}$ \\
CogVideoX-2B & .09 & $.56\mathsmaller{\pm .01}$ & $.63\mathsmaller{\pm .01}$ & $.09\mathsmaller{\pm .01}$ & $.09\mathsmaller{\pm .01}$ & $.59\mathsmaller{\pm .02}$ & $.72\mathsmaller{\pm .03}$ \\
VideoCrafter2  & .12 & $.55\mathsmaller{\pm .01}$ & $.58\mathsmaller{\pm .02}$ & $.08\mathsmaller{\pm .01}$ & $.06\mathsmaller{\pm .01}$ & $.53\mathsmaller{\pm .02}$ & $.73\mathsmaller{\pm .03}$ \\
Pyramid Flow & .10 & $.56\mathsmaller{\pm .01}$ & $.61\mathsmaller{\pm .02}$ & $.07\mathsmaller{\pm .00}$ & $.06\mathsmaller{\pm .01}$ & $.56\mathsmaller{\pm .01}$ & $.72\mathsmaller{\pm .03}$ \\
HunyuanVideo & .07 & $.58\mathsmaller{\pm .01}$ & $.63\mathsmaller{\pm .01}$ & $.08\mathsmaller{\pm .01}$ & $.07\mathsmaller{\pm .01}$ & $.57\mathsmaller{\pm .01}$ & $.70\mathsmaller{\pm .02}$ \\
\midrule
Pika & .10 & $.57\mathsmaller{\pm .01}$ & $.60\mathsmaller{\pm .01}$ & $.09\mathsmaller{\pm .00}$ & $.08\mathsmaller{\pm .01}$ & $.56\mathsmaller{\pm .01}$ & $.76\mathsmaller{\pm .02}$ \\
Hailuo & .07 & $.59\mathsmaller{\pm .01}$ & $.64\mathsmaller{\pm .01}$ & $.10\mathsmaller{\pm .00}$ & $.08\mathsmaller{\pm .01}$ & $.59\mathsmaller{\pm .01}$ & $.73\mathsmaller{\pm .02}$ \\
Gen-3 Alpha & .06 & $.63\mathsmaller{\pm .01}$ & $.63\mathsmaller{\pm .01}$ & $.08\mathsmaller{\pm .00}$ & $.08\mathsmaller{\pm .01}$ & $.57\mathsmaller{\pm .01}$ & $.74\mathsmaller{\pm .02}$ \\
Kling & .07 & $.63\mathsmaller{\pm .01}$ & $.64\mathsmaller{\pm .01}$ & $.07\mathsmaller{\pm .00}$ & $.07\mathsmaller{\pm .01}$ & $.57\mathsmaller{\pm .02}$ & $.71\mathsmaller{\pm .02}$ \\
\bottomrule
\end{tabular}

\end{table*}

In this section, we use the 60 causal systems from 20 different scenarios (collected in our crowd experiments in Section~\ref{subsec:auto_label}) as a testbed to evaluate the causal learning of prevailing VGMs. We found that these models occasionally generate videos that are off-topic or with missing subjects and confusing logic. To avoid the score being affected too severely by the low-quality generation, we allowed the VLLM to answer ``N/A'' (in addition to yes/no) during answer retrieval, filtering out all observations marked as ``N/A'' across all metrics and reporting the ``N/A'' ratio in results. Table~\ref{tab:main_benchmark} shows our benchmarking results on some prevailing open- or closed-source models.
Here, rule consistency is calculated as the average accuracy score. For details on the tested models, testing costs, the impact of N/A, and sample efficiency, see Appendix~\ref{app:model_details} to \ref{app:sample_size_exp}.

%In crowd experiments in Section~\ref{subsec:auto_label} we have collected 60 causal systems with 20 different scenarios. We use them as a testbed for prevailing VGMs to test their causal learning ability. %The cost of benchmark can be found in \ref{app:cost}.
%We found that these models often generate videos that are far from the subject (such as blur, missing subjects, and confusing logic). For this reason, we additionally allow VLLM to answer N/A (besides yes/no) in the answer retrieval step, filtering all observations with N/A in all metrics. %We discuss about the N/A more in Appendix~\ref{app:affect_of_na}. 
%The rule consistency is calculated by the average accuracy score. About the details of the models, the cost of benchmarking, the affect of N/A, sample efficiency, and threshold-based rule consistency, see Appendix~\ref{app:model_details} to \ref{app:thre_score}.

%Table~\ref{tab:main_benchmark} shows our benchmarking results on some prevailing open or closed-source models. There are some important observation: %The results of threshold score are shown in Appendix~\ref{app:thre_score}.

\begin{table*}[htbp]
\centering
\caption{Metrics for rule consistency by applying threshold for each rule.}
\label{tab:metric_level_3_threshold}
\begin{tabular}{l *{8}{c}}  
\toprule
\textbf{Name} & 
\multicolumn{4}{c}{Truth} & 
\multicolumn{4}{c}{Observe} \\
\cmidrule(lr){1-1} \cmidrule(lr){2-5} \cmidrule(lr){6-9}
\textbf{Threshold} & 0.65 & 0.75 & 0.85 & 0.95 & 0.65 & 0.75 & 0.85 & 0.95 \\
\midrule
CogVideoX1.5-5B  & $.19 \mathsmaller{\pm .04}$ & $.08 \mathsmaller{\pm .03}$ & $.02 \mathsmaller{\pm .01}$ & $.00 \mathsmaller{\pm .00}$ & $.61 \mathsmaller{\pm .05}$ & $.48 \mathsmaller{\pm .05}$ & $.30 \mathsmaller{\pm .05}$ & $.15 \mathsmaller{\pm .04}$ \\
CogVideoX-5B  & $.24 \mathsmaller{\pm .04}$ & $.10 \mathsmaller{\pm .03}$ & $.00 \mathsmaller{\pm .00}$ & $.00 \mathsmaller{\pm .00}$ & $.60 \mathsmaller{\pm .05}$ & $.45 \mathsmaller{\pm .05}$ & $.28 \mathsmaller{\pm .05}$ & $.14 \mathsmaller{\pm .04}$ \\
CogVideoX-2B  & $.32 \mathsmaller{\pm .05}$ & $.17 \mathsmaller{\pm .04}$ & $.06 \mathsmaller{\pm .02}$ & $.04 \mathsmaller{\pm .02}$ & $.59 \mathsmaller{\pm .05}$ & $.54 \mathsmaller{\pm .05}$ & $.34 \mathsmaller{\pm .05}$ & $.23 \mathsmaller{\pm .05}$ \\
VideoCrafter2 & $.18 \mathsmaller{\pm .04}$ & $.07 \mathsmaller{\pm .03}$ & $.01 \mathsmaller{\pm .02}$ 
& $.00 \mathsmaller{\pm .01}$ & $.59 \mathsmaller{\pm .05}$ & $.52 \mathsmaller{\pm .05}$ & $.36 \mathsmaller{\pm .05}$ & $.23 \mathsmaller{\pm .05}$ \\
Pyramid Flow   & $.23 \mathsmaller{\pm .04}$ & $.07 \mathsmaller{\pm .02}$ & $.00 \mathsmaller{\pm .00}$ & $.00 \mathsmaller{\pm .00}$ & $.63 \mathsmaller{\pm .04}$ & $.45 \mathsmaller{\pm .05}$ & $.32 \mathsmaller{\pm .05}$ & $.21 \mathsmaller{\pm .05}$ \\
HunyuanVideo  & $.30 \mathsmaller{\pm .05}$ & $.08 \mathsmaller{\pm .03}$ & $.03 \mathsmaller{\pm .02}$ & $.00 \mathsmaller{\pm .00}$ & $.56 \mathsmaller{\pm .04}$ & $.45 \mathsmaller{\pm .05}$ & $.30 \mathsmaller{\pm .05}$ & $.18 \mathsmaller{\pm .05}$ \\
\midrule
Pika  & $.27 \mathsmaller{\pm .05}$ & $.06 \mathsmaller{\pm .02}$ & $.00 \mathsmaller{\pm .00}$ & $.00 \mathsmaller{\pm .00}$ & $.66 \mathsmaller{\pm .05}$ & $.57 \mathsmaller{\pm .05}$ & $.35 \mathsmaller{\pm .05}$ & $.26 \mathsmaller{\pm .05}$ \\
Hailuo & $.28 \mathsmaller{\pm .05}$ & $.15 \mathsmaller{\pm .04}$ & $.05 \mathsmaller{\pm .02}$ & $.00 \mathsmaller{\pm .00}$ & $.66 \mathsmaller{\pm .05}$ & $.53 \mathsmaller{\pm .05}$ & $.35 \mathsmaller{\pm .06}$ & $.17 \mathsmaller{\pm .05}$ \\
Gen-3 Alpha  & $.26 \mathsmaller{\pm .05}$ & $.15 \mathsmaller{\pm .04}$ & $.03 \mathsmaller{\pm .02}$ & $.00 
\mathsmaller{\pm .00}$ & $.63 \mathsmaller{\pm .05}$ & $.51 \mathsmaller{\pm .05}$ & $.39 \mathsmaller{\pm .05}$ & $.23 \mathsmaller{\pm .05}$ \\
Kling  & $.23 \mathsmaller{\pm .04}$ & $.11 \mathsmaller{\pm .03}$ & $.05 \mathsmaller{\pm .02}$ & $.01 \mathsmaller{\pm .01}$ & $.60 \mathsmaller{\pm .04}$ & $.45 \mathsmaller{\pm .05}$ & $.29 \mathsmaller{\pm .06}$ & 
$.20 \mathsmaller{\pm .04}$ \\
\bottomrule
\end{tabular}
\end{table*}
\begin{table*}[htbp]
\centering

\caption{VACT benchmark on prevailing VGMs on human-sourced causal systems.}

\label{tab:human_benchmark}
\begin{tabular}{lc *{3}{cc}} 
\toprule
\multirow{2}{*}{\textbf{Model Names}} & \multirow{2}{*}{\textbf{N/A ratio}} & \multicolumn{2}{c}{\textbf{Text Consistency} $\uparrow$} & \multicolumn{2}{c}{\textbf{Generation Consistency} $\downarrow$} & \multicolumn{2}{c}{\textbf{Rule Consistency} $\uparrow$} \\
\cmidrule(lr){3-4} \cmidrule(lr){5-6} \cmidrule(lr){7-8}  
& & all & root & truth & observe & truth & observe \\
 \midrule
CogVideoX1.5-5B & .11 & $.58\mathsmaller{\pm .01}$ & $.58\mathsmaller{\pm .02}$ & $.09\mathsmaller{\pm .01}$ & $.08\mathsmaller{\pm .01}$ & $.54\mathsmaller{\pm .02}$ & $.69\mathsmaller{\pm .02}$\\
\midrule
Pika & .18 & $.57\mathsmaller{\pm .01}$ & $.55\mathsmaller{\pm .02}$ & $.07\mathsmaller{\pm .01}$ & $.06\mathsmaller{\pm .01}$ & $.54\mathsmaller{\pm .02}$ & $.67\mathsmaller{\pm .02}$ \\
Hailuo & .14 & $.63\mathsmaller{\pm .01}$ & $.62\mathsmaller{\pm .02}$ & $.07\mathsmaller{\pm .01}$ & $.08\mathsmaller{\pm .01}$ & $.55\mathsmaller{\pm .01}$ & $.70\mathsmaller{\pm .02}$ \\
\bottomrule
\end{tabular}
\end{table*}

In Table~\ref{tab:main_benchmark}, we observe that although certain closed-source models, such as Kling, demonstrated slightly better text consistency, and others, like Pika, showed marginally improved rule consistency, these differences were minor and not statistically significant. The fact that none of the existing models performed satisfactorily suggests that our benchmark can serve as a long-term objective for future research in this area. Given these observations, our main focus is to provide a comprehensive analysis and evaluation of the models. We explain the results as follows:

\paragraph{Text consistency}
The observed text accuracy ranged from 55\% to 65\%, only slightly above random guessing (50\%). This indicates that existing models \textbf{struggle to accurately generate variables from provided text}, regardless of whether causal or outcome variables are considered (with no significant difference between all and root cases). Although text fidelity has recently improved~\citep{sun2024soraseesurveytexttovideo}, our tests specifically require models to handle multiple variables simultaneously, including cases corresponding to less common scenarios. The low scores in our benchmarks highlight the models' difficulty in handling complex properties and uncommon situations. This implies that current models remain strongly \textit{limited to common scenarios} and \textit{lack the generalization capability} needed to effectively combine independent variables in more varied contexts, which is a necessary capacity for world simulators.

\paragraph{Generation consistency}
The generation consistency scores of all models are generally around 0.1. Consider that there is a random variable affecting the generation in each generation, we can use a two-point distribution to estimate the randomness. Because the variance of two-point distribution is $p(1-p)$, a value around 0.1 corresponds to approximately a 12\% deviation rate, meaning that 10 roughly 12\% of cases, model outputs vary significantly due to factors such as phrasing or random seed.

Overall, this indicates that models have learned relatively stable ``rule'', consistently producing similar outcomes for identical input variables $\mathbf{X}$.
However, this stability is not necessarily indicative of positive performance.
To better understand this, we consider text consistency and generation consistency results together. Although around 40\% of root variables being generated incorrectly (corresponding to around 60\% text consistency), the ``truth'' score and ``observation'' score in generation consistency remain closely aligned. This alignment occurs despite the fact that these scores categorize the same set of samples according to different criteria: the ground truth inputs $\mathbf{X}$ and the observed generated inputs $\hat{\mathbf{X}}$, respectively and there is a significant difference between these two values.

Such a pattern suggests that models may be \textbf{producing fixed outcomes} $\mathbf{Y}$ \textbf{regardless of variations in input variables} $\mathbf{X}$. This type of stability reflects what we call a ``\textit{degenerative}'' rule --- akin to a constant function. An illustrative example is shown in Figure~\ref{fig:bad_case}, where any object entering water invariably generates a splash. This behavior was further validated through manual inspection, as detailed in Appendix~\ref{app:analyze_gen_stable}.

\paragraph{Rule consistency} 
Finally, we directly evaluate the correctness of the rules learned by the models. The average ``truth'' rule accuracies are below 60\%, suggesting that the end-to-end rule accuracies are only slightly better than random guessing. Eliminating the influence of inaccurate conditions, the ``observe'' score shows that only around 70\% samples produces the same outcomes as in the real world. These findings indicate that poor performance in video generation consistent with real-world causal relationships is driven by both of these two main factors: models' misunderstanding of causal rules and their inability to accurately interpret and follow text instructions.

% Finally, we directly assess the correctness of the rules learned by the models. The average rule accuracies are below 60\% for the truth and only around 70\% for observation, with random guessing corresponding to 50\%. It suggests that both the misunderstanding of causal rules and the lacking to follow the correct text instruction leads to bad performance to generate videos obey the real-world causal relationships.

We calculate the threshold scores with threshold as $0.65$, $0.75$, $0.85$ and $0.95$. Here, we first calculate the average accuracy among samples for each causal rule and label it as $1$ if the average accuracy is larger than the threshold otherwise 0. The threshold score can reflect how many rules have been learned by VGMs with certain consistency. The results in Table~\ref{tab:metric_level_3_threshold} shows that only 14\%-26\% of the rules can match the groundtruth rules in 95\% of generation. This means that only a small number of causal relationships are learned by the model with a high degree of certainty.
And nearly 30\% of the rules have an accuracy below 65\%, which are almost unappreciated by the models. These findings clearly indicate that the models have not correctly understood the relationship between outcomes and causes (or parents), revealing weak rule learning of the current models.

\paragraph{Human source evaluation}
We also benchmark some models using the 60 human-annotated causal systems (\ie the control group in crowd experiments) as an ablation study to support our automatic test case generation.
Except for the factors and rules in the test cases, other steps follow the same as the VACT system. The results are shown in Table~\ref{tab:human_benchmark}, where we conducted experiments using three video generation models: CogVideoX1.5-5B, Hailuo, and Pika. We obtain similar results. This serves an evidence for both the effectiveness of our automatic annotation and the validity of our benchmark conclusions. Other results like N/A ratio and threshold score are shown in Appendix~\ref{app:human-source}.

\section{Conclusion}

In this paper, we propose an automated system for modeling causal relationships in scenarios and evaluating the causal behavior of VGMs. By combining LLM's commonsense understanding with intervention experiments, our automatic system can assess the causal learning in VGMs across diverse domains, scenarios, and rules without any human annotation. We validated its effectiveness through crowd experiments and manual checks. We introduced three progressive causal metrics to comprehensively analyze the models' causal behavior. Using this system, we created a benchmark and identified key causal flaws in existing models. As a long-term target, this work lays the foundation for large-scale detection of shortcut or biased learning, supplementing comprehensive training datasets, or reinforcement learning.

\section{Limitation}
We acknowledge several limitations in our current work. First, although the LLMs can already generate high-quality testbeds, occasional errors may still occur. For scenarios requiring extremely high quality assurance, human check and assistance is still recommended. Second, our causal system construction involves certain simplifications, such as focusing only on visualized factors and binarizing variables, which may need refinement for more complex scenarios, such as extending binarization to multiple discrete levels. This remains our future work. Lastly, our evaluation assumes that model generate high-quality videos but some models still struggle with text understanding and coherent video generation, hindering the analysis of their causal behavior. We view our system as a forward-looking tool, believing that as video generation models rapidly improve, causal behavior analysis will become more critical.

\section*{Acknowledgment}
\vspace{-5pt}
Z. Lin, M. Zhang and Y. He were supported by National Key R\&D Program of China (2022ZD0160300). Z. Lin was supported by the NSF China (No. 62276004). This work was supported by Kunpeng \& Ascend Center of Excellence, Peking University.

{
    \small
    \bibliographystyle{ieeenat_fullname}
    \bibliography{main}
}


\clearpage
\setcounter{page}{1}
\maketitlesupplementary
\appendix

\let\oldsection\section
\renewcommand{\section}[1]{%
  \oldsection{#1}%
  \addcontentsline{app}{section}{\protect\numberline{\thesection}#1}%
}

\let\oldsubsection\subsection
\renewcommand{\subsection}[1]{%
  \oldsubsection{#1}%
  \addcontentsline{app}{subsection}{\protect\numberline{\thesubsection}#1}%
}

\listofappendices
\newpage
\section{The ``stone'' and ``feather'' example}
\label{app:more_examples}

\begin{figure*}[ht]
    \centering
    \begin{subfigure}{1\linewidth}
    \includegraphics[width=1\linewidth]{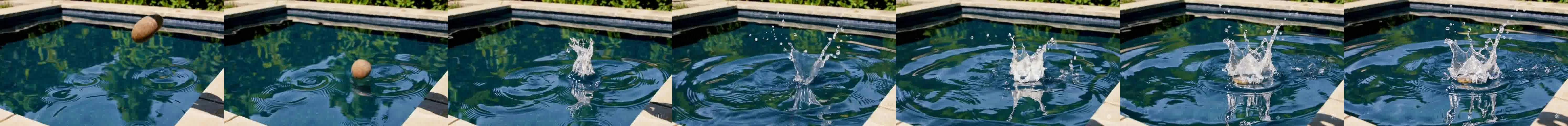}
    \includegraphics[width=1\linewidth]{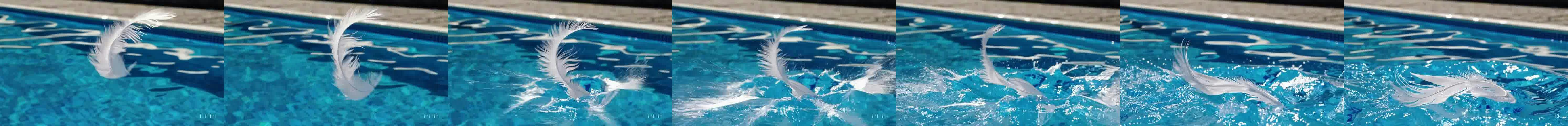}
    \caption{OpenAI Sora Generation}
    \end{subfigure}
    
    \begin{subfigure}{1\linewidth}
    \includegraphics[width=1\linewidth]{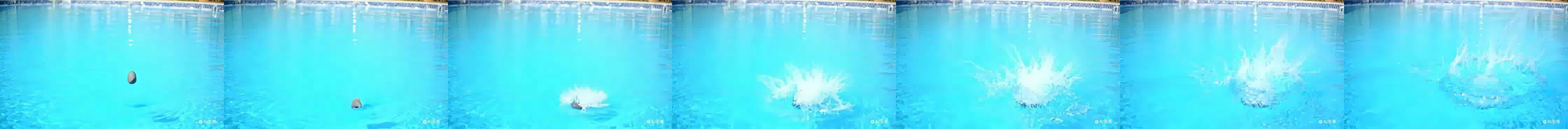}
    \includegraphics[width=1\linewidth]{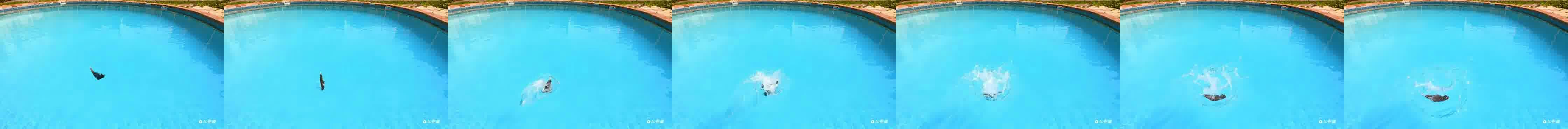}
    \caption{CogVideoX-2 Generation}
    \end{subfigure}

    \begin{subfigure}{1\linewidth}
    \includegraphics[width=1\linewidth]{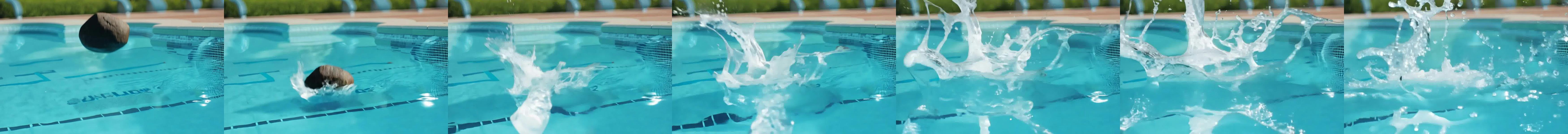}
    \includegraphics[width=1\linewidth]{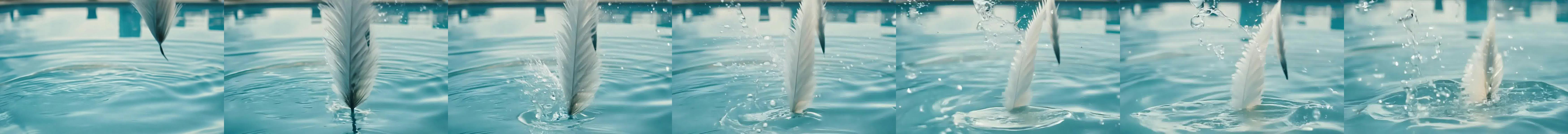}
    \caption{Gen-3 Alpha Generation}
    \end{subfigure}
    
    \caption{Videos generated by (a) OpenAI Sora, (b) CogVideoX-2 and (c) Gen-3 Alpha, shown as frames. For each model, the text prompt of the \textbf{Above} is: \textit{a stone is thrown into a swimming pool}; \textbf{Below} is: \textit{a feather is thrown into a swimming pool}. Both generation show \textit{noticeable splashes}, which is correct for the above (stone) scene but \textbf{incorrect} for the \textbf{below} (feather) scene.}
    \label{fig:more_bad_case}
\end{figure*}

%In figure~\ref{fig:bad_case}, we show that the OpenAI Sora~\citep{sora_is_here} cannot distinguish the different effects of a stone and a feather falling into water. 
In Figure~\ref{fig:bad_case}, we demonstrate that OpenAI’s Sora~\citep{sora_is_here} fails to distinguish between the different effects of a stone and a feather falling into water. This is not an isolated case of Sora. In figure~\ref{fig:more_bad_case}, we show the generation of CogVideoX-2~\citep{hong2022cogvideolargescalepretrainingtexttovideo} and Runway Gen-3 Alpha~\citep{runwaygen3}, showing that this spurious correlation is a common phenomenon that may exist in various models. 
%The models behave as if they ``directly'' replace the stone with the feather, without realizing that it changes many of the outcomes.
These models seem to ``directly'' substitute the stone with the feather, without understanding the significant differences in the outcomes.

On the one hand, this spurious correlation, we believe, comes from the distribution of the data set. 
%We found that there are a large number of videos on the Internet of throwing stones into the water, and there are significantly fewer videos of throwing feathers into the water.
We found that videos of stones being thrown into water are abundant online, while videos of feathers being thrown into water are significantly less common. 
 %As a side evidence, when we search ``thrown stone into water video'', there are \num{180000000} results while when we replace the stone with feather, the number reduces to \num{31000000}.
As supporting evidence, a search for “\textit{thrown stone into water video}” returns approximately \num{180000000} results on google, while replacing “stone” with “feather” reduces the results to around \num{31000000}.
This data bias means that the model may have seen enough scenes of stones entering the water during training but not enough scenes of feathers doing the same. 
%On the other hand, this also comes from the current widespread overfitting of the VGM models, which causes the model to rely too much on common data in the data set without sufficient understanding of scene rules;
Additionally, this issue stems from the widespread overfitting of current VGM models, which causes them to rely heavily on common data in the dataset without fully understanding the underlying rules of the scene.

In contrast, the current \textit{language} models like GPT-4o can more effectively grasp the different outcomes caused by various objects falling into the water. 
%can fully understand the different results caused by different objects falling into the water. 
In this case, the language model can distinguish that feathers falling into the water will not cause splashes. See the following chat (where some excerpts have been made for ease of reading):

\begin{promptbox}

\textbf{Question}: Please detailed describe the scenario about ``a stone is thrown into a swimming pool''

\textbf{Answer by GPT-4o}: 

\dots

The stone experiences air resistance as it moves, but due to its density, air drag has little effect.

\ 

As the stone contacts the water, it displaces the liquid, creating an initial splash.

\ 

Water droplets are ejected in different directions, forming a dynamic spray pattern.

\ 

The surrounding water rushes in to fill the space left by the stone, leading to the formation of ripples.

\ 

The stone rapidly sinks due to its density being greater than water.

\ 

The initial impact generates concentric ripples that spread outward across the water’s surface. If the throw was forceful, larger waves might be created, bouncing off the pool walls.

\ 

Over time, these waves dissipate due to friction and energy loss.

\end{promptbox}

\begin{promptbox}
\textbf{Question}: Please detailed describe the scenario about ``a feather is thrown into a swimming pool''

\textbf{Answer by GPT-4o}: 

\dots

As it falls, it flutters down slowly, almost like it's floating through the air. 

\ 

It lands gently on the surface of the water without much impact. The water ripples slightly where the feather touches, but there’s no big splash.

\

The feather floats on the surface, moving slowly with the tiny waves or currents in the pool. It stays on the water, shifting around a little as the surface moves, but it doesn’t sink or get pushed underwater.

\end{promptbox}

It is obvious that current language models have clear and comprehensive understanding of the scenario and even the physical mechanism in the scenario. It is the reason why we believe the LLMs can be utilized to automatically generate the groundtruth causal rules of test cases.

\section{Details of automatic generation of causal systems}
\label{app:automatic_generation_detail}
\subsection{Details of generating process}

We use the official API of OpenAI o1 model (o1-2024-12-17)~\citep{o1} to generate the causal systems. The three tasks are divided and prompted sequentially, with the LLM completing them through multiple rounds of dialogue. Throughout this process, the entire dialogue history is retained within the context window.
%Here, we split the three tasks, and prompt the LLM to complete these tasks in sequence through multiple rounds of dialogue. During this process, the complete dialogue history is always saved in the context window.
%The model will move to the next task until the maximum number of attempts is reached or the external check are passed and the LLM keep its answer after self-check.
The model will proceed to the next task either once the maximum number of attempts is reached or when the external checks are passed and the LLM retains its answer after a self-check.

%We require each step's generated content to include a file containing specific information, where:
We require that the generated content for each step includes a file containing specific information, where:
\begin{itemize}
    \item \textbf{Factor analysis}: a json file as a list of dictionary containing:
    \begin{itemize}
    \item ``\texttt{type}'': choices from ``factor'' or ``result''.
    \item ``\texttt{name}'': the name of the factor or result variable. They could be some words or a short sentence that can summarize the key meaning.
    \item ``\texttt{explanation}'': A short explanation about how the factor or result can affect the scenario and why the variable is visible, binary and important.
    \end{itemize}

    \item \textbf{Causal Graph}: a dot file that constructs a digraph, which first declares each factor as a node, then declares some directed edges between nodes. We use the Python library ``NetworkX''~\citep{SciPyProceedings_11} to verify the format.

    \item \textbf{Causal System}: a json file as a list of dictionary containing:
    \begin{itemize}
        \item ``\texttt{scenario}'': a string describing the event,
        \item ``\texttt{roots}'': a list of strings, each of which is a name of cause variable,
        \item ``\texttt{non\_roots}'': a list of strings, each of which is a name of outcome variable,
        \item ``\texttt{rules}'': a dictionary where each outcome variable corresponds to a Boolean function of its parents in the causal graph. The boolean function should be expressed as a disjunctive normal form (DNF), where each conjunctive clause are expressed as a dictionary ($A\land B\land \neg C$ expressed as \texttt{\{'A': True, 'B': True, 'C': False\}}. And the DNF is expressed as a list of the dictionary-expressed conjunctive clause.
    \end{itemize}
\end{itemize}

The complete generation process consumes roughly 20k reading tokens (10k cached) and 10k prediction tokens, costing about \$0.74 per causal system. This is approximately one-third the cost of manual labeling, which is 15 CNY per annotation.

\subsection{Requirement: rule-based \& self correction}
%We have some requirements for each internal results as well as final output causal systems. The concrete requirements can be found in the prompt in Appendix~\ref{app:full_prompts}. To make sure that the requirements should be met as much as possible, we design some check-and-correction loop. 
We have specific requirements for both the internal results and the final output causal systems. The detailed requirements can be found in the prompt in Appendix~\ref{app:full_prompts}. To ensure these requirements are met as thoroughly as possible, we have designed a check-and-correction loop.

Except for the first step ``factor analysis'', we use both the rule-based check and self-check for the answer generated by LLM. For the ``causal graph'', we check the following requirements by a Python program:
\begin{itemize}
    \item whether the generated answer consists of a legal dot file,
    \item whether the graph is a DAG, 
    \item whether there is an isolated node in the graph.
\end{itemize}
For the ``causal system'', we check that 
\begin{itemize}
    \item whether the returned rules keep the legal format, that is, it is a json file, with the correct keys (roots, non-roots, rules) and all values are in the correct format. Especially for the rules, we define a standard format to use a python list of dictionary to represent a disjunctive normal (DNF). We check whether the generated answer is a legal DNF.
    \item whether the rules leads to the same causal graph generated in the ``causal graph'' step,
    \item whether all the non-root nodes have exact one DNF and the root nodes do not have their DNF.
\end{itemize}
If any requirement has not been met, an error message will be the feedback to the LLM with the full history, and the LLM is required to regenerate its answer given the error message and the history information.

If the rule-based check has passed, we prompt the LLM to further check its answer by itself. 
The self-check prompts repeat the requirement in a more detailed way. These prompts are shown in Appendix~\ref{app:full_prompts}.

Although the current reasoning models like OpenAI o1 has learned to self-check during its thinking steps, we find the explicit self-check prompt can further help to improve the performance. For example, when asked to identify key factors in the scenario ``Knife slicing through (butter)'', o1 initially identifies ``\textit{Butter is cold and firm}'', where, while accurate, the temperature is not easily visible in the video. After a self-check process, o1 revises its answer to ``Butter is in block form'', a factor that can be more easily identified in the video. We believe that it could be because in this step, an LLM can think in more detail about whether the answer satisfies the condition without having to take into account the generation task at the same time.

Considering that we have adopted a step-by-step strategy, we also allow the model to regret the previous answer in the subsequent steps. For example, when generating causal rules, if the model finds that the previous causal graph is unreasonable during the process, we allow the model to generate \texttt{<regenerate\_graph>} to go back to the previous step. 
%We found that although such a situation is rare, it can effectively reduce the probability of the model generating low-quality answers.
While this situation is rare, we have found that it effectively reduces the likelihood of the model producing low-quality answers.

%We allow the model to generate the answer to \texttt{<keep\_answer>} after self-checking, if so, we will skip the subsequent checking steps. We found that a total of 3 checks can make almost all conditions satisfied where the model is almost satisfied with its answer and generates \texttt{<keep\_answer>}.
We allow the model to generate \texttt{<keep\_answer>} after self-checking. If this occurs, we skip the subsequent checking steps. We found that after a total of three checks, most conditions are met, and the model is typically satisfied with its answer, generating \texttt{<keep\_answer>}.
\subsection{Prompts}
\label{app:full_prompts}
In this section, we provide all of the prompts we use to facilitate the LLM to generate causal systems.

Prompt for Identifying Key Factors in a Scenario:
\begin{promptbox}
You will be provided with a brief description of a scenario. There could be some physical phenonmenon in this scenario. Please identify some **important** and **common** potential factors whose changes could significantly influence some important outcome of the scenario. These factors can fall into one of the following categories:

1. The objects or their properties in the scenario.

2. The object in the environment or the properties of the environment.

3. The actions or some properties of the action. 

For each factor, ensure that it meets the following criteria:

1. It should be **visible** and easily recognizable in a video.

2. It should be **binary**, meaning it can be clearly labeled as either ``yes'' or ``no'', rather than a continuous value.

3. It should be **independent**, not dependent on other factors.

4. Its effect on the outcome should be **deterministic** (i.e., it directly leads to a certain result, rather than just increasing or decreasing the probability).

5. The resulting effect should also be **visible** in a video.

If there is a pair bracket in the description, it means the content in the bracket is expected to be a variable (factors or outcome). For example, ``A (large) stone is thrown into a swimming pool (and splash water).'' means we expect ``does the water splash'' as one of the outcome and whether the stone is large enough is expected as one of ``factors''. But notice that it does not mean that other factors or outcomes are not allowed, you can also propose other factors or outcomes.

Please organize your answer as a **json** file as a list of dict, where each dict is like \{ ``type'': ``factor\_or\_result'', ``name'': ``factor\_or\_result\_name'', ``explanation'': ``how it affects the scenario and why you believe it is important and common''\}. Start your answer with a \textlangle json\textrangle tag and end with a \textlangle /json\textrangle tag.
\end{promptbox}

Prompt for Causal Graph Construction:
\begin{promptbox}
Based on the factors you proposed and their expected results, generate a causal graph that summarizes the physical relationships between them. In the graph, include only the most important and common factors or results; omit any overly detailed or trivial ones.

The graph should be a **directed acyclic graph**, where: 

 - Each **node** represents a factor or a result. 

 - Each **edge** represents a direct causal relationship between two nodes. 

The graph should be formatted in **DOT** format. Begin the DOT file with a \textlangle dot\textrangle tag and end it with a \textlangle /dot\textrangle tag. 
\end{promptbox}

Prompt for Causal Rule Generation:
\begin{promptbox}
Given the causal graph you generated, please create a Boolean expression for each **non-root** factor (factors with incoming edges) that represents the conditions under which that factor is **true**. The Boolean expression for each non-root factor should involve only the **parent factors** (i.e., the factors directly connected to it in the causal graph). The condition should be expressed as a **disjunctive normal form** (DNF), which is a disjunction (OR) of conjunctions (AND) of literals. 

Your response should include a set of boolean expressions, formatted as a `dict[str, list[dict[str, bool]]]', where the key is the name of this non-root factor and the value is a list of conditions (disjunctions), where each condition is a conjunction clauses (AND). Each condition is represented as a dictionary, where the key is the name of the parent factor and the value is a boolean value (True or False).

For example, if a factor A is true when B is true or (C is true and D is false), the boolean expression should be `\{``A'': [\{``B'': True\}, \{``C'': True, ``D'': False\}]\}'.

Your final answer should be a JSON file with the following keys

- ``roots'': a list of root factors.

- ``non\_roots'': a list of non-root factors.

- ``rules'': a dictionary where each non-root factor is associated with its corresponding Boolean expression.

Please begin your response with a \textlangle json\textrangle tag and end with a \textlangle /json\textrangle tag.
\end{promptbox}

For self-check prompt for factors:
\begin{promptbox}
Please review the factors you have proposed.  Ensure that each factor satisfies the following 5 requirements:

1. It should be **visible** and easily recognizable in a video.

2. It should be **binary**, meaning it can be clearly labeled as either ``yes'' or ``no'', rather than a continuous value.

3. It should be **independent**, not dependent on other factors.

4. Its effect on the outcome should be **deterministic** (i.e., it directly leads to a certain result, rather than just increasing or decreasing the probability).

5. The resulting effect should also be **visible** in a video.

Please ensure that the content in the bracket has been correctly identified as a variable (factor or outcome) in your answer.

Additionally, filter out any factors that are:

- **Too detailed**, **corner-case**, or **uncommon** in the scenario.

- Have an effect that is **too indirect** or difficult to understand.

If necessary, you may regenerate the factors to meet the criteria. It's OK to keep your previous answer by just generate \textlangle keep\_factor\textrangle\ without any other words  but you should carefully check every requirement for every factor and result.
\end{promptbox}

For self-check prompt for graph:
\begin{promptbox}
Please review your causal graph. Ensure that it meets the following criteria:

1. All nodes are **visible** and **binary**.

2. All root nodes are **independent** of each other, which means the choice of one root node should not influence the choice of another root node.

3. All edges in the graph is a **direct** and **deterministic** causal relation

4. Include all **important** causes and results, while omitting trivial or overly detailed nodes.

Please ensure that the content in the bracket has been correctly identified as a variable (factor or outcome) in your answer.

If necessary, regenerate the causal graph to meet these requirements. It's OK to keep your previous graph if it already meets the criteria by just generate \textlangle keep\_graph\textrangle\ without any other words but you should carefully check every requirement for every node and edge.
\end{promptbox}

For self-check prompt for rules:
\begin{promptbox}
Please review your answer. Ensure your answer meets the following criteria:

1. The ``roots'' and ``non\_roots'' list must be consistent with the causal graph. 

2. For the bool expressions: 

- All the nonroot factors are included in the rules dict, and no other factors are mistakenly included as keys.

- All variables in the Boolean expressions are exactly the parents of the corresponding non-root factors in the causal graph.

- The boolean expressions should correctly represent the physical rules in the real world.

If necessary, regenerate the json file to meet the requirements. It's OK to keep your previous rules if they already meet the criteria by just generate \textlangle keep\_rule\_json\textrangle\ without any other words but you should carefully check every requirement for every variable and rule.

If you find that you need to modify your generated causal graph, please generate \textlangle regenerate\_graph\textrangle \textlangle dot\textrangle ... \textlangle /dot\textrangle where the content between \textlangle dot\textrangle and \textlangle /dot\textrangle is the new causal graph.
\end{promptbox}

\section{20 scenarios in crowd experiments and benchmark}
\label{app:scenario_list}
%The 20 scenarios we use to operate our crowd experiments as well as our benchmark are listed as follows. These scenarios vary in the types of relationships involved, their complexity, and the extent to which they may contain variables. To simulate the situation where users may already have individual variables of interest, we also designed a ``bracket'' representation to prompt LLM that the content enclosed in the brackets MUST be considered as a variable.  Notice that we do not include the ``stone into water'' scenario into the list because we leave it as our debug case to adjust the prompt and provide an example for human annotators. 
The 20 scenarios used in our crowd experiments and benchmarks are listed below. These scenarios vary in the types of relationships they involve, their complexity, and the extent to which they include variables. To simulate situations where users may already have specific variables of interest, we also designed a ``bracket'' representation to prompt the LLM, indicating that the content within the brackets MUST be treated as a variable. Note that the "stone into water" scenario is not included in the list, as it serves as our debug case for adjusting the prompt and providing an example for human annotators.
\begin{enumerate}
    \item A small ball impacts the ground.
    \item A bullet is shot towards an object.
    \item A hand squeezes a sponge.
    \item A burning ball of paper was thrown into a pile of paper.
    \item A burning candle is placed with (wind and rain).
    \item A person strikes an ice block with a hammer.
    \item Sunlight shines on the water surface, (creating sparkling reflections).
    \item Two children of (different weights) are sitting on a seesaw.
    \item Pour one liquid into another.
    \item Rubber eraser rubs off (pencil) marks on paper.
    \item Knife slicing through (butter).
    \item Swinging a bat to hit a ball.
    \item A boot stomps into a puddle of mud.
    \item A ray of light is shining on a wooden block.
    \item Flag waving (in the wind) at the top of pole.
    \item A broom drags across the (dirty) ceramic floor.
    \item After being released, the ball rolls down the slope on its own.
    \item A paper airplane is thrown and glides through the air.
    \item Drop dye into the water.
    \item Sprinkle (iron) filings around a magnet.
\end{enumerate}

We also show some LLM-generated examples of various relationships between variables on the above 20 scenarios. These examples illustrate the diversity and effectiveness of automatic generation. 

In the scenario ``A hand squeezes a sponge'', the LLM identifies key factors like ``Sponge is wet'', ``Hand applies strong grip'', and ``Hand fully releases the sponge''. It generates diverse relationships by considering the states of the objects, the actions, and their sequence. The model recognizes state-based relationships (e.g., wet sponge, strong grip), causal relationships (hand’s grip expels water), and temporal relationships (the sequence of squeezing and releasing). Additionally, it captures interaction relationships, where the sponge’s wetness and the hand’s pressure influence the outcome, such as ``Water is expelled''.

%In the scenario ``Pour one liquid into another'', the LLM identifies factors like ``Immiscible liquids'', ``Exceeds container capacity'', ``and "Poured liquid is carbonated''. It generates relationships based on the properties of the liquids (e.g., immiscibility and carbonation), the container’s capacity, and their interactions. The model understands how these factors lead to outcomes such as ``Layered separation'', ``Overflow'', and ``Foam formation'', integrating both the liquids' characteristics and the container's limitations.
In the scenario  ``After being released, the ball rolls down the slope on its own'', the LLM identifies factors such as ``Is ball on slope'', ``Is slope steep enough'', and ``Is path clear of obstacles''. The model links the position of the ball and the steepness of the slope to the ball’s ability to roll, understanding that the ball will move if both conditions are satisfied. It also incorporates the influence of obstacles, recognizing that any obstruction along the path can prevent the ball from reaching the bottom. The LLM successfully identifies the relationship between the final outcome, "Ball reaches bottom," and the various factors involved, while considering the entire process, including the potential for obstacles to interrupt the ball’s descent.

In the scenario ``Rubber eraser rubs off (pencil) marks on paper'', the LLM identifies factors like ``Is pencil mark'', ``Eraser in contact'', and ``Rubbing motion present''. These factors work together to determine the outcome, ``Pencil mark removed''. The model recognizes that the presence of a pencil mark and the eraser’s contact are necessary for the process to start.  Additionally,  the rubbing motion, combined with the eraser’s pressure, results in the final outcome of mark removal.

\section{Details of crowd experiment}
\label{app:crowd_exp}
%We operate a crowd experiment to valid our automatic annotation of causal system from a scenario description. 
We conducted a crowd experiment to validate our automatic annotation of causal systems based on scenario descriptions.
We first invite three undergraduates (2 from physics school and 1 from computer science school) to annotate the same 20 text scenarios. We provide them with the same requirements as we provided to LLM. We first check their annotation with first 5 attempts and then feedback some obvious misalignment with our requirement. 
%We additionally ask annotators not to refer to (1) textbooks, because we want annotators to use commonsense when annotating rather than introduce overly professional background knowledge (2) LLM or other automatic annotation tools to faithfully reflect human intuition (3) avoid mutual communication that may cause bias. 
We also instructed the annotators to avoid (1) referencing textbooks, as we wanted them to rely on commonsense rather than professional background knowledge, (2) using LLMs or other automatic annotation tools, to ensure their annotations reflected human intuition, and (3) communicating with each other to prevent bias.
For human annotators, we prompted them to think in three steps similar to LLM; but we only collected the final rules. In order to ensure the seriousness of the annotators, we took a small number of samples and asked the annotators to explain their annotation reasons, which were checked by the authors. For the purpose of real comparison, we allowed a small number of non-systematic errors or deviations in the annotations --- because this reflects the true level of human annotators.

These 60 annotations collected for the 20 scenario will be randomly shuffled together with the 60 annotations generated by LLM and given to five other annotators for scoring. The five annotators were also undergraduates (3 from computer science, 1 from mathematics, and 1 from economics).

The scoring standard we provide is:
\begin{itemize}
    \item \textbf{Requirement}: whether the annotation meets all of our requirements including visibility, binary, and root node independence.
    \item \textbf{Rationality}: whether all the nodes in the causal system are consistent with public knowledge and common; and whether the most important factors and causal relations are included in the annotation.
    \item \textbf{Soundness}: whether all the rules in the causal graph are correct and definitive (from both physics and commonsense).
\end{itemize}
Each criterion is scored on a scale of 1-4, where 
\begin{itemize}
    \item 4: the annotation is completely correct (or meet the requirement),
    \item 3: there are minor errors,
    \item 2: there are obvious errors,
    \item 1: there are essential errors and the annotation needs to be rewritten.
\end{itemize} 

\begin{figure}
    \centering
    \begin{subfigure}{0.48\linewidth}
    \includegraphics[width=0.98\linewidth]{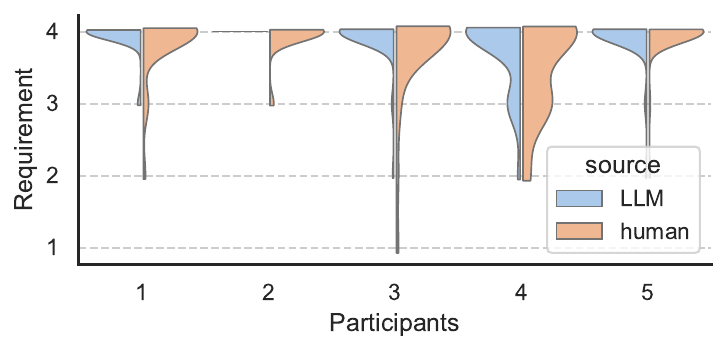}
    \caption{Requirement}
    
    \end{subfigure}
    \begin{subfigure}{0.48\linewidth}
    \includegraphics[width=0.98\linewidth]{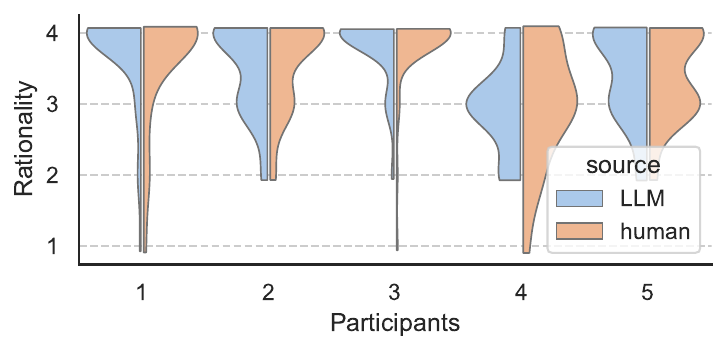}
    \caption{Rationality}
    
    \end{subfigure}

    \begin{subfigure}{0.48\linewidth}
    \includegraphics[width=0.98\linewidth]{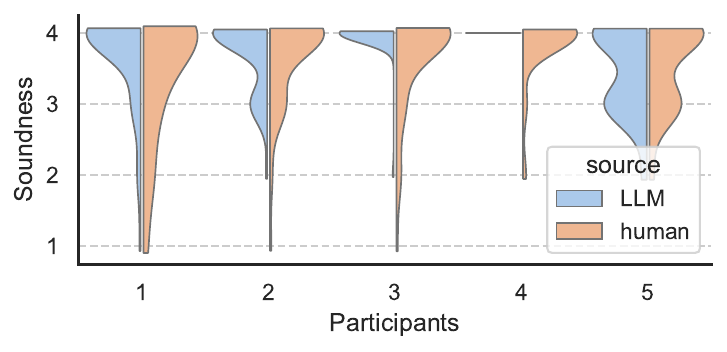}
    \caption{Soundness}
    \end{subfigure}
    \begin{subfigure}{0.48\linewidth}
    \includegraphics[width=0.98\linewidth]{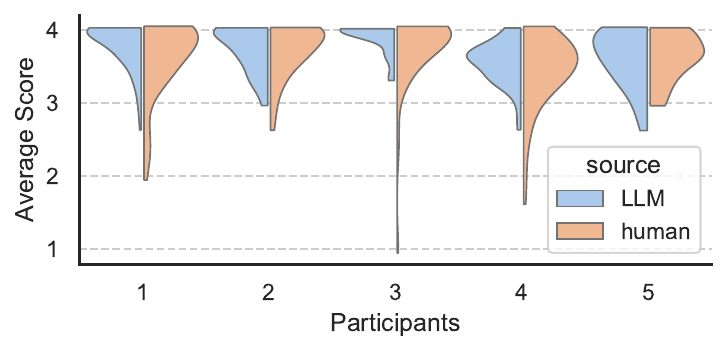}
    \caption{Average}
    \end{subfigure}
    \caption{The violin plot as detailed distribution of 5 scorers. The width shows the number of the samples. The x-axis represents the 5 annotators.}
    \label{fig:detail_crowd}
\end{figure}

The average scores have shown in Table~\ref{tab:crowd_average} in the main paper. Here, we provide the detailed distribution of each scorer in Figure~\ref{fig:detail_crowd}.

For ``requirement'' and ``soundness'', the LLM achieve excellent performance with a larger proportion of scores clustering around the top rating of 4 and the average score is significantly higher than human annotations. For rationality, the LLM- and human-annotation can not be clearly distinguished. The overall tendency of the five raters was consistent. Surprisingly, scorer 2 and 4 gave full marks of 4 points to all 60 items of LLM in requirement and soundness respectively.

Several examples highlight the reasons for the superior performance of the LLM in certain areas. Regarding the Requirement scores, the explicit guidelines provided in the prompt ensured that the LLM annotations generally met the requirements, resulting in consistently high scores. In contrast, human annotators occasionally failed to adhere to these requirements, either due to imprecise expressions or inadvertent oversights. For instance, in the scenario ``Pour one liquid into another'', one human annotator included the nodes ``the densities of the liquids differ greatly'' and ``the chemical structures of the liquids are similar'', both of which are unobservable factors. The LLM, however, avoided such missteps.

In terms of Soundness, where we require that the rules in the causal graph be both correct and definitive, human annotations displayed considerable variability across different scenarios. Some annotations included many nodes and rules, while others were sparse. In cases where a larger number of rules were included, human annotators sometimes overcomplicated their annotations, which led to errors. For example, in the scenario ``A bullet is shot towards an object'', a human annotator included the rule (A bullet hole appeared on the back of the object)= (The bullet moves quickly) $\land$ (The object is hard). The increased complexity of the rule, while addressing multiple factors, led to inaccuracies. The LLM, by contrast, considered fewer factors and produced simpler, more accurate rules.

For the Rationality criterion, which required the inclusion of the most important factors and causal relationships, human annotators excelled in some scenarios but failed to fully account for relevant factors in others. This variability resulted in a broader distribution of scores, with a greater number of high and low ratings for human annotations. Overall, the performance of both human and LLM annotations in this category was similar.

We took great care to ensure that all annotations generated in this experiment adhered to ethical guidelines, ensuring that no violent, pornographic, discriminatory, or offensive content was included in the annotated scenarios. We ensure that the collected data does not contain any personal privacy information. To safeguard against potential ethical violations, we closely monitored the content throughout the annotation process and implemented a strict review mechanism. Additionally, all annotators were explicitly instructed on the importance of maintaining a respectful and non-harmful approach in their work.

In recognition of the effort and time invested by the annotators, they were compensated at a rate of 100 CNY per hour, which is in line with standard industry practices for similar tasks. This compensation not only reflects the value of their contributions but also ensures that the annotators were fairly incentivized for their participation in the study. Furthermore, we provided a feedback loop for annotators, encouraging them to express any concerns or challenges they faced during the annotation process, fostering an open and transparent working environment.
\section{Details of test pipeline}
\label{app:test_pipeline}
Here we introduce the details of the test pipeline. For step ``prompt generation'', see Appendix~\ref{app:prompt_generation}. For step answer retrieval, see Appendix~\ref{app:probe_generate} and Appendix~\ref{app:answer_retrieval}. 
\subsection{Details of text prompt generation}
\label{app:prompt_generation}
Given a causal system as a test case, we need to generate some text prompts, which constrain the variable values in the scenario and are used to prompt the VGMs to generate corresponding videos. (In other words, they are used as the input of the tested T2V models.)

The step can be automated by an LLM. In this paper, we utilize the OpenAI gpt-4o (gpt-4o-2024-08-06) to finish it. To reduce communication overhead, we adopt the strategy of generating first and then sampling from the generated sentences, which is slightly different from the one described in the pipeline. Specifically, we provide the LLM with the original sentence description of the scenario and the list of variables (roots and non-roots separately). 
We require the model to generate $m$ sentences for every $2^N$ possible combinations of $\mathbf{X}$ where $N=|\mathbf{X}|$. % (or $\mathbf{V}$ if for calculating ``all'' score of level one metric text consistency $s^{all}_1$.)
We find for most situation $N<5$, the strategy of generating all value combinations at once is effective and works better than generating one value at a time. We observed that the former allows the model to consciously distinguish different values of $\mathbf{X}$. For cases where $N$ is too large, we take the approach of generating one value of $\mathbf{X}$ at a time. In our experiments, we set $m=10$. In this setting, for each causal system, About 500 tokens are reading and 500 - 1000 tokens are generated by gpt-4o, costing about \$0.005.

The prompt we use in the step is shown as follows:

Prompt for sentence generation without results:
\begin{promptbox}
You are a helpful assistant to generate corresponding short description about a scenario given some conditions. You will be provided with a short sentence to describe a scenario as well as some factors (variables) in the scenario. You should generate some short sentences which are slightly different from the originial sentence and describe the situation where the scenario is the same but the corresponding variables take given different value from original situation.

The scenario is: \{scenario\}

In this scenario, there are some factors are considered as (binary) variables and you should generate new description to change the original scenario to meet the corresponding value.
    
Factors: \{str(factors)\}.

There are also some results variables which are the outcome of the above factors: \{non\_roots\}. The values of these variables should not be mentioned in the generated sentences.

Each variable can take value as ``yes'' or ``no'' independently so that there are 2**\{num\_factors\} = \{num\_comb\} compositions. You should generate \{num\_sent\} sentences for each yes/no composition for these variables.

Please make sure (1) each sentence meet and explicitly express the corresponding value of variables and (2) the generated sentences as diverse as possible. Notice that you can add, delete or modify some words in original description to get the new sentence.

Your answer should be following the schema provided. Here,

- factors: The names of provides variables.

- compositions: Samples for all compositions. It is a list (len = 2**\{num\_factors\} = \{num\_comb\}) where each element has two parameters:

value: a list of bool. One-to-one correspondence with the values or the variables in the factors list.

samples: a list contains the given number of generated sentences.

\end{promptbox}

Prompt for sentence generation with results:
\begin{promptbox}
You are a helpful assistant to generate corresponding short description about a scenario given some conditions. You will be provided with a short sentence to describe a scenario as well as some factors (variables) in the scenario. You should generate some short sentences which are slightly different from the originial sentence and describe the situation where the scenario is the same but the corresponding variables take given different value from original situation.

The scenario is: \{scenario\}

In this scenario, there are some factors are considered as (binary) variables and you should generate new description to change the original scenario to meet the corresponding value.
    
Factors: \{str(factors)\}.

There are also some results variables which are the outcome of the above factors with their expected value: \{non\_roots\}. In each possible composition of factor values, you should first induce the corresponding value of the results variables and then generate the sentences. 

In these sentences, please explicitly and clearly express the corresponding value of both the factors and the results variables in the generated sentences. The rules of the results: \{``\textbackslash n''.join(rules)\}

Each variable can take value as ``yes'' or ``no'' independently so that there are 2**\{num\_factors\} = \{num\_comb\} compositions. You should generate \{num\_sent\} sentences for each yes/no composition for these variables.

Please make sure (1) each sentence meet and explicitly express the corresponding value of variables and (2) the generated sentences as diverse as possible. Notice that you can add, delete or modify some words in original description to get the new sentence.

Your answer should be following the schema provided. Here,

- factors: The names of provided factor variables.

- results: The names of provided results variables.

- compositions: Samples for all compositions. It is a list (len = 2**\{num\_factors\} = \{num\_comb\}) where each element has three parameters:

value: a list of bool. One-to-one correspondence with the values or the variables in the factors list.
    
results: a list of bool. One-to-one correspondence with the values or the variables in the results list. Calculated by the given rules.
    
samples: a list contains the given number of generated sentences.

\end{promptbox}

\subsection{Details of probe question generation}
\label{app:probe_generate}
We utilize GPT-4o-mini-2024-07-18 to generate questions for each variable. In a single conversation, we provide a short description of the scenario along with the factors that should be focused on. We instruct the model to generate questions for all root and non-root factors simultaneously. The prompt we design requires the model to generate a simple yes-no question for each factor in the scenario, ensuring that the questions are directly focused on the specific factor without incorporating any assumptions or conditions related to other factors.

The prompt we use in this step is shown as follows:

Prompts for Probe Question Generation:
\begin{promptbox}
You are a helpful assistant to help generate some questions about some factors in a scenario. You will be provided with a short description of a scenario and some factors that should be focused on. You should generate **ONE** yes-no questions for **EACH** of the factors in the scenario. These questions will be used to asked a video language model to test the actual situation in a video about the scenario. Notice that your questions should be simple, clear and direct to the target factor, and should not contain any assumption or conditions about other factors.

The scenario is \{scenario\}.

The factors are: \{factors\}.
\end{promptbox}

\subsection{Details of answer retrieval}
\label{app:answer_retrieval}
%We experimented with two models to accomplish the task of answering questions based on video content: Gemini and OpenAI 4o. Gemini possesses direct video reading capabilities, with an internal mechanism that extracts one frame per second for processing. 
We tested two models to answer questions based on video content: Gemini and OpenAI 4o. Gemini has built-in video reading capabilities, extracting one frame per second for processing.
In contrast, OpenAI 4o can process multiple images, so we extract one frame every 10 frames from the video and provide these key frames to the model for question answering. Ultimately, we adopted OpenAI 4o as the primary model for our experiments due to its superior performance.

For each video, we need to ask multiple questions. 
%To prevent the model from relying on commonsense or context in answering questions, rather than strictly using the video content, we explored two distinct questioning strategies. 
To ensure that the model relies strictly on the video content rather than commonsense or context, we explored two distinct questioning strategies.
The first strategy involves asking one question at a time, ensuring the independence of each answer, though this approach incurs higher costs. The second strategy involves asking all the questions in a single round, within a single prompt. To avoid the model inferring subsequent questions based on prior answers or external commonsense, we topologically sort the nodes in the causal graph, ensuring that result variables are queried before cause variables. This method prevents the model from reasoning through previous answers when addressing subsequent questions. Additionally, we specify in the prompt that the model should answer based solely on the video.

For each question, we allow the model to respond with True, False, or N/A. Some videos suffer from lower generation quality, or fail to align with the textual descriptions, causing critical factors to be unobservable. 
%In such cases, where the video does not provide sufficient evidence to answer the question, we permit the model to respond with N/A.
 In these cases, when the video does not provide enough evidence to answer the question, we allow the model to respond with N/A.

The prompt we use in this step is shown as follows:

Prompt for Video Analysis and Question Answering:
\begin{promptbox}
You are a professional video analysis expert, specialized in answering questions based on video content. Please answer the following question based **strictly** on the video provided. Ensure that your response is based on the video itself, and not on your own guesses or general knowledge. 
    
You will be provide some yes/no questions related to the video. Your answer should be in ``true'', ``false'' or ``N/A''. Besides, you should provide a brief explanation or evidence for your answer.
    
You should answer ``N/A'' if:

1. The video quality is too low, or the content is too unclear to make any meaningful inference.

2. The content in the video is not continuous or complete. The temporal and spatial discontinuities in the video make it impossible to make reasonable predictions.

3. The question asks about something that cannot be observed or recognized in the video (e.g., an object, event, or action that is not present).

4. The video does not provide enough context or evidence to form a conclusion.

5. The answer is unclear or could be interpreted in multiple ways, leading to ambiguity.

6. The question asks about an action, and the necessary prior action (for example, the ball hitting the ground before it can bounce) is not observed. Without the prior action, it is impossible to determine if the subsequent event occurred. 

if you believe you can answer yes or no with a reasonable degree of confidence, you should not answer ``N/A''. Especially, if the question asks about whether something is present, or an event has occurred, and the videos shows that it is absent or has not occurred, you should answer ``false'' instead of ``N/A''. For these questions, you can answer ``N/A'' only if the video quality is too low to make a meaningful inference.
   
If the question asks about an object, and the object is not observed, answer ``false''. Do not answer ``N/A''.
    
For detect an action, you should refer to some continuous frames to make sure the action is happening, instead of just one frame.
    
In addition, you should judge each question as independently as possible, and do not answer another question based on the content of another question. In particular, the content of another question itself should not be used as the basis for answering the current question.
    
Based on the above guidelines, please answer the following questions:

``\textbackslash n''.join(\{questions\})

\end{promptbox}

\section{Detailed definition for metrics}
\label{app:detailed_def_for_metrics}
In this subsection, we give a detailed definition for our proposed metrics in Section~\ref{sec: three levels}. 

First we review the definitions and symbols. Let $\mathbf{V}$ be a set of variables representing all factors of interest in a causal system. Let $G$ a directed acyclic graph with node set $\mathbf{V}$ and edge set $\mathbf{E}$. For every $V_j\in \mathbf{V}$, let $pa(V_j)=\{V_k\in \mathbf{V}: V_k\to V_j\in \mathbf{E}\}$ be the set of nodes that has a directed edge pointing to $V_j$. Suppose there is a deterministic structural equation model over $\mathbf{V}$. That is, for every $V_j\in \mathbf{V}$ such that $pa(V_j)\neq \emptyset$, there exists a function $f_j$ such that $V_j=f_j(pa(V_j))$. Denote $\mathbf{X}=\{V_j\in\mathbf{V}: pa(V_j)=\emptyset\}$ and $\mathbf{Y}=\mathbf{V} \setminus \mathbf{X}$. We also write $\mathbf{X}=(X_1,X_2,\dots,X_{m_1})$ and $\mathbf{Y}=(Y_1,Y_2,\dots,Y_{m_2})$ as random vectors. Then $\mathbf{X}$ is called the set of root (or cause) variables, and $\mathbf{Y}$ is called the set of non-root (or outcome) variables. In structural equation $Y_j=f_j(pa(Y_j))$ for every $Y_j\in \mathbf{Y}$, the function $f_j$ is called the rule of $Y_j$. The structural equations can be equivalently represented as $\mathbf{Y}=f(\mathbf{X})$. Since the value of non-root variables is determined by root variables, we also write $Y_j=f_j'(\mathbf{X})$ for every $Y_j\in \mathbf{Y}$. Let $D(\mathbf{X})=\{1,0\}^{|\mathbf{X}|}$ denote the domain of $\mathbf{X}$, that is, the set of all possible values of $\mathbf{X}$.

In our pipeline, we use a large language model for generating prompt from the given causal system and specified variables, a video generation model for generating video from the prompt, and an multi-modale LLM for retrieving the value of variables from the video. For specified $\mathbf{X},\mathbf{Y}$, let $f_P(\mathbf{X},\mathbf{Y})$ denote the generated prompt under the given causal system, with specifying both $\mathbf{X}$ and $\mathbf{Y}$. Let $f_P(\mathbf{X})$ denote the generated prompt under the given causal system with only specifying only $\mathbf{X}$. Note that $f_P$ includes an independent error $\varepsilon_P$ implicitly, so it is not a deterministic function of $\mathbf{X}$ and $\mathbf{Y}$. For a prompt $P$, let $f_V(P)$ denote the video generated by video generation model with prompt $P$. Finally, let $\hat{\mathbf{X}}, \hat{\mathbf{Y}} = f_A(f_V(P))$ denote the \textbf{observation} of all variables from the generated video. For simplicity, we also write $\hat{\mathbf{X}},\hat{\mathbf{Y}}=f_{V}(P)$. In this situation, we also call $\mathbf{X}, \mathbf{Y}$ the \textbf{ground truth}. For the $i$-th sample, let $\mathbf{X}^{(i)},\mathbf{Y}^{(i)}$ denote the ground truth and $\hat{\mathbf{X}}^{(i)},\hat{\mathbf{Y}}^{(i)}$ denote the observation. For any $V\in \mathbf{V}, X\in \mathbf{X}$ and $Y\in \mathbf{Y},$ we use $V^{(i)},X^{(i)},Y^{(i)}$ or $\hat{V}^{(i)},\hat{X}^{(i)},\hat{Y}^{(i)}$ to denote the corresponding component of $\mathbf{X}^{(i)},\mathbf{Y}^{(i)}$ or $\hat{\mathbf{X}}^{(i)},\hat{\mathbf{Y}}^{(i)}$, just as we use $V,X,Y$ to denote the corresponding component of $\mathbf{X},\mathbf{Y}$. We also use $V^{(i)}_j$ to denote the component $V_j$ in vector $\mathbf{V}^{(i)}$. For variable $Y_j\in \mathbf{Y}$, we use $\hat{pa}(Y_j)$ to denote the observed value of $pa(Y_j)$.

\subsection{Text consistency}

For text consistency, let $\mathbf{X}^{(1)},\mathbf{X}^{(2)},\dots,\mathbf{X}^{(n_1)}$ be $n_1$ samples that are i.i.d. are uniform distributed over $D(\mathbf{X})=\{1,0\}^{|\mathbf{X}|}$. Let $\mathbf{Y}^{(i)}=f(\mathbf{X}^{(i)})$ for $i=1,2,\dots,n_1$. 

Since we have specified the value of every variable in the prompt, we expect that the value of every observed variable matches with its ground truth. However, due to the internal causal mechanism in the video generation model, the value of outcome variables in the video may be influenced by the value of root variables in the video. Therefore, we propose two versions of metric: $s_1^{\mathrm{all}}$ by comparing the observed value of all variables with their ground truth, and $s_1^{\mathrm{roots}}$ by comparing the observed value of only root variables with their ground truth. For $s_1^{\mathrm{roots}}$, we generate prompt $P^{(i)}=f_P(\mathbf{X}^{(i)})$ by specifying only root variables, and for $s_1^{\mathrm{all}}$, we generate prompt $P^{(i)}=f_P(\mathbf{X}^{(i)},\mathbf{Y}^{(i)})$ by specifying both $\mathbf{X}^{(i)}$ and $\mathbf{Y}^{(i)}$. Finally, we get observation $\hat{\mathbf{X}}^{(i)},\hat{\mathbf{Y}}^{(i)}=f_{VA}(P^{(i)})$ by generating video from prompts and asking questions from videos.

The metrics for text consistency is defined as:
\begin{equation}
s_1^{\mathrm{all}}=\frac{1}{n_1|\mathbf{V}|}\sum_{i=1}^{n_1}\sum_{V\in \mathbf{V}}\mathds{1}(V^{(i)}=\hat{V}^{(i)}),
\end{equation}
and
\begin{equation}
s_1^{\mathrm{roots}}=\frac{1}{n_1|\mathbf{X}|}\sum_{i=1}^{n_1}\sum_{X\in \mathbf{X}}\mathds{1}(X^{(i)}=\hat{X}^{(i)}),
\end{equation}
where $\mathds{1}(\cdot)$ denotes the indicator function.

\subsection{Generation consistency}
\label{appsubsec: generation consistency}

For generation consistency, we construct some groups of samples. Samples within the same group should have the same ground truth. Therefore, by comparing observations within the same group, we can test whether generations for the same ground truth are consistent.

Formally, let $\mathbf{x}^{(1)},\mathbf{x}^{(2)},\dots,\mathbf{x}^{(n_2)}$ be $n_2$ different values that are randomly selected from $D(\mathbf{X})=\{1,0\}^{|\mathbf{X}|}$. We construct $n_2$ groups, with $r$ samples in each group, that is, letting 
\begin{equation}
\begin{aligned}
&\mathbf{X}^{(1)}=\cdots=\mathbf{X}^{(r)}=\mathbf{x}^{(1)},\\
&\dots\\
&\mathbf{X}^{((n_2-1)r+1)}=\cdots=\mathbf{X}^{(n_2r)}=\mathbf{x}^{(n_2)}.
\end{aligned}
\end{equation}
For $i=1,2,\dots,n_2r,$ let $P^{(i)}=f_P(\mathbf{X}^{(i)})$ be the generated prompt and $\hat{\mathbf{X}}^{(i)},\hat{\mathbf{Y}}^{(i)}=f_{VA}(P^{(i)})$ be the observation. 

To measure the inconsistency of observations within a group, we propose two versions of metric: $s_2^{\mathrm{truth}}$ and $s_2^{\mathrm{observe}}$. For $s_2^{\mathrm{truth}}$, we assume that text consistency holds, that is, observation of root variables should remains the same within each group. Therefore, we compare all variables for each group. For $s_2^{\mathrm{observe}}$, we allow for observation of root variables to be different within each group. Relatively, we see the observed root variables as the truth understood by the video generation model. So we reconstruct the groups by partitioning the samples by $\hat{\mathbf{X}}^{(i)}$,
and compare the observed outcome variables within each group.

Formally, for an index set $\mathbf{S}\subseteq \{1,2,\dots,n_2r\}$ and variable $V\in \mathbf{V}$, denote $\bar{V}_{\mathbf{S}}=\frac{1}{|\mathbf{S}|}\sum_{i\in \mathbf{S}}\hat{V}^{(i)}$ be the mean, and $d(V,\mathbf{S})=\frac{1}{|\mathbf{S}|}\sum_{i\in \mathbf{S}}\left(\hat{V}^{(i)}-\bar{V}_{\mathbf{S}}\right)^2$ be the sample variance of $V$ in subgroup $\mathbf{S}$. For group index $k=1,2,\dots,n_2,$ let $\mathbf{S}_k=\{(k-1)r+1,(k-1)r+2,\dots,kr\}$ be the index of samples within group $k$. Then we have
\begin{equation}
s_2^{\mathrm{truth}}=\frac{1}{n_2|\mathbf{Y}|}\sum_{k=1}^{n_2}\sum_{Y\in\mathbf{Y}}d(Y,\mathbf{S}_k).
\end{equation}
For definition of $s_2^{\mathrm{observe}}$, for each $\mathbf{x}\in D(\mathbf{X})$, let $\mathbf{S}_\mathbf{x}=\{i:\hat{\mathbf{X}}^{(i)}=\mathbf{x}\}$, and let $\mathcal{S}=\{\mathbf{S}_{\mathbf{x}}\neq \emptyset:\mathbf{x}\in D(\mathbf{X})\}$.
Then we have
\begin{equation}
s_2^{\mathrm{observe}}=\frac{1}{|\mathbf{Y}||\mathcal{S}|}\sum_{Y\in \mathbf{Y}}\sum_{\mathbf{S}_\mathbf{x}\in \mathcal{S}}d(Y,\mathbf{S}_\mathbf{x}).
\end{equation}

\subsection{Rule consistency}
\label{app:subsec rule consistency}

For rule consistency, we generate samples for each outcome variable independently. For each $Y_j\in \mathbf{Y},$ let $\mathbf{S}^T_j=\{\mathbf{x}\in D(\mathbf{X}):f_j'(\mathbf{x})=1\}$ be the set of values of $\mathbf{X}$ that making $Y_j=f_j'(\mathbf{X})=1,$ and let $\mathbf{S}^F_j=D(\mathbf{X})\setminus \mathbf{S}^T_j$. Then for ground truth $\mathbf{X}$ and $\mathbf{Y}=f(\mathbf{X})$, we have $Y_j=1$ if and only if $\mathbf{X}\in \mathbf{S}^T_j$. 

To test whether the video generation model has learned this rule, we draw $n_3$ samples $\mathbf{X}^{(1)},\mathbf{X}^{(2)},\dots,\mathbf{X}^{(n_3)}$ uniformly from $\mathbf{S}^T_j$, and $n_3$ samples $\mathbf{X}^{(n_3+1)},\mathbf{X}^{(n_3+2)},\dots,\mathbf{X}^{(2n_3)}$ uniformly from $\mathbf{S}^F_j$. Comparing to drawing sample uniformly from $D(\mathbf{X})$, this sampling method avoids the bias that may arise when $|\mathbf{S}^T_j|/|\mathbf{S}^F_j|$ is near $0$ or $1$. For $i=1,2,\dots,2n_3,$ let $P^{(i)}=f_P(\mathbf{X}^{(i)})$ be the generated prompt and $\hat{\mathbf{X}}^{(i)},\hat{\mathbf{Y}}^{(i)}=f_{V}(P^{(i)})$ be the observation.

We also propose two versions of metrics for rule consistency, $s_3^{\mathrm{truth}}$ and $s_3^{\mathrm{observe}}$. For $s_3^{\mathrm{truth}}$, we assume that text consistency holds, and check whether the value of observed outcome variables matches its ground truth. For $s_3^{\mathrm{observe}}$, we see the observed parents of each outcome variabe as the truth understood by the video generation model. Therefore, we calculate the value of outcome variables from the rules and its observed parents, and compare them with observed outcome variables. Remind that $Y^{(i)}_j$ denotes the component $Y_j$ in vector $Y^{(i)}$. Formally, we have
\begin{equation}
    s_3^{\mathrm{truth}}(Y_j)=\frac{1}{2n_3}\sum_{i=1}^{2n_3}\mathds{1}\left(Y^{(i)}_j=\hat{Y}^{(i)}_j\right),
\end{equation}
and then
\begin{equation}
    s_3^{\mathrm{truth}}=\frac{1}{|\mathbf{Y}|}\sum_{Y_j\in \mathbf{Y}}s_3^{\mathrm{truth}}(Y_j).
\end{equation}
For $s_3^{\mathrm{observe}}$, we propose a strategy to rebalance samples such that the expected value of $Y_j$, $f_j(\hat{pa}(Y_j))$, has equal weights over $\{0,1\}$. Let $\hat{pa}^{(i)}(Y_j)$ denote the observed value of parents of $Y_j$ in the $i$-th sample $\mathbf{Y}^{(i)}$. Then, denote $g_j=\sum_{i=1}^{2n_3} f_j(\hat{pa}^{(i)}(Y_j))$ as the total number of samples such that the expected value of $Y_j$ is $1$, then we reweight each sample and define $s_3^{\mathrm{observe}}(Y_j)$ as
\begin{equation}
\begin{aligned}
    s_3^{\mathrm{observe}}(Y_j)& = \frac{1}{2} \sum_{i=1}^{2n_3} \mathds{1}\left( \hat{Y}^{(i)}_j=f_j(\hat{pa}^{(i)}(Y_j)) \right)\cdot\\
    &\left(\frac{f_j(\hat{pa}^{(i)}(Y_j))}{g_j} + \frac{1-f_j(\hat{pa}^{(i)}(Y_j))}{2n_3-g_j}\right),
    \end{aligned}
\end{equation}
and then
\begin{equation}
    s_3^{\mathrm{observe}}=\frac{1}{|\mathbf{Y}|}\sum_{Y_j\in\mathbf{Y}}s_3^{\mathrm{observe}}(Y_j)
\end{equation}

For revealing more intuition under the evaluation of rule consistency, we also define the threshold-based metrics for rule consistency. 
Let $t\in (0,1)$ denote the threshold, the threshold-based metrics are defined as
\begin{equation}
  s_3^{\mathrm{truth, threshold}}(t)=\frac{1}{|\mathbf{Y}|}\sum_{Y_j\in \mathbf{Y}}\mathds{1}\left(s_3^{\mathrm{truth}}(Y_j)\ge t\right),  
\end{equation}
\begin{equation}
    s_3^{\mathrm{observe, threshold}}(t)=\frac{1}{|\mathbf{Y}|}\sum_{Y_j\in \mathbf{Y}}\mathds{1}\left(s_3^{\mathrm{observe}}(Y_j)\ge t\right).
\end{equation}
These metrics measures the probability that for a given causal rule, the model gives a correct value for the outcome variable corresponding to this rule with an accuracy over the threshold.

\subsection{Sample strategy for three-level metrics}
\label{app:sample_strategy}
We propose a unified sampling framework designed to optimize sample efficiency across different evaluation metrics. %The methodology employs a centralized sample registry implemented through a hierarchical data structure. At its core, a master table serves as the primary sample pool, systematically recording all generated samples with unique identifiers. This master table acts as the canonical reference for experimental observations, ensuring a consistent foundation for downstream analyses.
First, we perform sampling for each metric. Specifically, for Metric 1: text consistency, we collect $n_1$ samples, where the $\mathbf{X}$ values are uniformly random from the set $D(\mathbf{X})=\{1,0\}^{|\mathbf{X}|}$. For Metric 2: generation consistency, we collect $n_2$ groups, each containing $r$ samples with the same $\mathbf{X}$ value. For Metric 3: rule consistency, for each $Y_j\in \mathbf{Y}$, we collect $n_3$ samples from the positive set $\mathbf{S}_j^T$ and the negative set $\mathbf{S}_j^F$, respectively. During each sampling step, we record the number of samples corresponding to different $\mathbf{X}$.

With the separate sampling results, we construct a total sample set, where for each possible $\mathbf{X}$ value, the sample count is the \textbf{maximum} across the three metrics.  While each sample may be used multiple times to compute different metrics or different rule accuracies for $Y_j$, within the same metrics (or within metric 3 for the same $Y_j$), each sample is used only once. The framework ensures that no sample is reused within the calculation of any single metric. By doing so, we maintain the independent and identically distributed (IID) conditions for sampling, while preserving the integrity of each metric’s evaluation criteria. The architecture also achieves significant storage efficiency, reducing redundancy compared to traditional independent sampling approaches, without compromising the statistical validity of the results.
Finally, we use the total sample set to select the corresponding text prompts and generate videos.

%To accommodate the specific requirements of different evaluation metrics, the framework utilizes dedicated subtables for each assessment dimension. For rule consistency evaluation, each outcome variable maintains a separate subtable containing only the indices of samples from the primary pool that meet its specific observational criteria. Similarly, for generation consistency assessment, a specialized subtable tracks both sample indices and their associated generation groups. These subtables are designed to operate independently, preserving the distinct data requirements of each metric without cross-contamination.

%A critical feature of this architecture is its enforcement of strict uniqueness constraints across all subtables. By implementing hash-based index validation, the framework ensures that no sample is reused within the calculation of any single metric. This design guarantees that the sampling process satisfies the independent and identically distributed (IID) conditions, while simultaneously maintaining the integrity of each metric's evaluation criteria. The architecture also achieves significant storage efficiency, reducing redundancy compared to traditional independent sampling approaches, without compromising the statistical validity of the results.

\begin{figure}[h]
    \centering
    \includegraphics[width=0.8\linewidth]{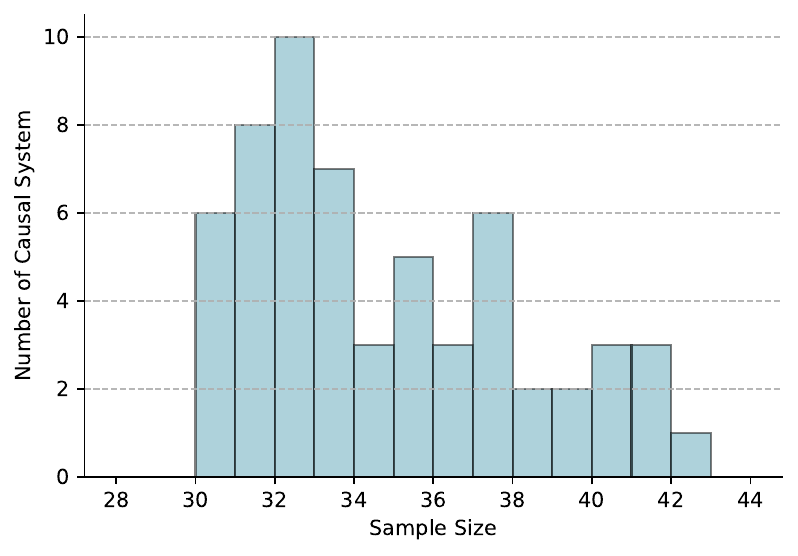}
    \caption{Distribution of sample sizes over causal systems.}
    \label{fig:sample_distribution}
\end{figure}

In our benchmark, we set the parameters as follows: $n_1=10$, $n_2=5$, $n_3=10$, and $r=3$. Using these values, we apply our strategy to draw samples. Appendix~\ref{app:sample_size_exp} demonstrate that this sample size is sufficient for distinguishing between metrics across different models. Specifically, we draw $n_1$ samples for the evaluation of text consistency, $n_2r$ samples for the evaluation of generation consistency, and $2n_3|\mathbf{Y}|$ samples for the evaluation of rule consistency. In contrast, without this strategy, a total of $N=25+20|\mathbf{Y}|$ samples would be required for each causal system, which could significantly increase computational costs. The distribution of sample sizes for each causal system is depicted in Figure~\ref{fig:sample_distribution}, which illustrates a considerable reduction in the number of samples needed by our approach.

\subsection{Sample-based scores}

Our metrics can also be applied to each sample, showing how each sample contributes to the evaluation. The definitions are as follows.

For text consistency, the metrics for a sample $i$ are defined as
\begin{equation}s_{1,i}^{\mathrm{all}}=\frac{1}{|\mathbf{V}|}\sum_{V\in\mathbf{V}}\mathds{1}(V^{(i)}=\hat{V}^{(i)}),
\end{equation}
and
\begin{equation}
s_{1,i}^{\mathrm{roots}}=\frac{1}{|\mathbf{X}|}\sum_{X\in\mathbf{X}}\mathds{1}(X^{(i)}=\hat{X}^{(i)}).
\end{equation}

For the sake of sample efficiency, samples with same ground truth $\mathbf{X}$ are reused in testing generation consistency and rule consistency. Let $i$ be the index of a sample in a group $\mathbf{S}$. Similarly, let $\bar{V}_\mathbf{S}=\frac{1}{|\mathbf{S}|}\sum_{i\in \mathbf{S}}\hat{V}^{(i)}$ be the mean of observed values for variable $V\in \mathbf{V}$. Then the metrics for generation consistency on the $i$-th sample are defined as
\begin{equation}
    s_{2,i}^{\mathrm{truth}}=\frac{1}{|\mathbf{Y}|}\sum_{Y\in\mathbf{Y}}(\hat{Y}^{(i)}-\bar{Y}_{\mathbf{S}_k})^2,
\end{equation}
and
\begin{equation}
     s_{2,i}^{\mathrm{observe}}=\frac{1}{|\mathbf{Y}|}\sum_{Y\in \mathbf{Y}}(\hat{Y}^{(i)}-\bar{Y}_{\mathbf{S}_\mathbf{x}})^2,
\end{equation}
where $\mathbf{S}_k$ and $\mathbf{S}_\mathbf{x}$, as defined in Appendix~\ref{appsubsec: generation consistency}, are groups which contain the $i$-th sample.

For rule consistency, samples are reused so that some samples are contained in the test samples for multiple outcome variables. Let $i$ be the index of a sample. Write $\mathbf{Y}=(Y_1,Y_2,\dots,Y_{m_2})$, and let $\mathbf{Z}^{(i)}\subseteq \{1,2,\dots,m_2\}$ be the index of all outcome variables whose test samples contains the $i$-th sample. Then the metric $s_{3,i}^{\mathrm{truth}}$ for the $i$-th sample is
\begin{equation}
    s_{3,i}^{\mathrm{truth}}=\frac{1}{|\mathbf{Z}^{(i)}|}\sum_{j\in \mathbf{Z}_i}\mathds{1}\left(Y^{(i)}_j=\hat{Y}^{(i)}_j\right).
\end{equation}
For the definition of $s_{3,i}^{\mathrm{observe}}$, we also rebalance the samples such that the expected value of $Y_j$, $f_j(\hat{pa}(Y_j))$, has equal weights over $\{0,1\}$. Recall that $g_j=\sum_{i=1}^{2n_3}f_j(\hat{pa}^{(i)}(Y_j))$ is the total number of samples such that the expected value of $Y_j$ is $1$. We define the weight of $Y_j$ in calculating the metric as
\begin{equation}
w_j = n_3\left(\frac{f_j(\hat{pa}^{(i)}(Y_j))}{g_j} + \frac{1-f_j(\hat{pa}^{(i)}(Y_j))}{2n_3-g_j}\right). 
\end{equation}
Then the metric $s_{3,i}^{\mathrm{observe}}$ for sample $i$ is defined as
\begin{equation}
\begin{aligned}
    s_{3,i}^{\mathrm{observe}} &= \frac{\sum_{j\in \mathbf{Z}_i}w_j\mathds{1}\left( \hat{Y}^{(i)}_j=f_j(\hat{pa}^{(i)}(Y_j)) \right)}{\sum_{j\in\mathbf{Z}_i}w_j}.
\end{aligned}
\end{equation}

\section{Manual verification of automatic results}
\label{app:manual_check}
For automatic annotation of causal systems, we have verified the effectiveness through crowd experiments. Here we verify other automatic steps, including:
\begin{itemize}
    \item Section~\ref{app:prompt_correct}: generating text prompts from value combinations,
    \item Section~\ref{app:probe_correct}: generation probe questions from factors,
    \item Section~\ref{app:ask_correct}: retrieve observed value from videos.
\end{itemize}
For each step, we randomly choice some automatic generation in our 60 test cases from our VACT benchmark and manually check whether the automatic annotation is correct.

\subsection{Manual verification of prompt correctness}
\label{app:prompt_correct}
Below are randomly selected scenarios and corresponding prompts from our dataset. We manually verify the correctness by checking whether the two types of prompts (with and without non-root nodes ) match the given values for the variables.

We examined a total of four scenarios and their corresponding 106 prompts. Nearly all of the prompts passed inspection, with the exception of \textbf{two}. The issue with the first prompt arises from our setting the variable ``\textit{sponge is wet}'' to false, while the prompt only specifies that the hand is dry and fails to clarify the condition of the sponge. The second issue pertains to a prompt that was expected to contain only root nodes; however, it includes the word ``\textit{slide}'', which introduces a non-root value.

We show these 106 samples as follows, where we marked the correct ones with \textcolor{green}{\textbf{\checkmark}} and the 2 incorrect results with \textcolor{red}{\textbf{\texttimes}}.

Scenario:
\begin{promptbox}
Rules for each \textcolor{orange}{non-root} node:

\hspace*{2em} \textcolor{orange}{Pile Catches Fire} = (\textcolor{blue}{Ball Actively Burning} $ \wedge $ \textcolor{blue}{Ball Contact Pile}) 

\hspace*{2em} $X_1$ = \textcolor{blue}{Ball Actively Burning} \quad $X_2$ = \textcolor{blue}{Ball Contact Pile} \quad  $Y$ = \textcolor{orange}{Pile Catches Fire}
\end{promptbox}
Prompts:
\begin{promptbox}
                 
Without non-root nodes:

($X_1$ = False , $X_2$ = True)

\hspace*{2em} A smoldering paper ball, now unlit, was tossed but missed the stack of old newspapers.\textcolor{green}{\textbf{\checkmark}}

\hspace*{2em} A ball of paper that had stopped burning eventually settled into a stack of paper.\textcolor{green}{\textbf{\checkmark}}

\hspace*{2em} A cooling ball of paper was placed carelessly into a mound of papers.\textcolor{green}{\textbf{\checkmark}}

\hspace*{2em} An extinguished paper sphere was accidentally dropped into a heap of documents. \textcolor{green}{\textbf{\checkmark}}

\hspace*{2em} A barely expired paper ball softly landed in a collection of scraps. \textcolor{green}{\textbf{\checkmark}}

($X_1$ = True , $X_2$ = True)

\hspace*{2em} A flaming ball of paper crashed into a stack of old newspapers.  \textcolor{green}{\textbf{\checkmark}}

\hspace*{2em} A lit paper ball was hurled into a heap of documents.\textcolor{green}{\textbf{\checkmark}}

\hspace*{2em} A burning paper sphere landed directly in a pile of loose-leaf papers. \textcolor{green}{\textbf{\checkmark}}

\hspace*{2em} A fireball of paper was tossed straight into a mound of papers. \textcolor{green}{\textbf{\checkmark}}

\hspace*{2em} An ignited ball of paper rolled into a collection of scraps. \textcolor{green}{\textbf{\checkmark}}

($X_1$ = True , $X_2$ = False)

\hspace*{2em} A burning ball of paper was thrown close to but missed hitting a pile of paper. \textcolor{green}{\textbf{\checkmark}}

\hspace*{2em} A flaming ball of paper flew past a stack of old newspapers without making contact. \textcolor{green}{\textbf{\checkmark}}

\hspace*{2em} A lit paper ball was launched near a heap of documents, but it didn't touch them. \textcolor{green}{\textbf{\checkmark}}

($X_1$ = False , $X_2$ = False)

\hspace*{2em} A ball of paper, which had extinguished, was thrown away from a pile of paper.\textcolor{green}{\textbf{\checkmark}}

\hspace*{2em} A smoldering paper ball, now unlit, was tossed but missed the stack of old newspapers. \textcolor{green}{\textbf{\checkmark}}

\hspace*{2em} A once aflame ball of paper, now out, was hurled and did not touch the paper heap. \textcolor{green}{\textbf{\checkmark}}

With non-root nodes:

($X_1$ = False , $X_2$ = False, $Y$ = False)

\hspace*{2em} An unlit ball of paper passed by the paper pile without touching it, leaving the pile unburned.\textcolor{green}{\textbf{\checkmark}}

($X_1$ = False , $X_2$ = True, $Y$ = False)

\hspace*{2em} Since the ball wasn't on fire upon contact, the paper pile stayed unharmed.\textcolor{green}{\textbf{\checkmark}}

\hspace*{2em} A non-burning, thrown ball of paper landed on the pile but didn't ignite it. \textcolor{green}{\textbf{\checkmark}}

\hspace*{2em} Though the ball reached the pile, it was not burning, and thus the pile remained safe. \textcolor{green}{\textbf{\checkmark}}

\hspace*{2em} A ball that wasn't actively burning was thrown onto the pile, and the pile stayed unignited. \textcolor{green}{\textbf{\checkmark}}

\hspace*{2em} The paper ball made contact with the pile, but without being on fire, the pile did not catch alight. \textcolor{green}{\textbf{\checkmark}}

($X_1$ = True , $X_2$ = True, $Y$ = True)

\hspace*{2em} The flaming paper sphere, still ablaze, was thrown and hit the paper pile, which then caught fire.\textcolor{green}{\textbf{\checkmark}}

\hspace*{2em} A blazing ball of paper made contact with a stack of paper, causing the pile to ignite, since the ball was burning and it struck the pile.\textcolor{green}{\textbf{\checkmark}}

($X_1$ = True , $X_2$ = False, $Y$ = False)

\hspace*{2em} Despite being actively on fire, the paper ball missed the pile, and as a result, the pile did not catch fire. \textcolor{green}{\textbf{\checkmark}}

\hspace*{2em} Even though the ball was burning, it did not make contact with the pile of paper, so the pile remained unburned. \textcolor{green}{\textbf{\checkmark}}

\end{promptbox}
Scenario:
\begin{promptbox}
Rules for each \textcolor{orange}{non-root} node:

\hspace*{2em} \textcolor{orange}{Butter Sliced} = (\textcolor{blue}{Butter Solid} $ \wedge $ \textcolor{blue}{Downward Slicing Motion Applied}) 

\hspace*{2em} $X_1$ = \textcolor{blue}{Butter Solid} \quad $X_2$ = \textcolor{blue}{Downward Slicing Motion Applied} \quad  $Y$ = \textcolor{orange}{Butter Sliced}
\end{promptbox}
Prompts:
\begin{center}
\begin{promptbox}
                 
Without non-root nodes:

($X_1$ = False , $X_2$ = True)

\hspace*{2em} A knife pierces the soft butter with effortless downward motion.\textcolor{green}{\textbf{\checkmark}}

\hspace*{2em} The knife sweeps downward, slicing perfectly through softened butter.\textcolor{green}{\textbf{\checkmark}}

\hspace*{2em} Swiftly moving downwards, the knife glides through the creamy butter easily. \textcolor{green}{\textbf{\checkmark}}

($X_1$ = True , $X_2$ = True)

\hspace*{2em} A knife slides effortlessly downward through a solid block of butter.  \textcolor{green}{\textbf{\checkmark}}

\hspace*{2em} The solid butter yields smoothly as a knife slices through it with a downward motion.\textcolor{green}{\textbf{\checkmark}}

\hspace*{2em} With a straight down slice, the knife cuts cleanly through the solid butter. \textcolor{green}{\textbf{\checkmark}}

\hspace*{2em} Cutting a solid piece of butter with a knife moving downward feels like slicing through soft clay.\textcolor{green}{\textbf{\checkmark}}

\hspace*{2em} A sturdy push downward sends the knife through the solidified butter seamlessly. \textcolor{green}{\textbf{\checkmark}}

($X_1$ = True , $X_2$ = False)

\hspace*{2em}Simply pressing a knife against the solid butter won't cut it.\textcolor{green}{\textbf{\checkmark}}

\hspace*{2em} The knife doesn't glide through the solid butter without a downward push. \textcolor{green}{\textbf{\checkmark}}

\hspace*{2em} A knife pressed horizontally against the solid butter fails to cut through.\textcolor{green}{\textbf{\checkmark}}

($X_1$ = False , $X_2$ = False)

\hspace*{2em} A knife resting on soft butter is ineffective without downward force. \textcolor{green}{\textbf{\checkmark}}

\hspace*{2em} No downward motion makes the knife linger atop the soft butter. \textcolor{green}{\textbf{\checkmark}}

\hspace*{2em} Simply resting a knife on soft butter won't achieve a cut. \textcolor{green}{\textbf{\checkmark}}

\hspace*{2em} Without cutting downward, a knife barely breaks the soft butter surface. \textcolor{green}{\textbf{\checkmark}}

\hspace*{2em} The knife sits idle against the soft butter, lacking downward pressure. \textcolor{green}{\textbf{\checkmark}}

With non-root nodes:

($X_1$ = False , $X_2$ = False, $Y$ = False)

\hspace*{2em} There is no slicing of the butter, as it is neither solid nor subjected to a downward motion. \textcolor{green}{\textbf{\checkmark}}

\hspace*{2em} With the butter not solid and without a downward motion, no slicing occurs. \textcolor{green}{\textbf{\checkmark}}

\hspace*{2em} Neither solid state nor downward motion is present, leaving the butter unsliced.\textcolor{green}{\textbf{\checkmark}}

\hspace*{2em} The butter is not solid, and no downward slicing motion is applied, so the butter is not sliced. \textcolor{green}{\textbf{\checkmark}}

($X_1$ = True , $X_2$ = True, $Y$ = True)

\hspace*{2em} Since the butter is solid and a downward force is used, the knife slices the butter.\textcolor{green}{\textbf{\checkmark}}

\hspace*{2em} Solid butter is easily sliced through as a downward slicing motion is applied.\textcolor{green}{\textbf{\checkmark}}

\hspace*{2em} The butter, being solid, is sliced through as a downward slicing motion is applied. \textcolor{green}{\textbf{\checkmark}}

\hspace*{2em} The butter is solid, and a downward slicing motion is applied, resulting in the butter being sliced. \textcolor{green}{\textbf{\checkmark}}

\hspace*{2em} With the butter in a solid state and a downward slicing motion in action, the butter gets sliced. \textcolor{green}{\textbf{\checkmark}}

($X_1$ = True , $X_2$ = False, $Y$ = False)

\hspace*{2em} The butter is solid but no downward slicing motion is applied, so the butter is not sliced.\textcolor{green}{\textbf{\checkmark}}
\end{promptbox}
\end{center}
Scenario:
\begin{center}
\begin{promptbox}
Rules for each \textcolor{orange}{non-root} node:

\hspace*{2em} \textcolor{orange}{Water Emerges from Sponge} = (\textcolor{blue}{Sponge is Wet} $ \wedge $ \textcolor{blue}{Hand Fully Compresses Sponge}) 

\hspace*{2em} \textcolor{orange}{Sponge Shape Visibly Changes} = (\textcolor{blue}{Hand Fully Compresses Sponge})

\hspace*{2em} $X_1$ = \textcolor{blue}{Sponge is Wet} \quad $X_2$ = \textcolor{blue}{Hand Fully Compresses Sponge} 

\hspace*{2em} $Y_1$ = \textcolor{orange}{Water Emerges from Sponge} \quad $Y_2$ = \textcolor{orange}{Sponge Shape Visibly Changes}
\end{promptbox}
\end{center}
Prompts:
\begin{center}
\begin{promptbox}
                 
Without non-root nodes:

($X_1$ = False , $X_2$ = True)

\hspace*{2em} A dry sponge is entirely compressed by a hand squeezing it.\textcolor{green}{\textbf{\checkmark}}

\hspace*{2em} The hand fully compresses a dry sponge with its grip.\textcolor{green}{\textbf{\checkmark}}

\hspace*{2em} A hand squeezes a dry sponge until it's fully compressed. \textcolor{green}{\textbf{\checkmark}}

\hspace*{2em} Fully closing, a hand compresses a dry sponge.\textcolor{green}{\textbf{\checkmark}}

\hspace*{2em} The hand squeezes a dry sponge as much as it will go. \textcolor{green}{\textbf{\checkmark}}

($X_1$ = True , $X_2$ = True)

\hspace*{2em} The hand squeezes a wet sponge, fully compressing it.  \textcolor{green}{\textbf{\checkmark}}

\hspace*{2em} A hand grips a wet sponge and fully squeezes it.\textcolor{green}{\textbf{\checkmark}}

\hspace*{2em} The hand exerts force on a wet sponge, squeezing it flat.\textcolor{green}{\textbf{\checkmark}}

\hspace*{2em} A wet sponge is completely compressed by a hand. \textcolor{green}{\textbf{\checkmark}}

\hspace*{2em} A wet sponge is gripped and fully squeezed by a hand. \textcolor{green}{\textbf{\checkmark}}

($X_1$ = True , $X_2$ = False)

\hspace*{2em} Squeezing a wet sponge, the hand stops before fully compressing it.\textcolor{green}{\textbf{\checkmark}}

\hspace*{2em} A hand grips a wet sponge, compressing it only slightly. \textcolor{green}{\textbf{\checkmark}}

\hspace*{2em} The hand applies pressure but doesn't fully squeeze the wet sponge.\textcolor{green}{\textbf{\checkmark}}

\hspace*{2em} A hand gently squeezes a wet sponge without fully compressing it. \textcolor{green}{\textbf{\checkmark}}

\hspace*{2em} A wet sponge is partially squeezed by a hand. \textcolor{green}{\textbf{\checkmark}}

($X_1$ = False , $X_2$ = False)

\hspace*{2em} The hand applies some pressure to a dry sponge but doesn't compress it completely.\textcolor{green}{\textbf{\checkmark}}

\hspace*{2em} A hand holds and gently squeezes a dry sponge without full compression. \textcolor{green}{\textbf{\checkmark}}

\hspace*{2em} The hand grips and squeezes a dry sponge lightly, without full compression. \textcolor{green}{\textbf{\checkmark}}

\hspace*{2em} A hand partially squeezes a dry sponge without complete compression. \textcolor{green}{\textbf{\checkmark}}

With non-root nodes:

($X_1$ = False , $X_2$ = False, $Y_1$ = False, $Y_2$ = False)

\hspace*{2em} The dry sponge remains unchanged when the hand gives it a gentle squeeze both in terms of shape and water release. \textcolor{green}{\textbf{\checkmark}}

($X_1$ = True , $X_2$ = True, $Y$ = True, $Y_2$ = True)

\hspace*{2em} When the hand squeezes the wet sponge completely, the sponge visibly deforms and water emerges.\textcolor{green}{\textbf{\checkmark}}

\hspace*{2em} With a wet sponge being fully pressed by the hand, water seeps out and the sponge's form changes.\textcolor{green}{\textbf{\checkmark}}

\hspace*{2em} The wet sponge is fully compressed by the hand, resulting in a change in its shape and water being squeezed out. \textcolor{green}{\textbf{\checkmark}}

($X_1$ = False , $X_2$ = True, $Y_1$ = False , $Y_2$ = True)

\hspace*{2em} A dry hand compresses the sponge completely, causing its shape to change, but no water releases.\textcolor{red}{\textbf{\texttimes}}

($X_1$ = True , $X_2$ = False, $Y_1$ = False , $Y_2$ = False)

\hspace*{2em} The damp sponge is only partially squeezed by the hand, meaning no water is released and the shape remains consistent.\textcolor{green}{\textbf{\checkmark}}

\hspace*{2em} A hand lightly squeezes the wet sponge, leaving its shape and water content unchanged.\textcolor{green}{\textbf{\checkmark}}

\hspace*{2em} Although the sponge is wet, the hand does not fully compress it, so no water comes out, and its shape stays the same. \textcolor{green}{\textbf{\checkmark}}

\end{promptbox}
\end{center}

Scenario:
\begin{center}
\begin{promptbox}
Rules for each \textcolor{orange}{non-root} node:

\hspace*{2em} \textcolor{orange}{Ice Block Moves} = ($\neg$ \textcolor{blue}{Ice Block On Stable Surface}) 

\hspace*{2em} \textcolor{orange}{Ice Block Cracks} = (\textcolor{blue}{Hammer Head Metal})

\hspace*{2em} $X_1$ = \textcolor{blue}{Ice Block On Stable Surface} \quad $X_2$ = \textcolor{blue}{Hammer Head Metal} 

\hspace*{2em} $Y_1$ = \textcolor{orange}{Ice Block Moves} \quad $Y_2$ = \textcolor{orange}{Ice Block Cracks}
\end{promptbox}
\end{center}
Prompts:
\begin{center}
\begin{promptbox}
                 
Without non-root nodes:

($X_1$ = False , $X_2$ = True)

\hspace*{2em} A person strikes an ice block with a metal hammer, causing it to slide on the surface.\textcolor{red}{\textbf{\texttimes}}

\hspace*{2em} A metal-headed hammer is wielded by a person to hit an ice block that's not stably placed.\textcolor{green}{\textbf{\checkmark}}

\hspace*{2em} Someone hits a sliding ice block with a metal hammer. \textcolor{green}{\textbf{\checkmark}}

\hspace*{2em} An individual uses a metal hammer to strike an ice block that isn't on stable footing. \textcolor{green}{\textbf{\checkmark}}

($X_1$ = True , $X_2$ = True)

\hspace*{2em} A person uses a metal-headed hammer to hit an ice block resting on a stable base. \textcolor{green}{\textbf{\checkmark}}

\hspace*{2em} An individual strikes a stable ice block with a metallic hammer.\textcolor{green}{\textbf{\checkmark}}

\hspace*{2em} A hammer with a metal head is used by a person to hit a stable ice block.\textcolor{green}{\textbf{\checkmark}}

\hspace*{2em} An ice block on a stable platform is struck by someone wielding a metal hammer. \textcolor{green}{\textbf{\checkmark}}

($X_1$ = True , $X_2$ = False)

\hspace*{2em}An individual hits a secure ice block with a hammer that lacks a metal head. \textcolor{green}{\textbf{\checkmark}}

\hspace*{2em} Someone uses a non-metallic hammer to hit an ice block resting stably.\textcolor{green}{\textbf{\checkmark}}

\hspace*{2em} A person uses a hammer with a non-metal head to hit an ice block on a stable surface.\textcolor{green}{\textbf{\checkmark}}

\hspace*{2em} Striking a solidly placed ice block with a hammer that doesn't have a metal head. \textcolor{green}{\textbf{\checkmark}}

($X_1$ = False , $X_2$ = False)

\hspace*{2em} A person hits an ice block with a non-metal hammer, and the block is not stable.\textcolor{green}{\textbf{\checkmark}}

\hspace*{2em} Striking a shifting ice block with a hammer that has a non-metal head. \textcolor{green}{\textbf{\checkmark}}

\hspace*{2em} The hand fully compresses a dry sponge with its grip.Using a non-metal headed hammer, a person hits an unsteady block of ice.\textcolor{green}{\textbf{\checkmark}}

\hspace*{2em} The ice block, not secure, is struck by a person with a non-metal hammer. \textcolor{green}{\textbf{\checkmark}}

\hspace*{2em} Someone uses a hammer without a metal head to hit a loosely sitting ice block. \textcolor{green}{\textbf{\checkmark}}

With non-root nodes:

($X_1$ = True , $X_2$ = True, $Y_1$ = False, $Y_2$ = True)

\hspace*{2em} The ice block, resting securely on a stable surface, is struck by a hammer with a metal head, which causes it to crack. \textcolor{green}{\textbf{\checkmark}}

($X_1$ = False , $X_2$ = False, $Y$ = True, $Y_2$ = False)

\hspace*{2em} The ice block on an unsteady surface moves but does not crack when struck with a non-metal hammer. \textcolor{green}{\textbf{\checkmark}}

\hspace*{2em} Even on an unsteady surface, the ice block only shifts without cracking when hit by a non-metal hammer.\textcolor{green}{\textbf{\checkmark}}

\hspace*{2em} A non-metal hammer causes the ice block on an unstable surface to move but avoids cracking. \textcolor{green}{\textbf{\checkmark}}

\hspace*{2em} An ice block shifts on its unstable foundation, though uncracked, under a non-metal hammer blow. \textcolor{green}{\textbf{\checkmark}}

($X_1$ = True , $X_2$ = False, $Y_1$ = False , $Y_2$ = False)

\hspace*{2em} The ice block, placed securely on a stable surface, does not move or crack when struck by a non-metal hammer.\textcolor{green}{\textbf{\checkmark}}

\hspace*{2em} Striking the ice block on a stable foundation with a non-metal hammer results in no movement or cracking.\textcolor{green}{\textbf{\checkmark}}

\hspace*{2em} A hammer with a non-metal head hits an ice block on stable ground, neither moves nor cracks it.\textcolor{green}{\textbf{\checkmark}}

\hspace*{2em} The ice block, secured by its stable surface, withstands the non-metal hammer blow without cracking or shifting. \textcolor{green}{\textbf{\checkmark}}

\hspace*{2em} A non-metal hammer strikes the ice block on stable ground, leaving it neither cracked nor moved. \textcolor{green}{\textbf{\checkmark}}

\end{promptbox}
\end{center}

\subsection{Manual Verification of Factor-Question Alignment}
\label{app:probe_correct}

To evaluate whether the generated videos comply with causal rules, we utilize a VLLM to extract the values of both root and non-root nodes. When posing ``yes-no'' questions about the video, it is essential to ensure that the questions are appropriately aligned with the relevant factors in each specific scenario.

In this section, We randomly selected 6 scenarios, comprising a total of 20 factor-question pairs, all of which were found to be correct.

Scenario: A small ball impacts the ground.
\begin{center} 
\small
\begin{tabular}{p{0.35\linewidth} p{0.45\linewidth} c}
\toprule
Factor & Question & Correct \\
\midrule
ball is deflated & Is the ball deflated? & \textcolor{green}{\textbf{\checkmark}} \\
the ground is soft & Is the ground soft? & \textcolor{green}{\textbf{\checkmark}} \\
ball bounces & Does the ball bounce? & \textcolor{green}{\textbf{\checkmark}} \\
\bottomrule
\end{tabular}
\end{center}

Scenario: Sunlight shines on the water surface, creating sparkling reflections.
\begin{center} 
\small
\begin{tabular}{p{0.42\linewidth} p{0.38\linewidth} c}
\toprule
Factor & Question & Correct \\
\midrule
direct sunlight present & Is direct sunlight present on the water surface? & \textcolor{green}{\textbf{\checkmark}} \\
water ripples visible & Are water ripples visible on the surface? & \textcolor{green}{\textbf{\checkmark}} \\
unobstructed water surface & Is the water surface unobstructed? & \textcolor{green}{\textbf{\checkmark}} \\
\bottomrule

\end{tabular}
\end{center}

Scenario: A person strikes an ice block with a hammer.
\begin{center} 
\small
\begin{tabular}{p{0.35\linewidth} p{0.45\linewidth} c}
\toprule
Factor & Question & Correct \\
\midrule
block is small & Is the ice block small? & \textcolor{green}{\textbf{\checkmark}} \\
direct hammer strike & Is the hammer striking the ice block directly? & \textcolor{green}{\textbf{\checkmark}} \\
block breaks & Does the ice block break when struck? & \textcolor{green}{\textbf{\checkmark}} \\
\bottomrule

\end{tabular}
\end{center}

Scenario: Flag waving in the wind at the top of pole.
\begin{center} 
\small
\begin{tabular}{p{0.25\linewidth} p{0.55\linewidth} c}
\toprule
Factor & Question & Correct \\
\midrule
is flag hoisted & Is the flag hoisted at the top of the pole? & \textcolor{green}{\textbf{\checkmark}} \\
is there wind & Is there wind present in the environment? & \textcolor{green}{\textbf{\checkmark}} \\
flag waving & Is the flag waving? & \textcolor{green}{\textbf{\checkmark}} \\
\bottomrule

\end{tabular}
\end{center}

Scenario: A broom drags across the dirty ceramic floor.
\begin{center} 
\small
\begin{tabular}{p{0.45\linewidth} p{0.4\linewidth} c}
\toprule
Factor & Question & Correct \\
\midrule
broom bristles contact floor & Are the broom bristles making contact with the floor? & \textcolor{green}{\textbf{\checkmark}} \\  
floor is wet & Is the floor wet? & \textcolor{green}{\textbf{\checkmark}} \\
obstruction on floor & Is there an obstruction on the floor? & \textcolor{green}{\textbf{\checkmark}} \\
floor becomes clean & Does the floor become clean after using the broom? & \textcolor{green}{\textbf{\checkmark}} \\
\bottomrule

\end{tabular}
\end{center}

Scenario: Drop dye into the water.
\begin{center} 
\small
\begin{tabular}{p{0.35\linewidth} p{0.45\linewidth} c}
\toprule
Factor & Question & Correct \\
\midrule
dye is water soluble & Is the dye water soluble? & \textcolor{green}{\textbf{\checkmark}} \\ 
water is stirred & Is the water stirred? & \textcolor{green}{\textbf{\checkmark}} \\
water becomes colored & Does the water become colored? & \textcolor{green}{\textbf{\checkmark}} \\
water becomes uniformly colored & Does the water become uniformly colored? & \textcolor{green}{\textbf{\checkmark}} \\
\bottomrule

\end{tabular}
\end{center}

\subsection{Manual verification of VLLM answer retrieval correctness}
\label{app:ask_correct}

The answers provided by VLLM serve as the foundation for calculating the final score of the generated videos. Therefore, it is essential to manually verify the accuracy of these responses. In this section, we select four models and examine three distinct scenarios, each accompanied by three corresponding prompts.

The sampled scenarios encompass both challenging and easy prompts, with and without non-root nodes, and feature answers classified as True, False, or N/A. A comprehensive explanation of the conditions under which VLLM provides an N/A response is available in ~\ref{app:affect_of_na}. For example, the explanation provided by VLLM for the N/A response regarding a video generated by Pika, as presented in Table~\ref{tab:s2p3}, is: ``\textit{The images do not provide a clear view of the top of the boot. It is not possible to determine if it is sealed or not from the given angles.}'' for the factor ``\textit{boot top sealed}'', which is consistent with our observations.

Regarding the accuracy of model responses, we find that VLLM demonstrates sufficient capability to handle simple scenarios and prompts (such as those in Table~\ref{tab:s1p3}, Table~\ref{tab:s2p2},Table~\ref{tab:s2p3},Table~\ref{tab:s3p1},Table~\ref{tab:s3p2}, and Table~\ref{tab:s3p3}). However, its performance declines when addressing more complex questions (such as those in Table~\ref{tab:s1p1}, Table~\ref{tab:s1p2}, and Table~\ref{tab:s2p1}). Currently, the accuracy of this approach hovers around 95\%, which is acceptable but still leaves room for improvement. The shortcomings in correctness primarily stem from two factors. First, the VGMs often generate videos with low quality and ambiguity, which increases the difficulty for VLLM to provide accurate answers. Additionally, VLLMs still lack the ability to clearly understand intricate details in images or videos, particularly when dealing with more complex questions. Nevertheless, we are optimistic that as the foundational capabilities of VLLMs continue to improve, the performance of this video description system will experience significant enhancement.

The checked question-answer pairs are shown below, accompanied by the generated videos. Our verification results are presented in a table that closely follows each prompt.

\textbf{Scenario: A ray of light is shining on a wooden block.}

\begin{figure}[H]
    \centering
    \begin{subfigure}{1\linewidth}
    \includegraphics[width=1\linewidth]{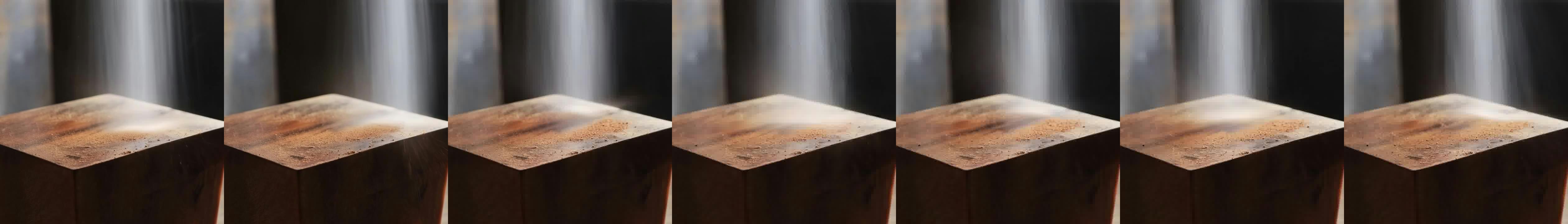}
    \caption{HunyuanVideo Generation}
    
    \end{subfigure}
    \begin{subfigure}{1\linewidth}
    \includegraphics[width=1\linewidth]{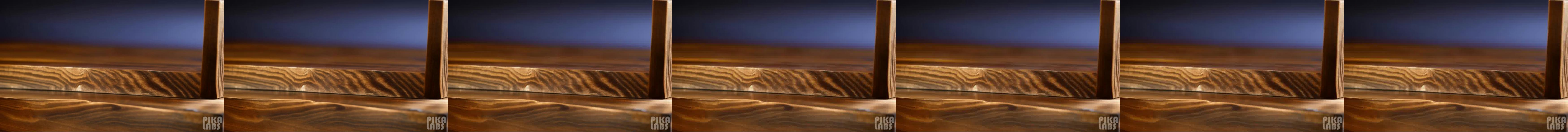}
    \caption{Pika Generation}

    \end{subfigure}
    \begin{subfigure}{1\linewidth}
    \includegraphics[width=1\linewidth]{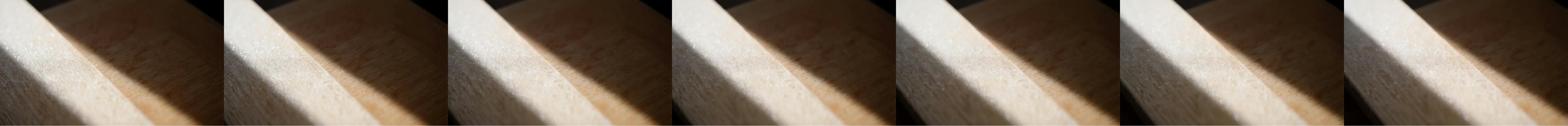}
    \caption{Hailuo Generation}

    \end{subfigure}
    \begin{subfigure}{1\linewidth}
    \includegraphics[width=1\linewidth]{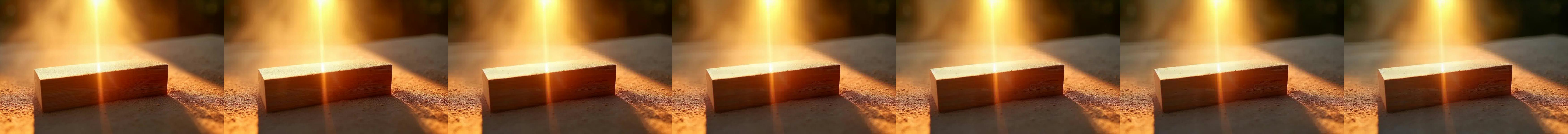}    
    \caption{Pyramid Generation}
    
    \end{subfigure}
    
    \caption{Model Generation}
    \label{fig:s1p1}
\end{figure}

Prompt-1: A beam of light grazes the polished surface of a wooden block in dust. (Videos: Figure~\ref{fig:s1p1}; Results: Tabel~\ref{tab:s1p1})

\begin{table}[H]
\centering
\begin{tabular}{c c c c}
\toprule
\multirow{2}{*}{Model} & \multirow{2}{*}{Factor} & Model & \multirow{2}{*}{Correct} \\ & & Answer & \\
\midrule
\multirow{6}{*}{Hunyuan} & in direct path & True & \textcolor{green}{\textbf{\checkmark}} \\ 
& surface polished & False & \textcolor{red}{\textbf{\texttimes}} \\
& environment dusty & False & \textcolor{green}{\textbf{\checkmark}} \\
& block illuminated & True & \textcolor{green}{\textbf{\checkmark}} \\
& reflection visible & False & \textcolor{green}{\textbf{\checkmark}} \\
& beam visible in air & True & \textcolor{green}{\textbf{\checkmark}} \\
\midrule
\multirow{6}{*}{Pika} & in direct path & False & \textcolor{green}{\textbf{\checkmark}} \\ 
& surface polished & True & \textcolor{green}{\textbf{\checkmark}} \\
& environment dusty & False & \textcolor{green}{\textbf{\checkmark}} \\
& block illuminated & False & \textcolor{green}{\textbf{\checkmark}} \\
& reflection visible & False & \textcolor{green}{\textbf{\checkmark}} \\
& beam visible in air & False & \textcolor{green}{\textbf{\checkmark}} \\
\midrule
\multirow{6}{*}{Hailuo} & in direct path & N/A & \textcolor{green}{\textbf{\checkmark}} \\ 
& surface polished & False & \textcolor{green}{\textbf{\checkmark}} \\
& environment dusty & False & \textcolor{green}{\textbf{\checkmark}} \\
& block illuminated & True & \textcolor{green}{\textbf{\checkmark}} \\
& reflection visible & True & \textcolor{red}{\textbf{\texttimes}} \\
& beam visible in air & False & \textcolor{green}{\textbf{\checkmark}} \\
\midrule
\multirow{6}{*}{Pyramid} & in direct path & True & \textcolor{green}{\textbf{\checkmark}} \\ 
& surface polished & False & \textcolor{green}{\textbf{\checkmark}} \\
& environment dusty & True & \textcolor{green}{\textbf{\checkmark}} \\
& block illuminated & True & \textcolor{green}{\textbf{\checkmark}} \\
& reflection visible & False & \textcolor{green}{\textbf{\checkmark}} \\
& beam visible in air & True & \textcolor{green}{\textbf{\checkmark}} \\
\bottomrule
\end{tabular}
\caption{Verification of VLLM Answer Correctness}
\label{tab:s1p1}
\end{table}

\begin{figure}[H]
    \centering
    \begin{subfigure}{1\linewidth}
    \includegraphics[width=1\linewidth]{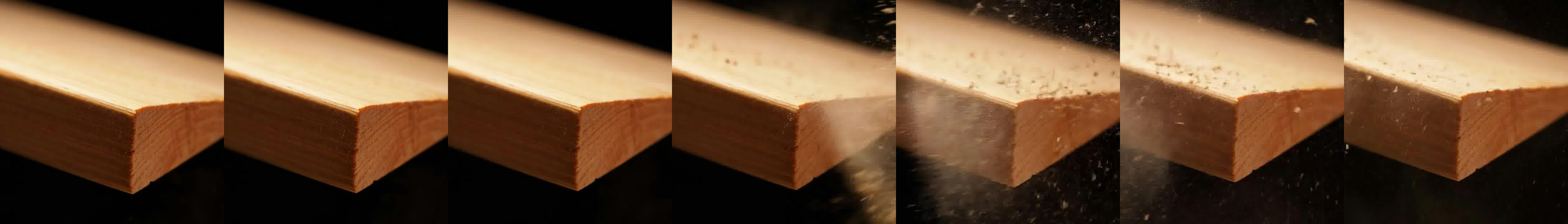}
    \caption{HunyuanVideo Generation}
    
    \end{subfigure}
    \begin{subfigure}{1\linewidth}
    \includegraphics[width=1\linewidth]{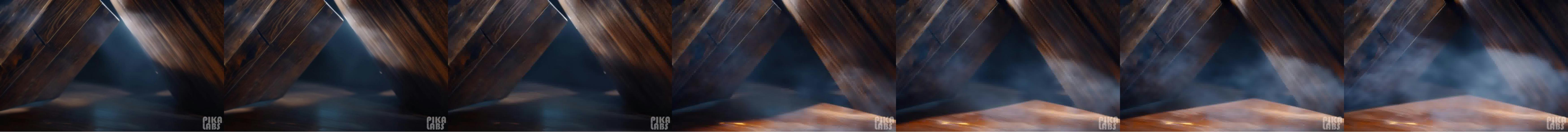}
    \caption{Pika Generation}

    \end{subfigure}
    \begin{subfigure}{1\linewidth}
    \includegraphics[width=1\linewidth]{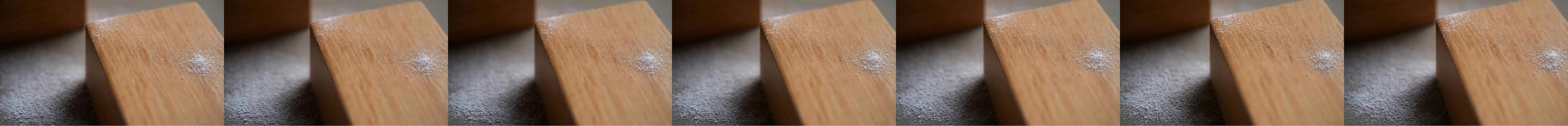}
    \caption{Hailuo Generation}

    \end{subfigure}
    \begin{subfigure}{1\linewidth}
    \includegraphics[width=1\linewidth]{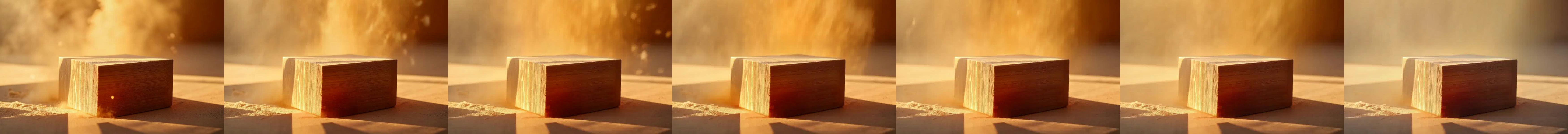}
    \caption{Pyramid Generation}
    
    \end{subfigure}

    \caption{Model Generation}
    \label{fig:s1p2}
\end{figure}

Prompt-2:The polished surface of a wooden block directly catches the light amid dust. (Videos:Figure~\ref{fig:s1p2};Results: Tabel~\ref{tab:s1p2})

\begin{table}[H]
\centering
\begin{tabular}{c c c c}
\toprule
\multirow{2}{*}{Model} & \multirow{2}{*}{Factor} & Model & \multirow{2}{*}{Correct} \\ & & Answer & \\
\midrule
\multirow{6}{*}{Hunyuan} & in direct path & False & \textcolor{green}{\textbf{\checkmark}} \\ 
& surface polished & False & \textcolor{green}{\textbf{\checkmark}} \\
& environment dusty & True & \textcolor{green}{\textbf{\checkmark}} \\
& block illuminated & False & \textcolor{green}{\textbf{\checkmark}} \\
& reflection visible & False & \textcolor{green}{\textbf{\checkmark}} \\
& beam visible in air & False & \textcolor{green}{\textbf{\checkmark}} \\
\midrule
\multirow{6}{*}{Pika} & in direct path & False & \textcolor{green}{\textbf{\checkmark}} \\ 
& surface polished & False & \textcolor{red}{\textbf{\texttimes}} \\
& environment dusty & True & \textcolor{green}{\textbf{\checkmark}} \\
& block illuminated & False & \textcolor{green}{\textbf{\checkmark}} \\
& reflection visible & False & \textcolor{green}{\textbf{\checkmark}} \\
& beam visible in air & True & \textcolor{green}{\textbf{\checkmark}} \\
\midrule
\multirow{6}{*}{Hailuo} & in direct path & False & \textcolor{green}{\textbf{\checkmark}} \\ 
& surface polished & False & \textcolor{green}{\textbf{\checkmark}} \\
& environment dusty & True & \textcolor{green}{\textbf{\checkmark}} \\
& block illuminated & False & \textcolor{green}{\textbf{\checkmark}} \\
& reflection visible & False & \textcolor{green}{\textbf{\checkmark}} \\
& beam visible in air & False & \textcolor{green}{\textbf{\checkmark}} \\
\midrule
\multirow{6}{*}{Pyramid} & in direct path & True & \textcolor{green}{\textbf{\checkmark}} \\ 
& surface polished & False & \textcolor{green}{\textbf{\checkmark}} \\
& environment dusty & True & \textcolor{green}{\textbf{\checkmark}} \\
& block illuminated & True & \textcolor{green}{\textbf{\checkmark}} \\
& reflection visible & True & \textcolor{red}{\textbf{\texttimes}} \\
& beam visible in air & False & \textcolor{green}{\textbf{\checkmark}} \\
\bottomrule
\end{tabular}
\caption{Verification of VLLM Answer Correctness}
\label{tab:s1p2}
\end{table}

\begin{figure}[H]
    \centering
    \begin{subfigure}{1\linewidth}
    \includegraphics[width=1\linewidth]{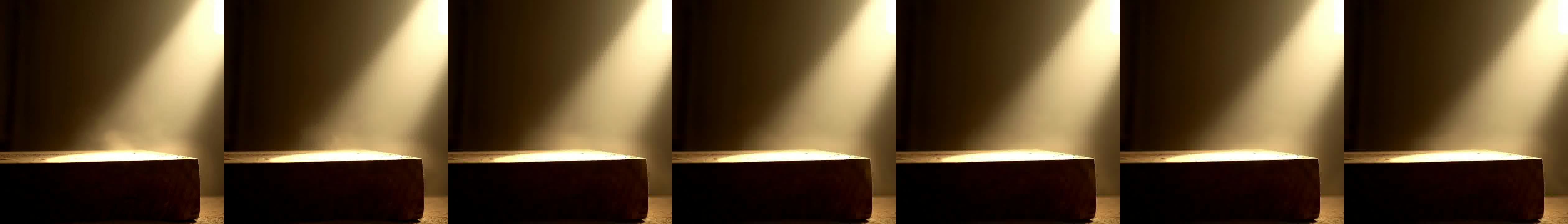}
    \caption{HunyuanVideo Generation}
    
    \end{subfigure}
    \begin{subfigure}{1\linewidth}
    \includegraphics[width=1\linewidth]{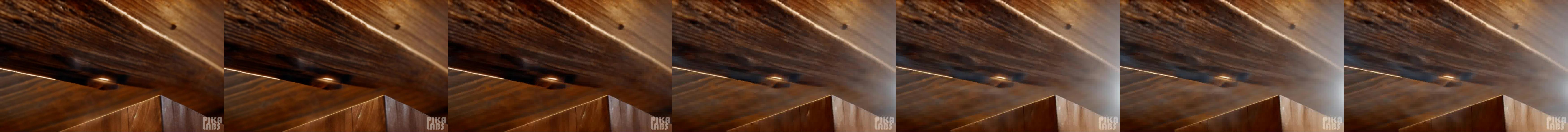}
    \caption{Pika Generation}
    
    \end{subfigure}

    \begin{subfigure}{1\linewidth}
    \includegraphics[width=1\linewidth]{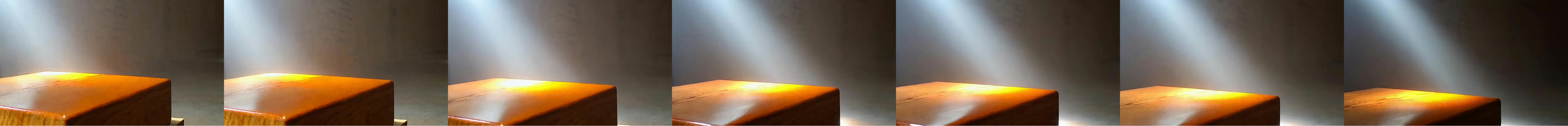}
    \caption{Hailuo Generation}
    
    \end{subfigure}

    \begin{subfigure}{1\linewidth}
    \includegraphics[width=1\linewidth]{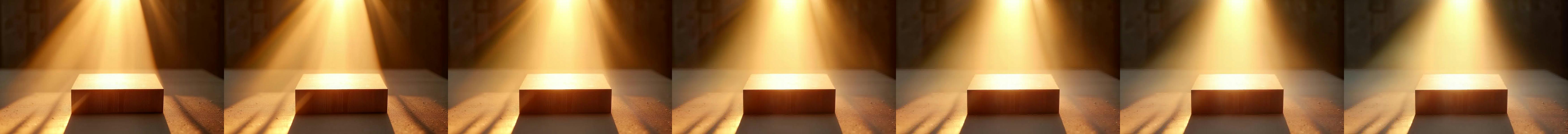}
    \caption{Pyramid Generation}
    
    \end{subfigure}

    \caption{Model Generation}
    \label{fig:s1p3}
\end{figure}

Prompt-3:A ray of light directly illuminates a polished wooden block and the environment is dusty, causing both the block to be lit and reflections to be visible, with the beam clearly seen in the air. (Videos:Figure~\ref{fig:s1p3};Results: Tabel~\ref{tab:s1p3})

\begin{table}[H]
\centering
\begin{tabular}{c c c c}
\toprule
\multirow{2}{*}{Model} & \multirow{2}{*}{Factor} & Model & \multirow{2}{*}{Correct} \\ & & Answer & \\
\midrule
\multirow{6}{*}{Hunyuan} & in direct path & True & \textcolor{green}{\textbf{\checkmark}} \\ 
& surface polished & False & \textcolor{green}{\textbf{\checkmark}} \\
& environment dusty & True & \textcolor{green}{\textbf{\checkmark}} \\
& block illuminated & True & \textcolor{green}{\textbf{\checkmark}} \\
& reflection visible & False & \textcolor{green}{\textbf{\checkmark}} \\
& beam visible in air & True & \textcolor{green}{\textbf{\checkmark}} \\
\midrule
\multirow{6}{*}{Pika} & in direct path & True & \textcolor{green}{\textbf{\checkmark}} \\ 
& surface polished & False & \textcolor{green}{\textbf{\checkmark}} \\
& environment dusty & True & \textcolor{green}{\textbf{\checkmark}} \\
& block illuminated & False & \textcolor{green}{\textbf{\checkmark}} \\
& reflection visible & False & \textcolor{green}{\textbf{\checkmark}} \\
& beam visible in air & True & \textcolor{green}{\textbf{\checkmark}} \\
\midrule
\multirow{6}{*}{Hailuo} & in direct path & True & \textcolor{green}{\textbf{\checkmark}} \\ 
& surface polished & True & \textcolor{green}{\textbf{\checkmark}} \\
& environment dusty & False & \textcolor{green}{\textbf{\checkmark}} \\
& block illuminated & True & \textcolor{green}{\textbf{\checkmark}} \\
& reflection visible & False & \textcolor{green}{\textbf{\checkmark}} \\
& beam visible in air & True & \textcolor{green}{\textbf{\checkmark}} \\
\midrule
\multirow{6}{*}{Pyramid} & in direct path & True & \textcolor{green}{\textbf{\checkmark}} \\ 
& surface polished & False & \textcolor{green}{\textbf{\checkmark}} \\
& environment dusty & True & \textcolor{green}{\textbf{\checkmark}} \\
& block illuminated & True & \textcolor{green}{\textbf{\checkmark}} \\
& reflection visible & False & \textcolor{green}{\textbf{\checkmark}} \\
& beam visible in air & True & \textcolor{green}{\textbf{\checkmark}} \\
\bottomrule
\end{tabular}
\caption{Verification of VLLM Answer Correctness}
\label{tab:s1p3}
\end{table}

\textbf{Scenario: A boot stomps into a puddle of mud.}

\begin{figure}[H]
    \centering
    \begin{subfigure}{1\linewidth}
    \includegraphics[width=1\linewidth]{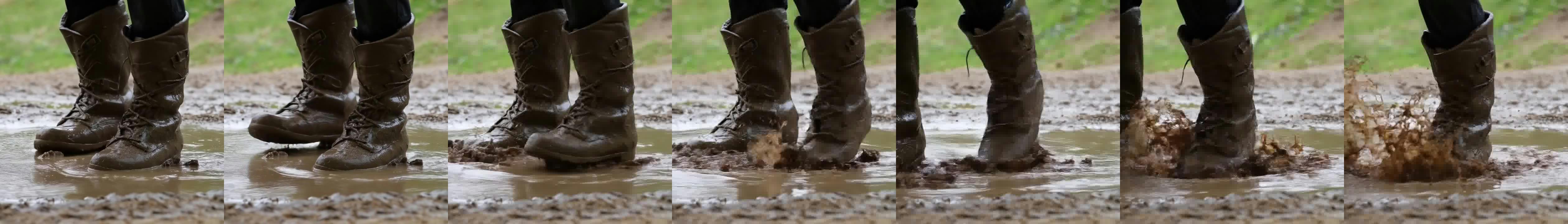}
    \caption{HunyuanVideo Generation}
    
    \end{subfigure}
    \begin{subfigure}{1\linewidth}
    \includegraphics[width=1\linewidth]{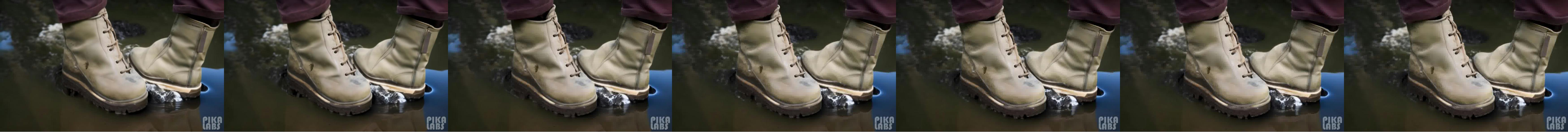}
    \caption{Pika Generation}

    \end{subfigure}
    \begin{subfigure}{1\linewidth}
    \includegraphics[width=1\linewidth]{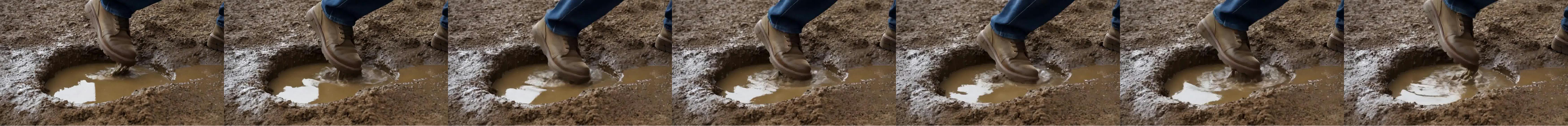}
    \caption{Hailuo Generation}

    \end{subfigure}
    \begin{subfigure}{1\linewidth}
    \includegraphics[width=1\linewidth]{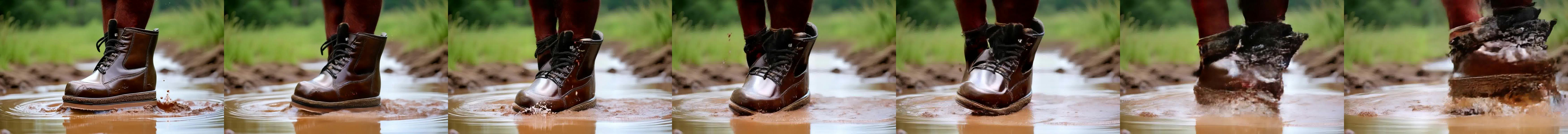}    
    \caption{Pyramid Generation}
    
    \end{subfigure}

    \caption{Model Generation}
    \label{fig:s2p1}
\end{figure}

Prompt-1:An intense stomp by an open-topped boot into a puddle of watery mud occurs. (Videos:Figure~\ref{fig:s2p1};Results: Tabel~\ref{tab:s2p1})

\begin{table}[H]
\centering
\begin{tabular}{c c c c}
\toprule
\multirow{2}{*}{Model} & \multirow{2}{*}{Factor} & Model & \multirow{2}{*}{Correct} \\ & & Answer & \\
\midrule
\multirow{5}{*}{Hunyuan} & watery mud & True & \textcolor{green}{\textbf{\checkmark}} \\ 
& big downward stomp & True& \textcolor{green}{\textbf{\checkmark}} \\
& boot top sealed & False & \textcolor{green}{\textbf{\checkmark}} \\
& mud splashes out of puddle & True & \textcolor{green}{\textbf{\checkmark}} \\
& mud enters the boot & False & \textcolor{green}{\textbf{\checkmark}} \\
\midrule
\multirow{5}{*}{Pika} & watery mud & True & \textcolor{green}{\textbf{\checkmark}} \\ 
& big downward stomp & False & \textcolor{green}{\textbf{\checkmark}} \\
& boot top sealed & True & \textcolor{red}{\textbf{\texttimes}} \\
& mud splashes out of puddle & False & \textcolor{green}{\textbf{\checkmark}} \\
& mud enters the boot & False & \textcolor{green}{\textbf{\checkmark}} \\
\midrule
\multirow{5}{*}{Hailuo} & watery mud & True & \textcolor{green}{\textbf{\checkmark}} \\ 
& big downward stomp & True & \textcolor{green}{\textbf{\checkmark}} \\
& boot top sealed & N/A & \textcolor{green}{\textbf{\checkmark}} \\
& mud splashes out of puddle & True & \textcolor{red}{\textbf{\texttimes}} \\
& mud enters the boot & False & \textcolor{green}{\textbf{\checkmark}} \\
\midrule
\multirow{5}{*}{Pyramid} & watery mud & True & \textcolor{green}{\textbf{\checkmark}} \\ 
& big downward stomp & True & \textcolor{green}{\textbf{\checkmark}} \\
& boot top sealed & False & \textcolor{green}{\textbf{\checkmark}} \\
& mud splashes out of puddle & True & \textcolor{green}{\textbf{\checkmark}} \\
& mud enters the boot & False & \textcolor{green}{\textbf{\checkmark}} \\

\bottomrule
\end{tabular}
\caption{Verification of VLLM Answer Correctness}
\label{tab:s2p1}
\end{table}

\begin{figure}[H]
    \centering
    \begin{subfigure}{1\linewidth}
    \includegraphics[width=1\linewidth]{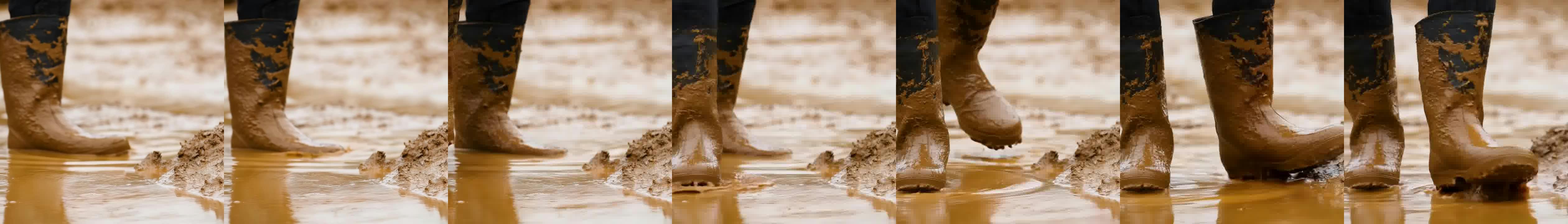}
    \caption{HunyuanVideo Generation}
    
    \end{subfigure}
    \begin{subfigure}{1\linewidth}
    \includegraphics[width=1\linewidth]{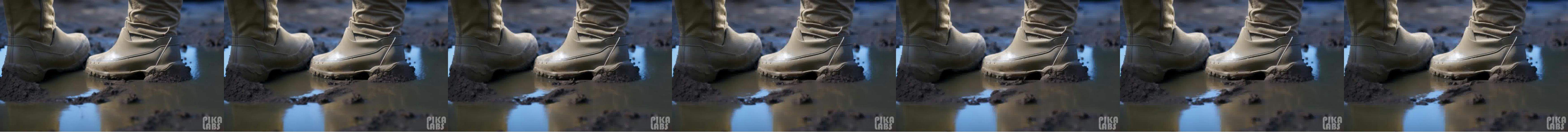}
    \caption{Pika Generation}

    \end{subfigure}
    \begin{subfigure}{1\linewidth}
    \includegraphics[width=1\linewidth]{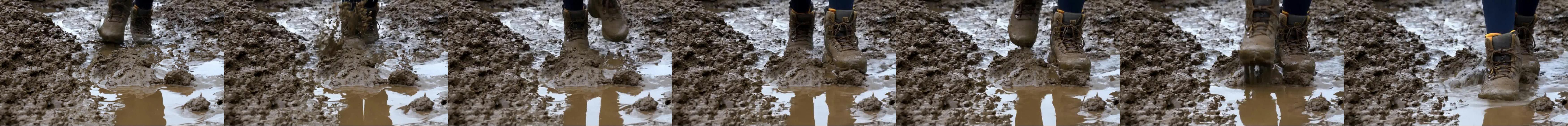}
    \caption{Hailuo Generation}

    \end{subfigure}
    \begin{subfigure}{1\linewidth}
    \includegraphics[width=1\linewidth]{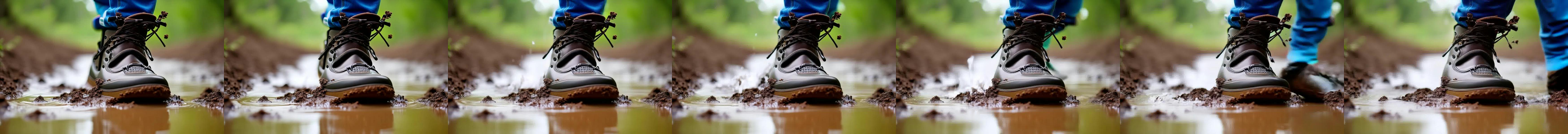}    
    \caption{Pyramid Generation}
    
    \end{subfigure}

    \caption{Model Generation}
    \label{fig:s2p2}
\end{figure}

Prompt-2:In non-watery mud, no splashes occur, but mud enters an unsealed boot during light stepping. (Videos:Figure~\ref{fig:s2p2};Results: Tabel~\ref{tab:s2p2})

\begin{table}[H]
\centering
\begin{tabular}{c c c c}
\toprule
\multirow{2}{*}{Model} & \multirow{2}{*}{Factor} & Model & \multirow{2}{*}{Correct} \\
& & Answer & \\
\midrule
\multirow{5}{*}{Hunyuan} & watery mud & True & \textcolor{green}{\textbf{\checkmark}} \\ 
& big downward stomp & False & \textcolor{green}{\textbf{\checkmark}} \\
& boot top sealed & False & \textcolor{green}{\textbf{\checkmark}} \\
& mud splashes out of puddle & False & \textcolor{green}{\textbf{\checkmark}} \\
& mud enters the boot & False & \textcolor{green}{\textbf{\checkmark}} \\
\midrule
\multirow{5}{*}{Pika} & watery mud & True & \textcolor{green}{\textbf{\checkmark}} \\ 
& big downward stomp & False & \textcolor{green}{\textbf{\checkmark}} \\
& boot top sealed & True & \textcolor{green}{\textbf{\checkmark}} \\
& mud splashes out of puddle & False & \textcolor{green}{\textbf{\checkmark}} \\
& mud enters the boot & False & \textcolor{green}{\textbf{\checkmark}} \\
\midrule
\multirow{5}{*}{Hailuo} & watery mud & True & \textcolor{green}{\textbf{\checkmark}} \\ 
& big downward stomp & True & \textcolor{green}{\textbf{\checkmark}} \\
& boot top sealed & True & \textcolor{green}{\textbf{\checkmark}} \\
& mud splashes out of puddle & True & \textcolor{green}{\textbf{\checkmark}} \\
& mud enters the boot & False & \textcolor{green}{\textbf{\checkmark}} \\
\midrule
\multirow{5}{*}{Pyramid} & watery mud & True & \textcolor{green}{\textbf{\checkmark}} \\ 
& big downward stomp & True & \textcolor{green}{\textbf{\checkmark}} \\
& boot top sealed & False & \textcolor{green}{\textbf{\checkmark}} \\
& mud splashes out of puddle & True & \textcolor{green}{\textbf{\checkmark}} \\
& mud enters the boot & N/A & \textcolor{green}{\textbf{\checkmark}} \\

\bottomrule
\end{tabular}
\caption{Verification of VLLM Answer Correctness}
\label{tab:s2p2}
\end{table}

\begin{figure}[H]
    \centering
    \begin{subfigure}{1\linewidth}
    \includegraphics[width=1\linewidth]{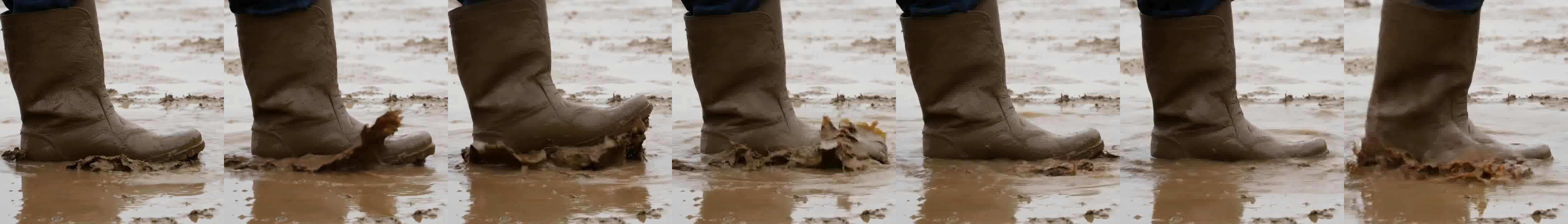}
    \caption{HunyuanVideo Generation}
    
    \end{subfigure}
    \begin{subfigure}{1\linewidth}
    \includegraphics[width=1\linewidth]{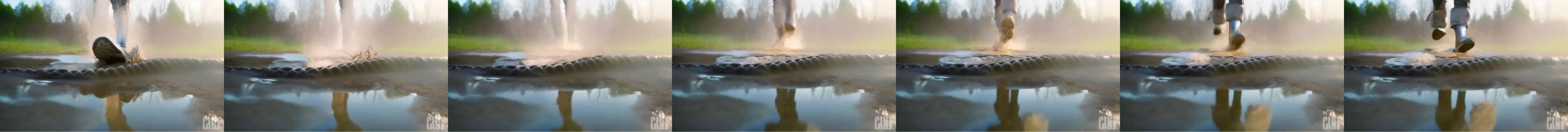}
    \caption{Pika Generation}

    \end{subfigure}
    \begin{subfigure}{1\linewidth}
    \includegraphics[width=1\linewidth]{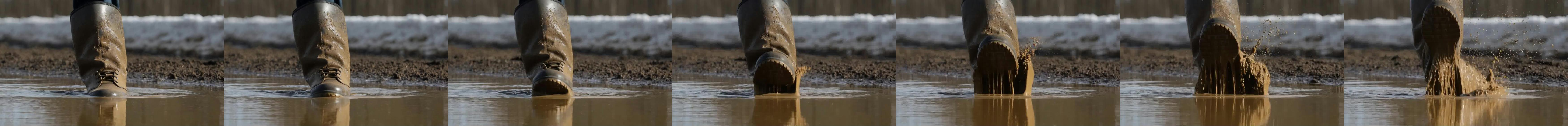}
    \caption{Hailuo Generation}

    \end{subfigure}
    \begin{subfigure}{1\linewidth}
    \includegraphics[width=1\linewidth]{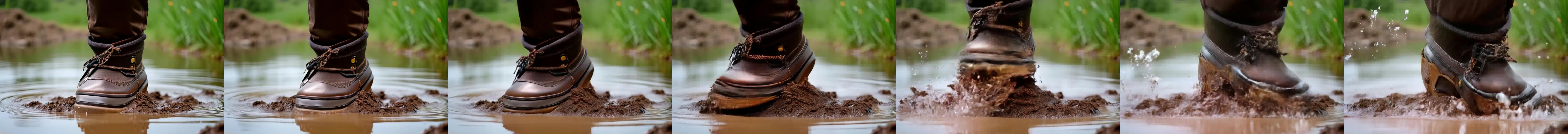}    
    \caption{Pyramid Generation}
    
    \end{subfigure}

    \caption{Model Generation}
    \label{fig:s2p3}
\end{figure}

Prompt-3:A boot with a sealed top makes a big downward stomp into watery mud, causing mud to splash out of the puddle but none enters the boot. (Videos:Figure~\ref{fig:s2p3};Results: Tabel~\ref{tab:s2p3})

\begin{table}[H]
\centering
\begin{tabular}{c c c c}
\toprule
\multirow{2}{*}{Model} & \multirow{2}{*}{Factor} & Model & \multirow{2}{*}{Correct} \\ & & Answer & \\
\midrule
\multirow{5}{*}{Hunyuan} & watery mud & True & \textcolor{green}{\textbf{\checkmark}} \\ 
& big downward stomp & True & \textcolor{green}{\textbf{\checkmark}} \\
& boot top sealed & True & \textcolor{green}{\textbf{\checkmark}} \\
& mud splashes out of puddle & True & \textcolor{green}{\textbf{\checkmark}} \\
& mud enters the boot & False & \textcolor{green}{\textbf{\checkmark}} \\
\midrule
\multirow{5}{*}{Pika} & watery mud & True & \textcolor{green}{\textbf{\checkmark}} \\ 
& big downward stomp & True & \textcolor{green}{\textbf{\checkmark}} \\
& boot top sealed & N/A & \textcolor{green}{\textbf{\checkmark}} \\
& mud splashes out of puddle & True & \textcolor{green}{\textbf{\checkmark}} \\
& mud enters the boot & N/A & \textcolor{green}{\textbf{\checkmark}} \\
\midrule
\multirow{5}{*}{Hailuo} & watery mud & True & \textcolor{green}{\textbf{\checkmark}} \\ 
& big downward stomp & True & \textcolor{green}{\textbf{\checkmark}} \\
& boot top sealed & N/A & \textcolor{green}{\textbf{\checkmark}} \\
& mud splashes out of puddle & True & \textcolor{green}{\textbf{\checkmark}} \\
& mud enters the boot & False & \textcolor{green}{\textbf{\checkmark}} \\
\midrule
\multirow{5}{*}{Pyramid} & watery mud & True & \textcolor{green}{\textbf{\checkmark}} \\ 
& big downward stomp & True & \textcolor{green}{\textbf{\checkmark}} \\
& boot top sealed & True & \textcolor{green}{\textbf{\checkmark}} \\
& mud splashes out of puddle & True & \textcolor{green}{\textbf{\checkmark}} \\
& mud enters the boot & False & \textcolor{green}{\textbf{\checkmark}} \\

\bottomrule
\end{tabular}
\caption{Verification of VLLM Answer Correctness}
\label{tab:s2p3}
\end{table}

\textbf{Scenario: Knife slicing through butter.}

\begin{figure}[H]
    \centering
    \begin{subfigure}{1\linewidth}
    \includegraphics[width=1\linewidth]{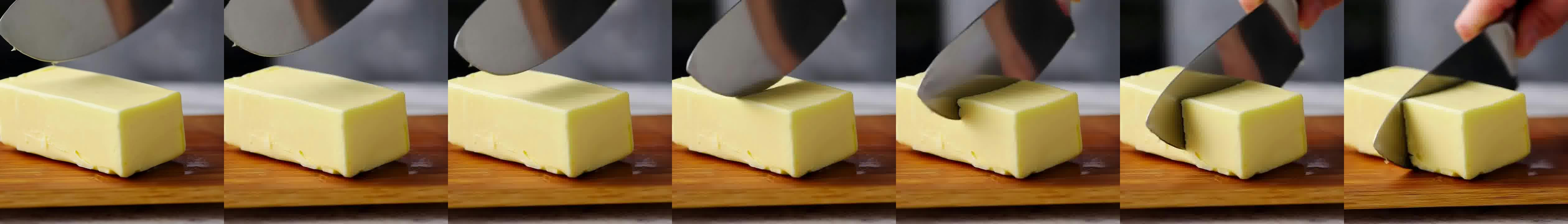}
    \caption{HunyuanVideo Generation}
    
    \end{subfigure}
    \begin{subfigure}{1\linewidth}
    \includegraphics[width=1\linewidth]{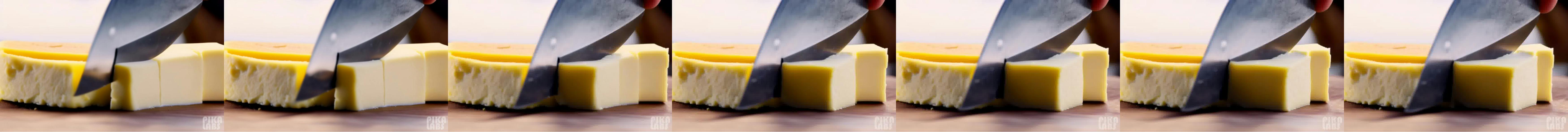}
    \caption{Pika Generation}

    \end{subfigure}
    \begin{subfigure}{1\linewidth}
    \includegraphics[width=1\linewidth]{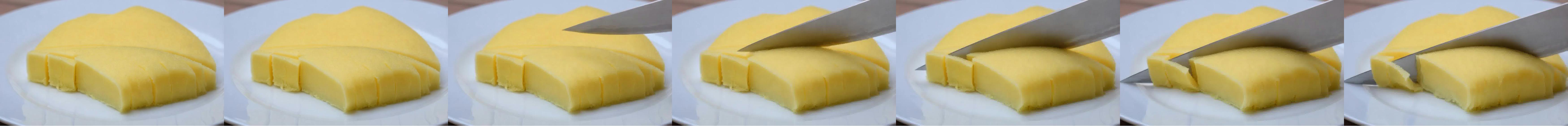}
    \caption{Hailuo Generation}

    \end{subfigure}
    \begin{subfigure}{1\linewidth}
    \includegraphics[width=1\linewidth]{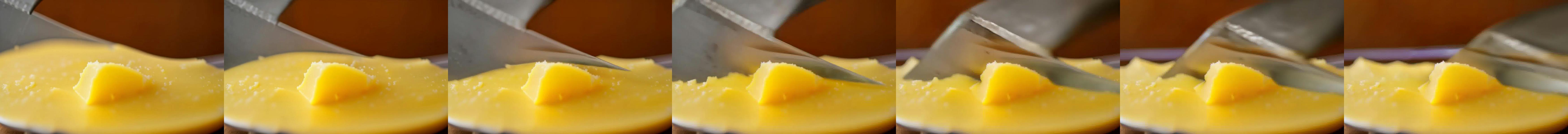}    
    \caption{Pyramid Generation}
    
    \end{subfigure}

    \caption{Model Generation}
    \label{fig:s3p1}
\end{figure}

Prompt-1:The knife meets little opposition as it slices through the butter. (Videos:Figure~\ref{fig:s3p1};Results: Tabel~\ref{tab:s3p1})

\begin{table}[H]
\centering
\begin{tabular}{c c c c}
\toprule
\multirow{2}{*}{Model} & \multirow{2}{*}{Factor} & Model & \multirow{2}{*}{Correct} \\ & & Answer & \\
\midrule
\multirow{3}{*}{Hunyuan} & blade in contact with butter & True & \textcolor{green}{\textbf{\checkmark}} \\ 
& Knife is moving against butter & True & \textcolor{green}{\textbf{\checkmark}} \\
& Butter is sliced & True & \textcolor{green}{\textbf{\checkmark}} \\
\midrule
\multirow{3}{*}{Pika} & blade in contact with butter & True & \textcolor{green}{\textbf{\checkmark}} \\ 
& Knife is moving against butter & True & \textcolor{green}{\textbf{\checkmark}} \\
& Butter is sliced & True & \textcolor{green}{\textbf{\checkmark}} \\
\midrule
\multirow{3}{*}{Hailuo} & blade in contact with butter & True & \textcolor{green}{\textbf{\checkmark}} \\ 
& Knife is moving against butter & True & \textcolor{green}{\textbf{\checkmark}} \\
& Butter is sliced & True & \textcolor{green}{\textbf{\checkmark}} \\
\midrule
\multirow{3}{*}{Pyramid} & blade in contact with butter & True & \textcolor{green}{\textbf{\checkmark}} \\ 
& Knife is moving against butter & True & \textcolor{green}{\textbf{\checkmark}} \\
& Butter is sliced & False & \textcolor{green}{\textbf{\checkmark}} \\

\bottomrule
\end{tabular}
\caption{Verification of VLLM Answer Correctness}
\label{tab:s3p1}
\end{table}

\begin{figure}[H]
    \centering
    \begin{subfigure}{1\linewidth}
    \includegraphics[width=1\linewidth]{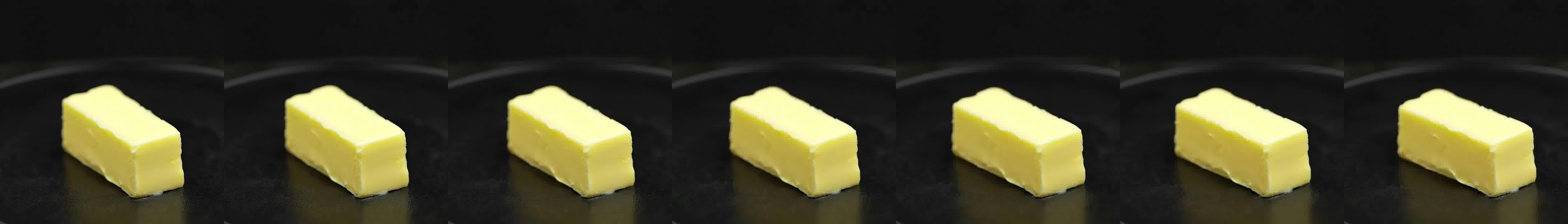}
    \caption{HunyuanVideo Generation}
    
    \end{subfigure}
    \begin{subfigure}{1\linewidth}
    \includegraphics[width=1\linewidth]{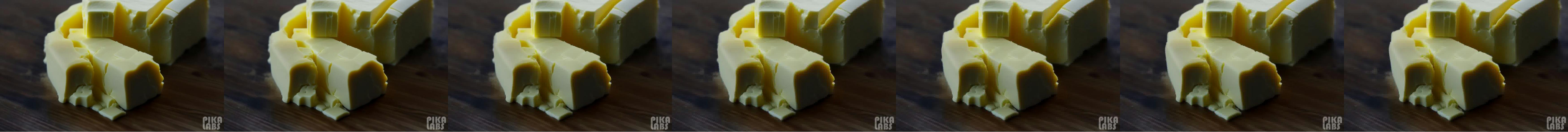}
    \caption{Pika Generation}

    \end{subfigure}
    \begin{subfigure}{1\linewidth}
    \includegraphics[width=1\linewidth]{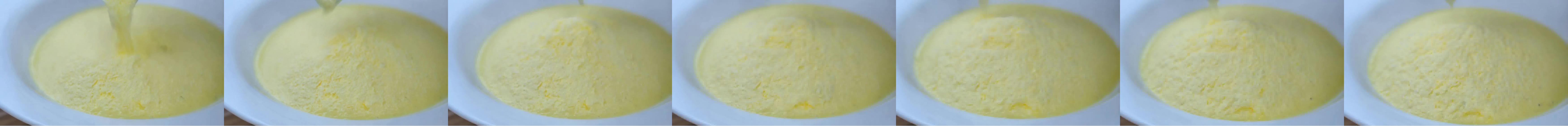}
    \caption{Hailuo Generation}

    \end{subfigure}
    \begin{subfigure}{1\linewidth}
    \includegraphics[width=1\linewidth]{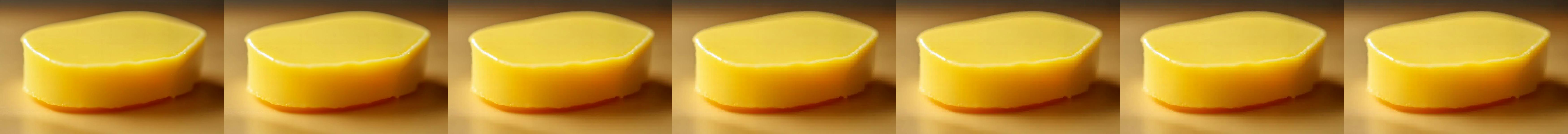}    
    \caption{Pyramid Generation}
    
    \end{subfigure}

    \caption{Model Generation}
    \label{fig:s3p2}
\end{figure}

Prompt-2:With no movement or contact, the butter sits undisturbed. (Videos:Figure~\ref{fig:s3p2};Results: Tabel~\ref{tab:s3p2})

\begin{table}[H]
\centering
\begin{tabular}{c c c c}
\toprule
\multirow{2}{*}{Model} & \multirow{2}{*}{Factor} & Model & \multirow{2}{*}{Correct} \\ & & Answer & \\
\midrule
\multirow{3}{*}{Hunyuan} & blade in contact with butter & False & \textcolor{green}{\textbf{\checkmark}} \\ 
& Knife is moving against butter & False & \textcolor{green}{\textbf{\checkmark}} \\
& Butter is sliced & False & \textcolor{green}{\textbf{\checkmark}} \\
\midrule
\multirow{3}{*}{Pika} & blade in contact with butter & False & \textcolor{green}{\textbf{\checkmark}} \\ 
& Knife is moving against butter & False & \textcolor{green}{\textbf{\checkmark}} \\
& Butter is sliced & False & \textcolor{green}{\textbf{\checkmark}} \\
\midrule
\multirow{3}{*}{Hailuo} & blade in contact with butter & False & \textcolor{green}{\textbf{\checkmark}} \\ 
& Knife is moving against butter & False & \textcolor{green}{\textbf{\checkmark}} \\
& Butter is sliced & False & \textcolor{green}{\textbf{\checkmark}} \\
\midrule
\multirow{3}{*}{Pyramid} & blade in contact with butter & False & \textcolor{green}{\textbf{\checkmark}} \\ 
& Knife is moving against butter & False & \textcolor{green}{\textbf{\checkmark}} \\
& Butter is sliced & False & \textcolor{green}{\textbf{\checkmark}} \\

\bottomrule
\end{tabular}
\caption{Verification of VLLM Answer Correctness}
\label{tab:s3p2}
\end{table}

\begin{figure}[H]
    \centering
    \begin{subfigure}{1\linewidth}
    \includegraphics[width=1\linewidth]{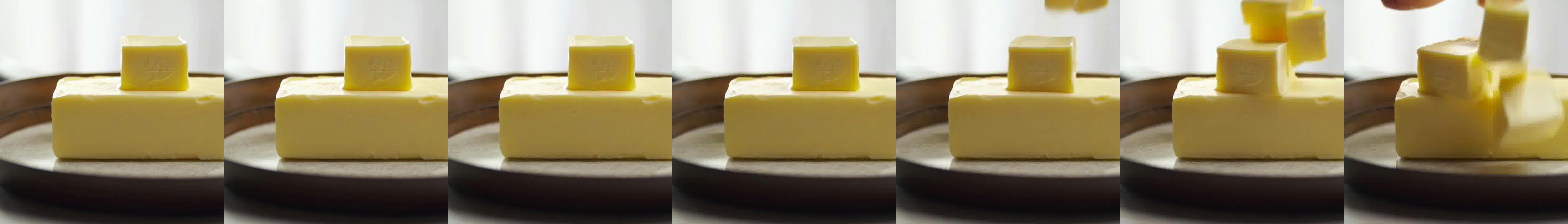}
    \caption{HunyuanVideo Generation}
    
    \end{subfigure}
    \begin{subfigure}{1\linewidth}
    \includegraphics[width=1\linewidth]{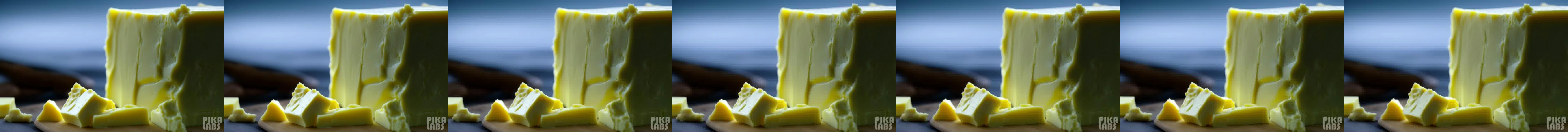}
    \caption{Pika Generation}

    \end{subfigure}
    \begin{subfigure}{1\linewidth}
    \includegraphics[width=1\linewidth]{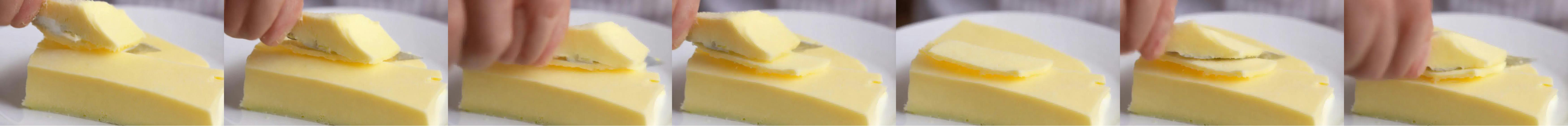}
    \caption{Hailuo Generation}

    \end{subfigure}
    \begin{subfigure}{1\linewidth}
    \includegraphics[width=1\linewidth]{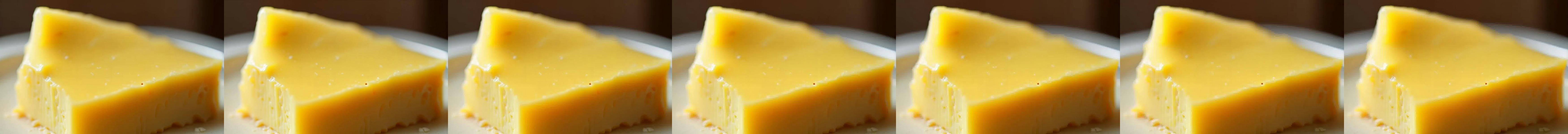}    
    \caption{Pyramid Generation}
    
    \end{subfigure}

    \caption{Model Generation}
    \label{fig:s3p3}
\end{figure}

Prompt-3:Contact with the butter is established, but without motion, the butter remains unsliced. (Videos:Figure~\ref{fig:s3p3};Results: Tabel~\ref{tab:s3p3})

\begin{table}[H]
\centering
\begin{tabular}{c c c c}
\toprule
\multirow{2}{*}{Model} & \multirow{2}{*}{Factor} & Model & \multirow{2}{*}{Correct} \\ & & Answer & \\
\midrule
\multirow{3}{*}{Hunyuan} & blade in contact with butter & False & \textcolor{green}{\textbf{\checkmark}} \\ 
& Knife is moving against butter & False & \textcolor{green}{\textbf{\checkmark}} \\
& Butter is sliced & False & \textcolor{green}{\textbf{\checkmark}} \\
\midrule
\multirow{3}{*}{Pika} & blade in contact with butter & False & \textcolor{green}{\textbf{\checkmark}} \\ 
& Knife is moving against butter & False & \textcolor{green}{\textbf{\checkmark}} \\
& Butter is sliced & False & \textcolor{green}{\textbf{\checkmark}} \\
\midrule
\multirow{3}{*}{Hailuo} & blade in contact with butter & True & \textcolor{green}{\textbf{\checkmark}} \\ 
& Knife is moving against butter & True & \textcolor{green}{\textbf{\checkmark}} \\
& Butter is sliced & True & \textcolor{green}{\textbf{\checkmark}} \\
\midrule
\multirow{3}{*}{Pyramid} & blade in contact with butter & False & \textcolor{green}{\textbf{\checkmark}} \\ 
& Knife is moving against butter & False & \textcolor{green}{\textbf{\checkmark}} \\
& Butter is sliced & False & \textcolor{green}{\textbf{\checkmark}} \\

\bottomrule
\end{tabular}
\caption{Verification of VLLM Answer Correctness}
\label{tab:s3p3}
\end{table}
\section{Details and more discussion about benchmarks}

\subsection{Evaluated Models}
\label{app:model_details}
To conduct a comprehensive benchmark, we evaluate a total of 6 open-source models and 4 closed-source models. Detailed information about the models included in our evaluation is provided in this section.

\textbf{Open-Source Models:}

For the open-source models, we benchmark 
\begin{itemize}
    \item CogVideoX~\citep{hong2022cogvideolargescalepretrainingtexttovideo}, a recent state-of-the-art video generation model. Specifically, we use three versions in our experiment: CogVideoX1.5-5B, CogVideoX-5B, and CogVideoX-2B;
    \item VideoCrafter2~\citep{chen2024videocrafter2}, the latest version of the VideoCrafter series, which is an open-source toolbox for video generation and editing;
    \item Pyramid Flow miniFLUX~\citep{jin2024pyramidalflowmatchingefficient}, utilizing its 768p checkpoint. This variant of the Pyramid Flow series supports the generation of both high-quality images and videos;
    \item HunyuanVideo~\citep{kong2025hunyuanvideosystematicframeworklarge}, developed by Tencent. HunyuanVideo is currently the largest open-source video generation model, with over 13 billion parameters, and provides performance comparable to leading closed-source models.
\end{itemize} 

All of the open-source models used in our experiments were downloaded from the Huggingface website.

%\textbf{CogVideoX~\citep{hong2022cogvideolargescalepretrainingtexttovideo}}: CogVideoX is a series of open-source video generation models developed by QingYing. In our experiment, we use three versions of CogVideoX: CogVideoX1.5-5B, CogVideoX-5B, and CogVideoX-2B, all of which were downloaded from the Huggingface website.

%\textbf{VideoCrafter~\citep{chen2024videocrafter2}}: VideoCrafter is an open-source video generation and editing toolbox for crafting video content. In our experiments, we use the latest version of its T2V model, VideoCrafter2, which was downloaded from the Huggingface website.

%\textbf{Pyramid Flow~\citep{jin2024pyramidalflowmatchingefficient}}: Pyramid Flow is an autoregressive video generation method that prioritizes training efficiency and is based on flow matching. In our experiments, we utilize the 768p checkpoint of the miniFLUX variant, which was sourced from the Huggingface website.

%\textbf{HunyuanVideo~\citep{kong2025hunyuanvideosystematicframeworklarge}}: HunyuanVideo, developed by Tencent, is currently the largest open-source video generation model, with over 13 billion parameters, and delivers performance comparable to leading closed-source models. It has been integrated into Diffusers, and we obtained the pretrained model from the Huggingface website for our experiments.

\textbf{Close-Source Models:}

For the close-source models, we benchmark 

\begin{itemize}
    \item Gen-3 Alpha~\citep{runwaygen3}, The latest version released by Runway shows improvements in fidelity, consistency, and motion compared to Gen-2;
    \item Pika~\citep{Pika}, developed by Pika Labs, is used in its free beta version, accessed through the Pika Discord Bot;
    \item  Hailuo~\citep{Hailuo}, developed by MiniMax, is used in its T2V-01 version;
    \item Kling 1.0~\citep{Kling}, a closed VGM released by Kuaishou.
\end{itemize} 

 We access all the closed-source models by calling their APIs, either through their official websites or third-party interfaces. Detailed information can be found in ~\ref{app:cost}. Some of the models provide an additional prompt enhancement trick but for fair comparison, we do not turn it on if there is an option. See discussion about this trick in Appendix~\ref{app:prompt_enhance}.

\subsection{Cost of benchmarking}
\label{app:cost}
We report the time and money cost of benchmarking each model here.

Open-Source Models: see Table~\ref{tab: open-source models}.

\begin{table*}[htbp]
\caption{The time and money cost for open-source models.}
\label{tab: open-source models}
\centering
\begin{tabular}{c c c c}
\toprule
Name & Device & Time / Video & Total Time (above 2000 videos)\\
\midrule
CogVideoX1.5-5B & NVIDIA A800-SXM4-80GB & $ \sim $ 15min & $ \sim $ 500 GPU hours \\
CogVideoX-5B & NVIDIA A800-SXM4-80GB & $ \sim $ 3min & $ \sim $ 100 GPU hours \\
CogVideoX-2B & NVIDIA A800-SXM4-80GB & $ \sim $ 1min & $ \sim $ 33 GPU hours \\
Pyramid Flow & NVIDIA A800-SXM4-80GB & $ \sim $ 2.5min & $ \sim $ 83 GPU hours \\
HunyuanVideo & NVIDIA A800-SXM4-80GB & $ \sim $ 10min & $ \sim $ 330 GPU hours \\
VideoCrafter2 & NVIDIA A800-SXM4-80GB & $ \sim $ 3min & $ \sim $ 100 GPU hours \\
\bottomrule
\end{tabular}
\end{table*}

Close-Source Models: see Table~\ref{table: closed-source models}.

\begin{table*}[htbp]
\caption{The time and money cost for close-source models.}
\label{table: closed-source models}
\centering
\begin{tabular}{c c c c}
\toprule
Name & API Source & Cost / Video & Total Cost (above 2000 videos)\\
\midrule
Gen-3 Alpha & Useapi.net & Unlimited Subscription & \$ 95 \\
Pika & Useapi.net & Pika Discord Bot & Free \\
Kling & PiAPI & \$ 0.13 & \$ 260 \\
Hailuo & Official & Unlimited Subscription & \$ 94.99 \\
\bottomrule
\end{tabular}
\end{table*}

\subsection{About N/A results}
\label{app:affect_of_na}
When we retrieve the observed values in a video by a VLLM, we allow the model to answer `N/A' besides yes or no. We prompt the model the conditions of answering N/A as follows:
\begin{promptbox}
1. The video quality is too low, or the content is too unclear to make any meaningful inference.

2. The content in the video is not continuous or complete. The temporal and spatial discontinuities in the video make it impossible to make reasonable predictions.

3. The question asks about something that cannot be observed or recognized in the video (e.g., an object, event, or action that is not present).

4. The video does not provide enough context or evidence to form a conclusion.

5. The answer is unclear or could be interpreted in multiple ways, leading to ambiguity.

6. The question asks about an action, and the necessary prior action (for example, the ball hitting the ground before it can bounce) is not observed. Without the prior action, it is impossible to determine if the subsequent event occurred. 

\end{promptbox}

We report the N/A ratio in all observation in Table~\ref{tab:main_benchmark} and we also report the `N/A : correct : incorrect' ratio for all test we used in Level 1 all $s_1^{all}$ in Table~\ref{tab:nan_ratio}. 

We acknowledge that the appearance of N/A may introduce some bias to subsequent metrics. 
%For example, assuming that the model generates N/A in scenarios that it is less good at, removing N/A from the results may help the model avoid scenarios that it is not good at, thereby making the score higher than its actual performance, or the performance gap between different models is erroneously narrowed. 
For example, if the model generates N/A in scenarios where it performs poorly, removing these N/A responses could lead to inflated scores. This would make the model appear better than it actually is, or falsely narrow the performance gap between different models. 
But as we mentioned in the introduction (Section~\ref{sec:intro}), as a longer-term goal, our evaluation focuses more on the evaluation of the ``world simulator'', and the guarantee of general video generation quality should be taken as a prerequisite rather than the focus of this article. At the same time, we observe that better (newer, larger) models tend to have a lower N/A ratio, which is in line with our expectations and  shows that as the model generation capability continues to improve, the probability of obvious serious errors will gradually decrease.

\begin{table}[htbp]
\centering
\caption{The ratio of N/A variables, correct variables and incorrect variables for text consistency.}
\label{tab:nan_ratio}
\begin{tabular}{lccc}
\toprule
\textbf{Name} & \textbf{N/A ratio} & \textbf{correct ratio} & \textbf{incorrect ratio} \\
\midrule
CogVideoX1.5-5B  & .06 & .53 & .41 \\
CogVideoX-5B  & .06 & .55 & .39 \\
CogVideoX-2B  & .08 & .52 & .40 \\
VideoCrafter2  & .14 & .48 & .39 \\
Pyramid Flow  & .10 & .51 & .39 \\
HunyuanVideo  & .07 & .55 & .39 \\
\midrule
Pika  & .11 & .52 & .38 \\
Hailuo  & .07 & .55 & .38 \\
Gen-3 Alpha & .07 & .60 & .33 \\
Kling & .07 & .58 & .35 \\
\bottomrule
\end{tabular}
\end{table}

\subsection{Experiment for sample size}
\label{app:sample_size_exp}
\begin{figure*}
    \centering
    \begin{subfigure}{0.48\linewidth}
    \includegraphics[width=0.98\linewidth]{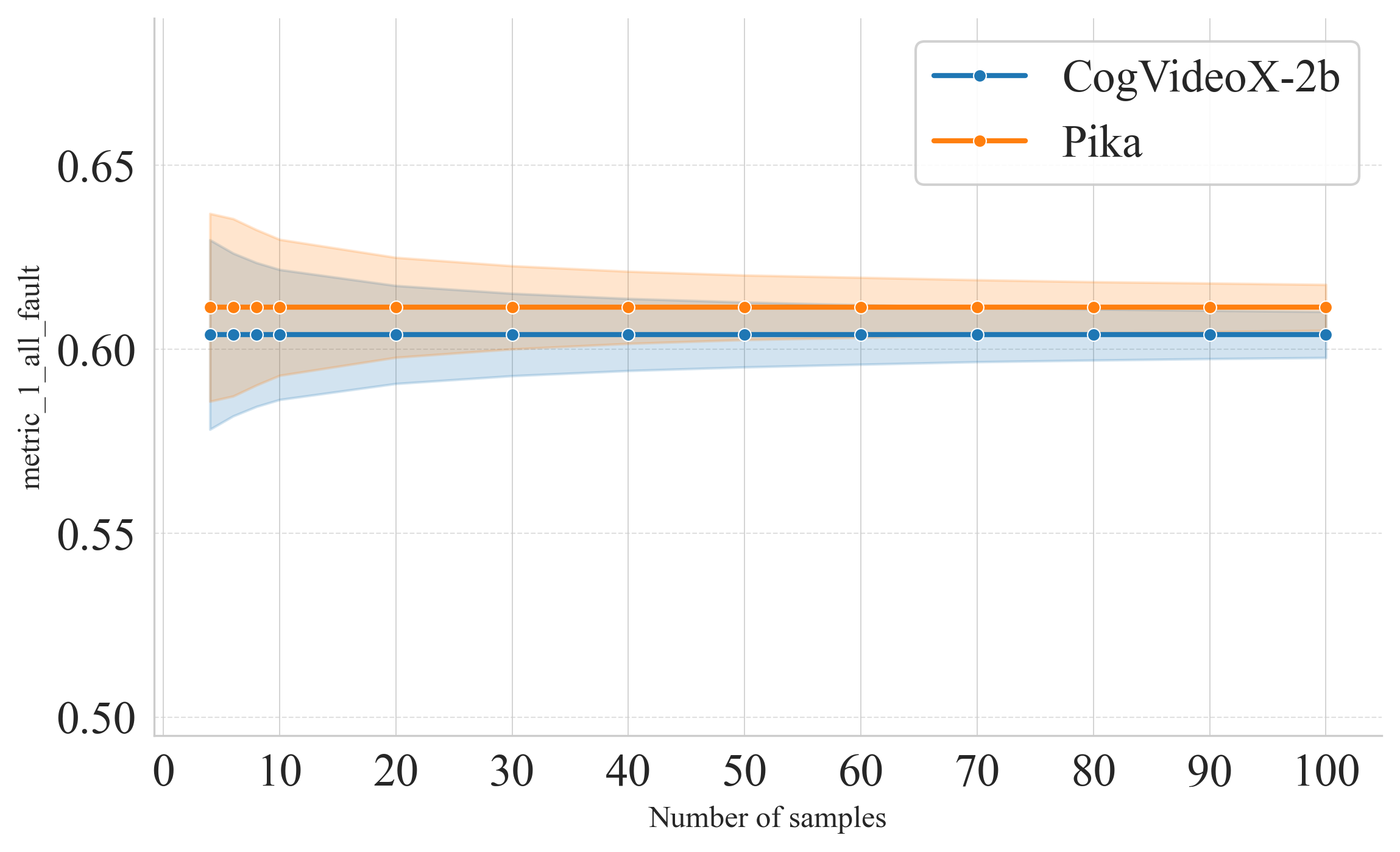}
    \caption{$s_1^{\mathrm{all}}$ with N/A as incorrect}
    
    \end{subfigure}
    \begin{subfigure}{0.48\linewidth}
    \includegraphics[width=0.98\linewidth]{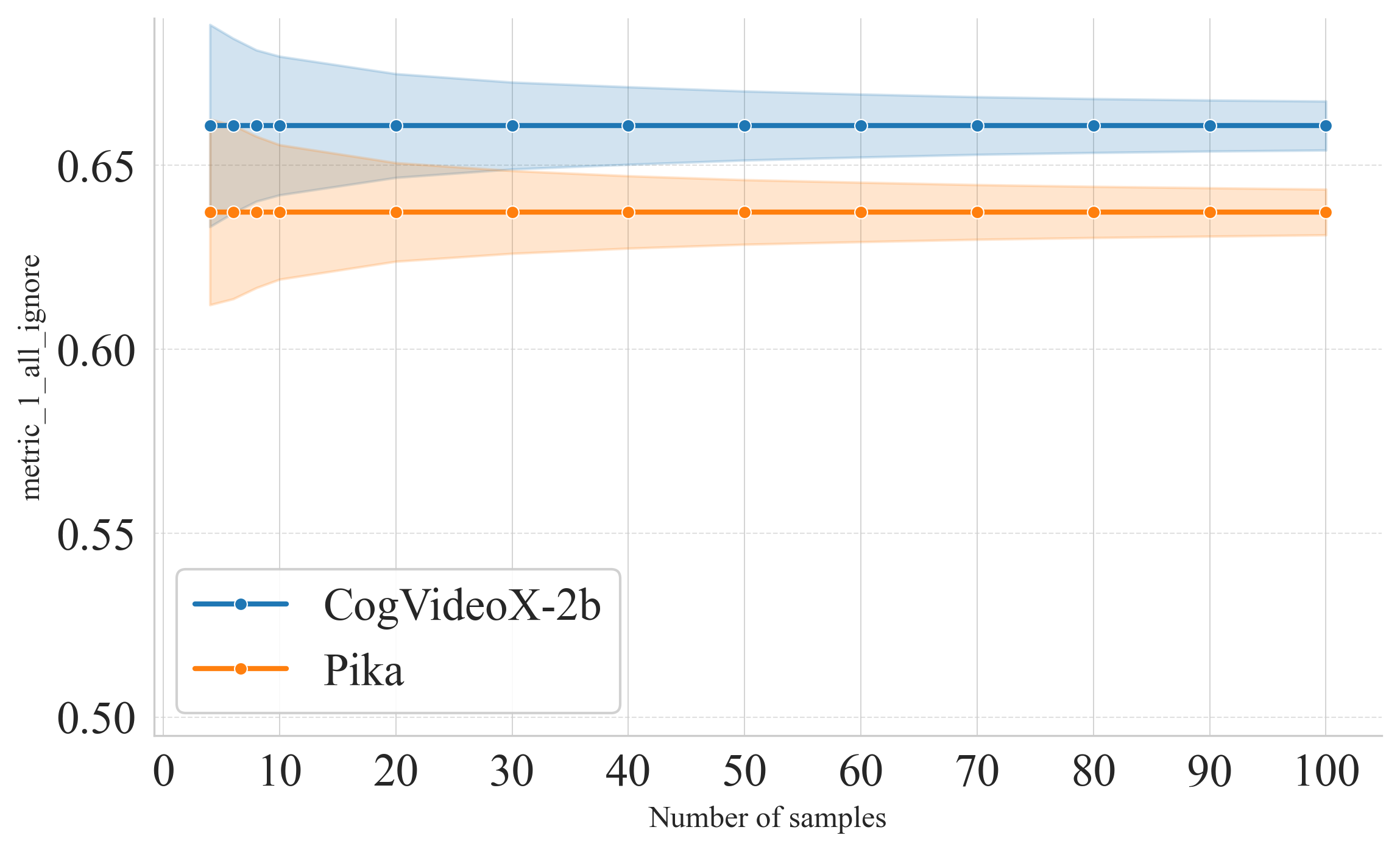}
    \caption{$s_1^{\mathrm{all}}$ with N/A ignored}
    
    \end{subfigure}

    \begin{subfigure}{0.48\linewidth}
    \includegraphics[width=0.98\linewidth]{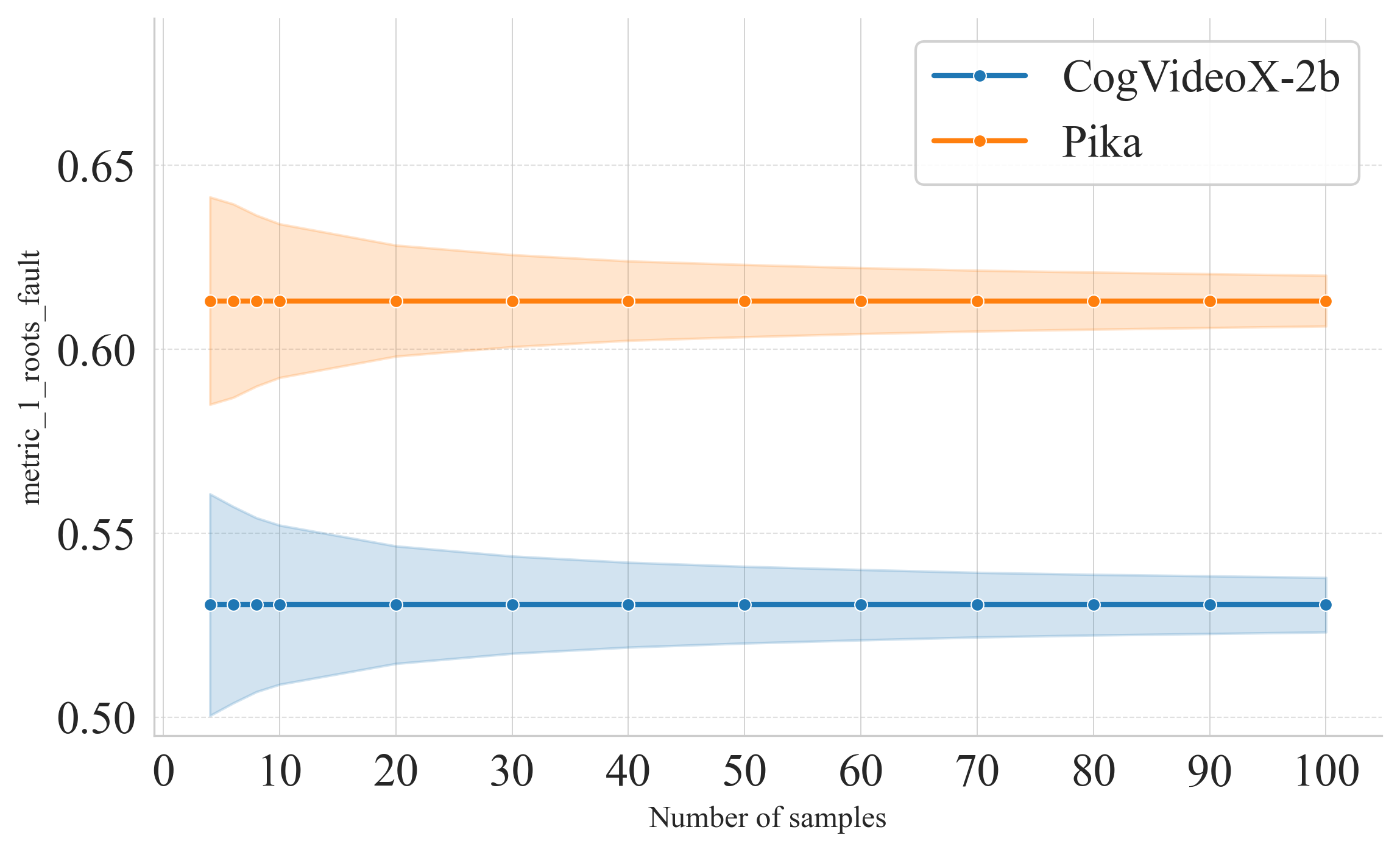}
    \caption{$s_1^{\mathrm{roots}}$ with N/A as incorrect}
    \end{subfigure}
    \begin{subfigure}{0.48\linewidth}
    \includegraphics[width=0.98\linewidth]{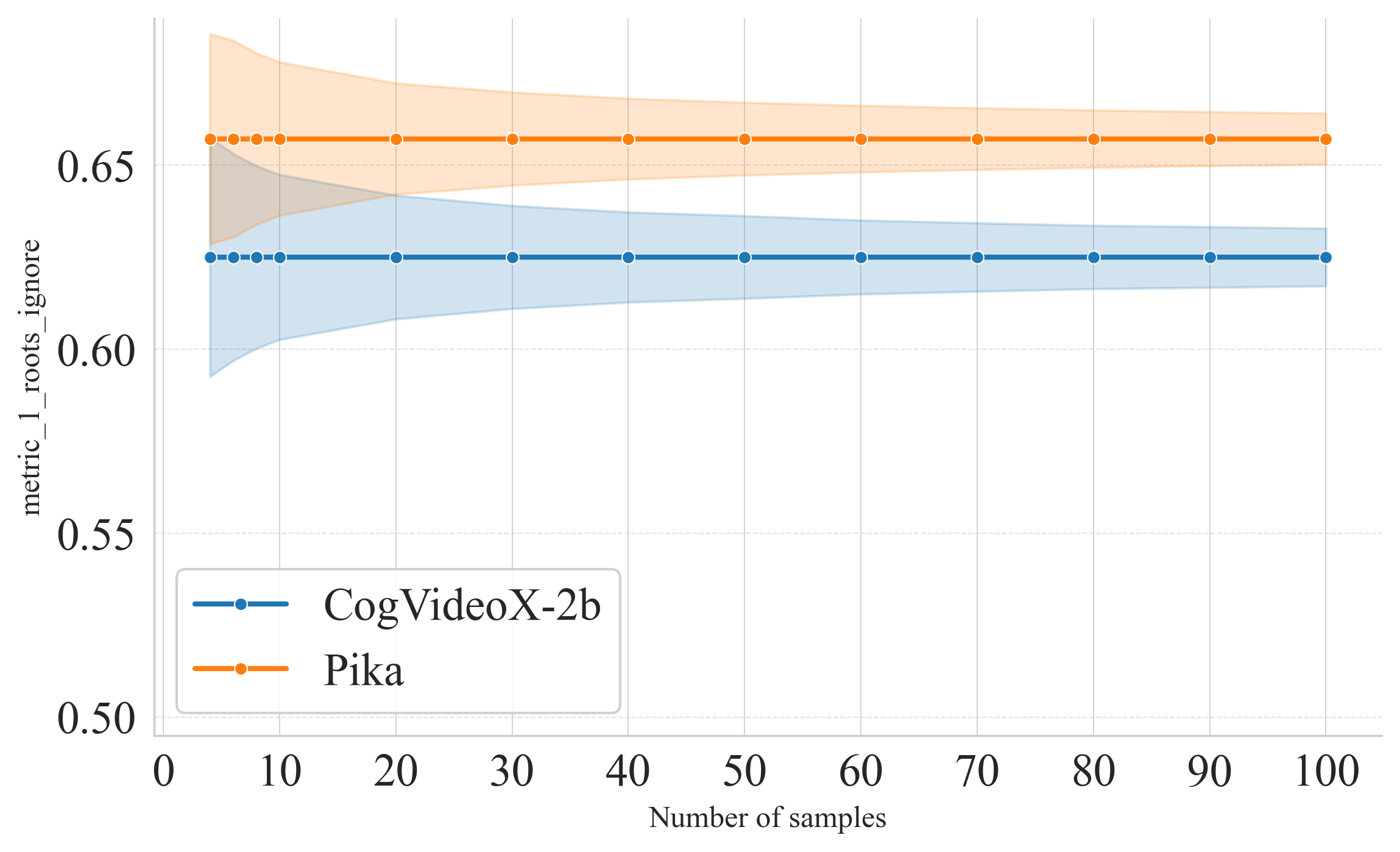}
    \caption{$s_1^{\mathrm{roots}}$ with N/A ignored}
    \end{subfigure}

    \begin{subfigure}{0.48\linewidth}
    \includegraphics[width=0.98\linewidth]{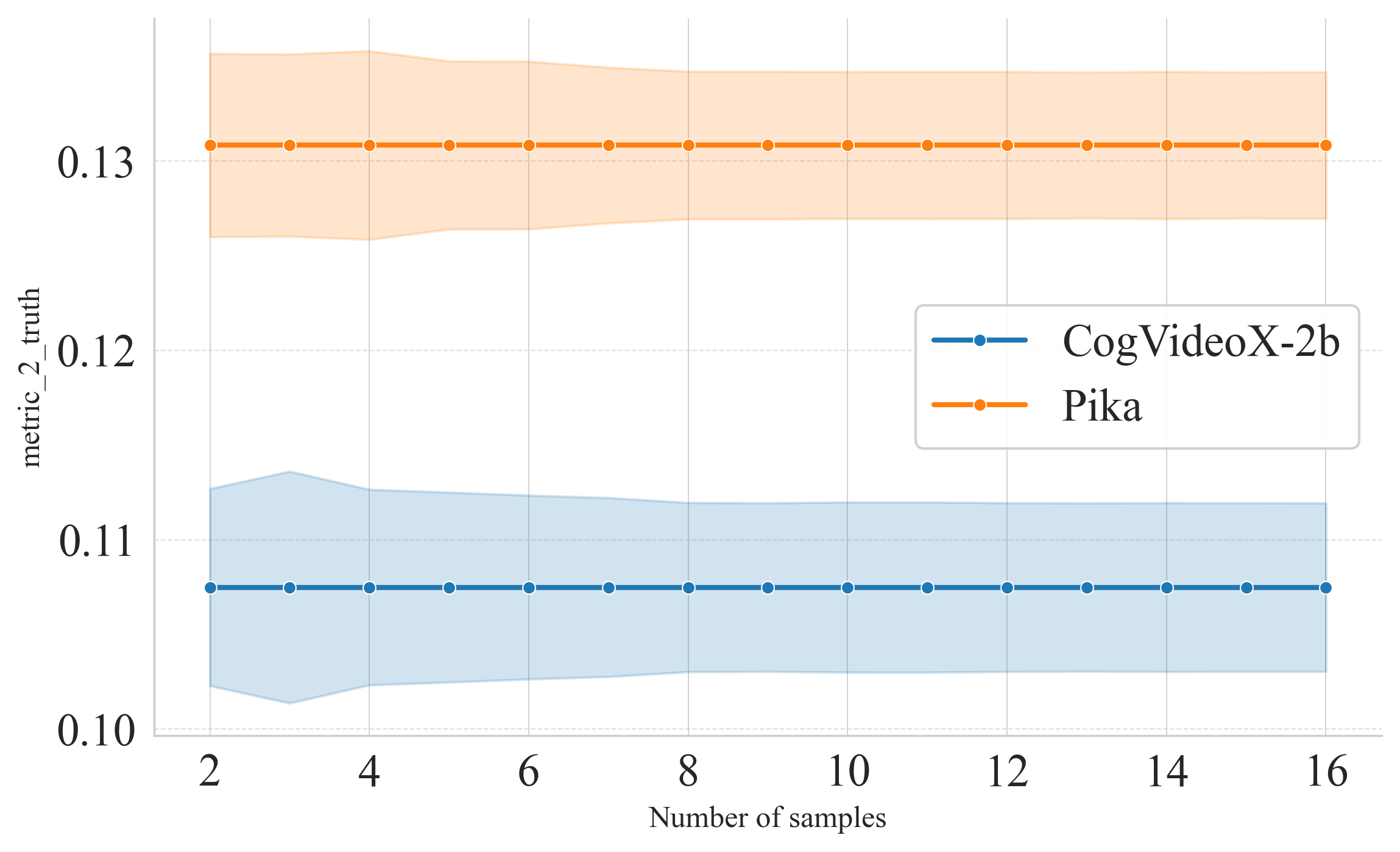}
    \caption{$s_2^{\mathrm{truth}}$}
    \end{subfigure}
    \begin{subfigure}{0.48\linewidth}
    \includegraphics[width=0.98\linewidth]{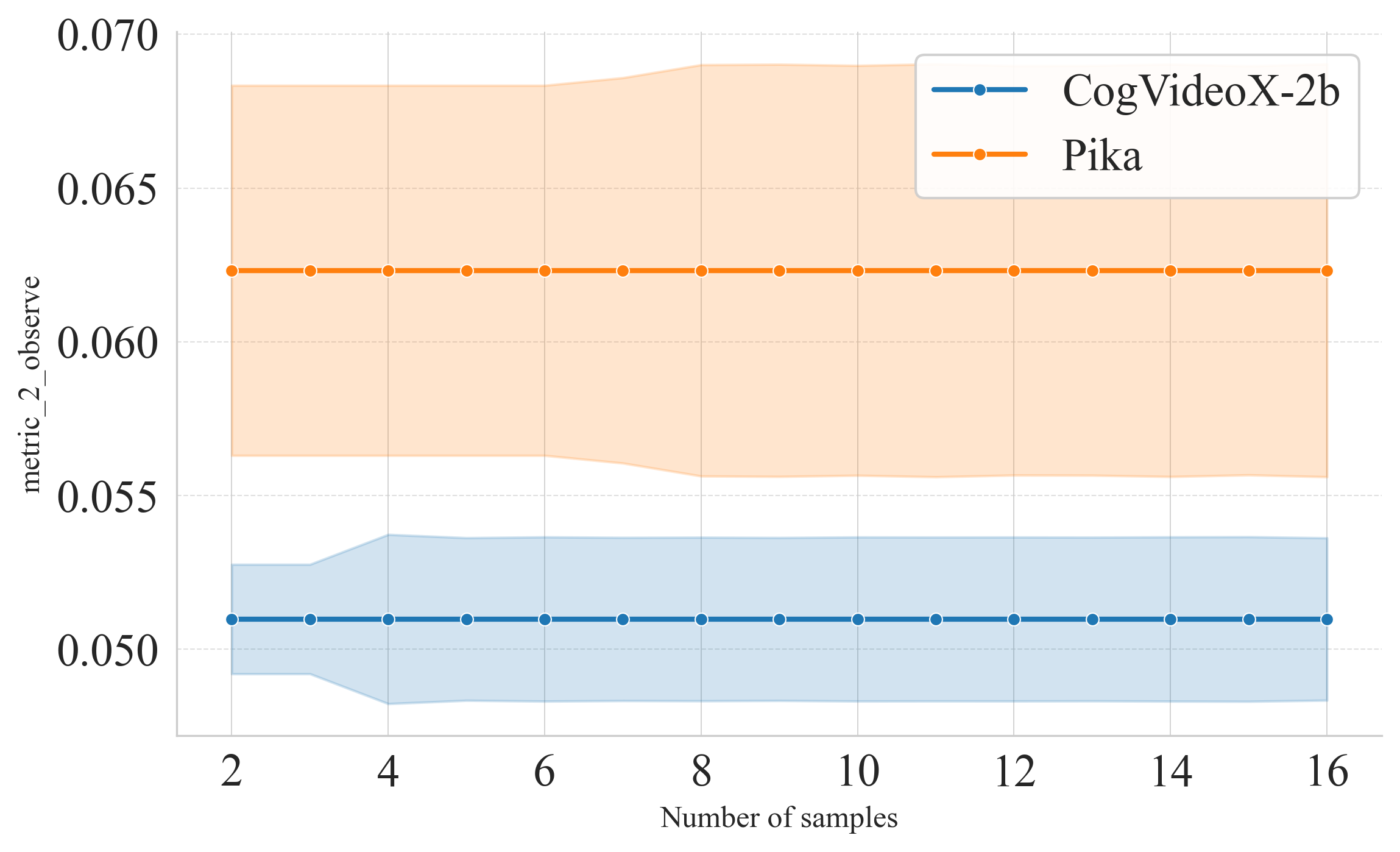}
    \caption{$s_2^{\mathrm{observe}}$}
    \end{subfigure}

    \begin{subfigure}{0.48\linewidth}
    \includegraphics[width=0.98\linewidth]{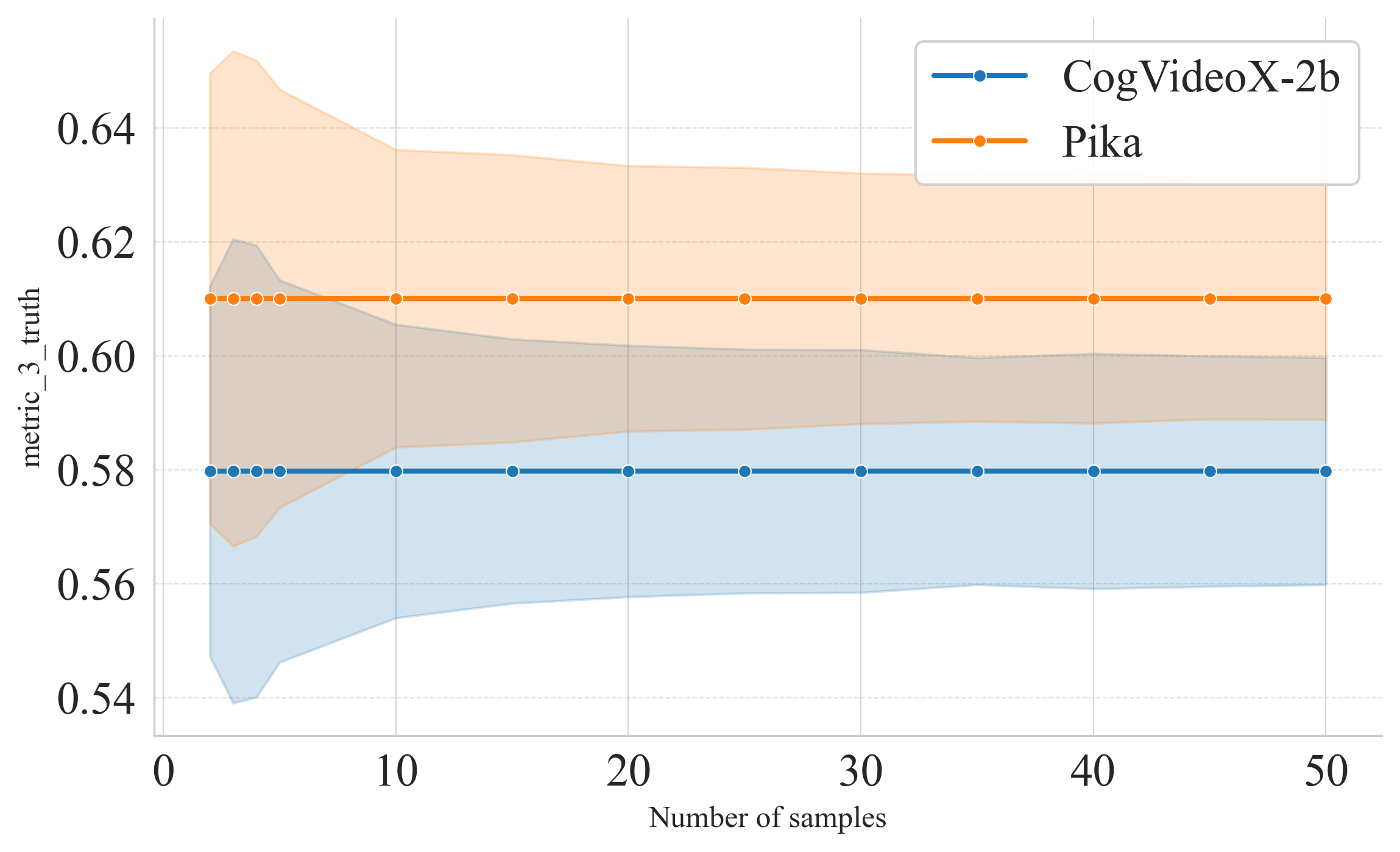}
    \caption{$s_3^{\mathrm{truth}}$}
    \end{subfigure}
    \begin{subfigure}{0.48\linewidth}
    \includegraphics[width=0.98\linewidth]{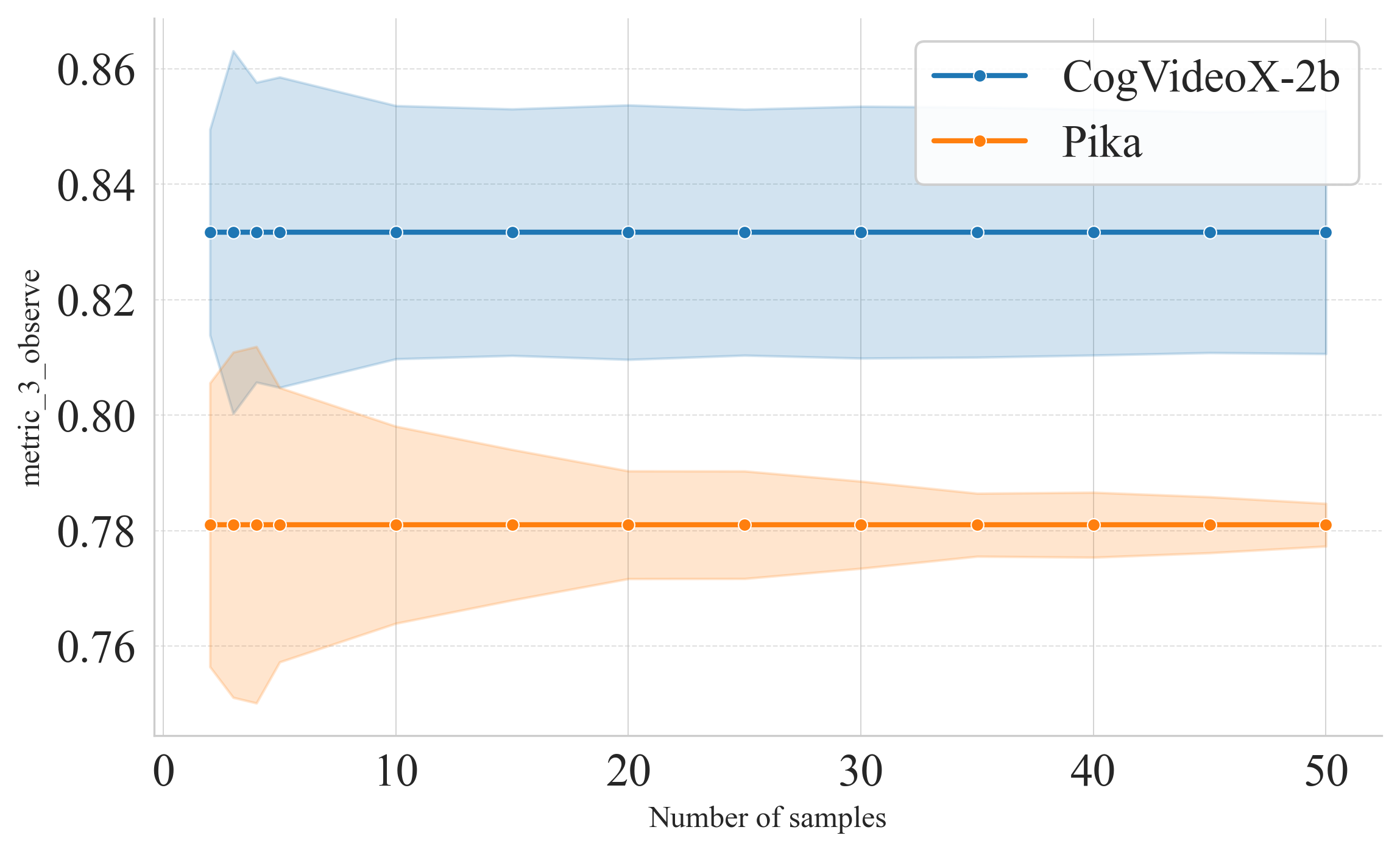}
    \caption{$s_3^{\mathrm{observe}}$}
    \end{subfigure}
    \caption{Estimated confidence interval for each metric as the sample size increases. }
    \label{fig:sample_exp}
\end{figure*}

We conduct an empirical study to determine the minimum sample size required for statistically distinguishing performance metrics between two video generation models (VGMs). The experiment compares CogVideoX-2B (representing open-source models) and Pika (representing closed-source models) under a specific causal system where both models exhibited competent video generation quality. We vary sample sizes from $2$ to $100$ for text consistency, group sizes from $2$ to $16$ for generation consistency, and sample sizes for each outcome variable from $2$ to $50$ samples for rule consistency. To ensure statistical validity, we employ bootstrap resampling (1,000 iterations) with finite-population correction to estimate standard deviations of metric estimators. Standard deviations are adjusted for matching our scenario pool ($60$ causal systems). For text consistency metrics, we implement two evaluation protocols: 1) excluding missing (N/A) observations, and 2) treating N/A values as incorrect responses. Confidence intervals (95\% coverage) are constructed using bias-corrected accelerated bootstrap methods centered on the minimum-variance unbiased estimator.

\begin{figure}[htpb]
      \centering
      \includegraphics[width=0.65\linewidth]{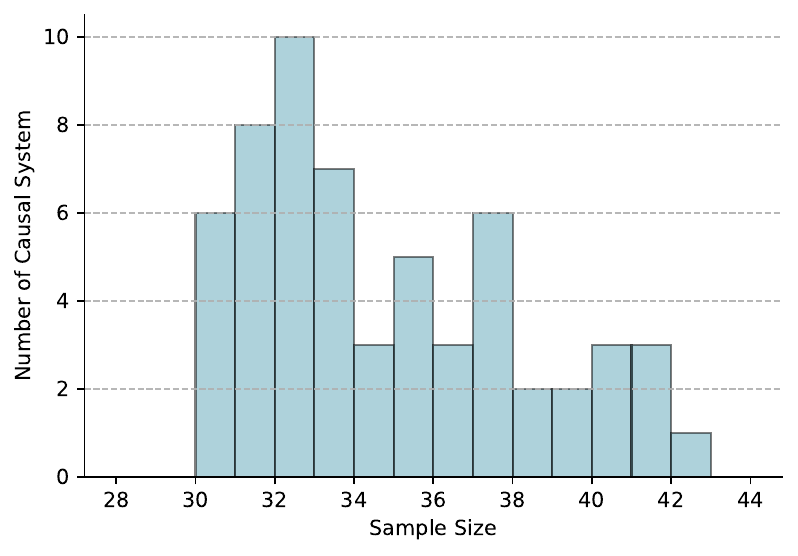}
      \caption{Sample numbers of the 60 causal systems in VACT benchmark.}
      \label{fig:sample_number}
  \end{figure}

The results, visualized in Figure~\ref{fig:sample_exp}, reveal distinct sample size requirements across metrics. As a efficiency-accuracy trade-off, we established an operational criterion where the minimal sufficient sample size occurs when the confidence interval of one model's metric no longer overlaps with the point estimate of the competitor model. From the figure we can see that:
\begin{itemize}
    \item For text consistency, drawing $n_1=10$ samples is enough to distinguish metrics between two models in most cases. When N/A observed variables are seen as incorrect, $s_1^{\mathrm{all}}$ between two models cannot be distinguished for any number of samples. 
    \item For generation consistency, drawing $n_2=5$ groups can distinguish metrics between two models.
    \item For rule consistency, drawing $n_3=10$ samples for each outcome variable can distinguish metrics between two models.
\end{itemize}
  Based on these findings, our benchmark protocol adopts $n_1=10$, $n_2=5$, and $n_3=10$ as optimal parameters balancing statistical power and evaluation efficiency, leading to total \num{2079} video samples. The sample numbers of these 60 causal systems are shown in Figure~\ref{fig:sample_number}.

\subsection{Human-sourced benchmarking}
\label{app:human-source}
To validate the effectiveness of automatically generated causal systems, we manually annotated an additional $60$ causal systems for these 20 scenarios through crowd experiments under identical instructions. For these human-annotated causal systems, we conducted experiments using three video generation models: CogVideoX1.5-5B, Hailuo, and Pika. The metric results are presented in Table~\ref{tab:human_benchmark_app}, with missing value (N/A) cases analogous to Appendix~\ref{app:affect_of_na} shown in Table~\ref{tab:human_nan_ratio}, and threshold sensitivity experiments summarized in Table~\ref{tab:human_metric_level_3_threshold}.

\begin{table*}[htbp]
\centering

\caption{VACT benchmark on prevailing VGMs on human-sourced causal systems.}

\label{tab:human_benchmark_app}
\small
\begin{tabular}{lc *{3}{cc}} 
\toprule
\multirow{2}{*}{Model Names} & \multirow{2}{*}{N/A ratio} & \multicolumn{2}{c}{Text Consistency $\uparrow$} & \multicolumn{2}{c}{Generation Consistency $\downarrow$} & \multicolumn{2}{c}{Rule Consistency $\uparrow$} \\
\cmidrule(lr){3-4} \cmidrule(lr){5-6} \cmidrule(lr){7-8}  
& & all & root & truth & observe & truth & observe \\
 \midrule
CogVideoX1.5-5B & .11 & $.58\mathsmaller{\pm .01}$ & $.58\mathsmaller{\pm .02}$ & $.09\mathsmaller{\pm .01}$ & $.08\mathsmaller{\pm .01}$ & $.54\mathsmaller{\pm .02}$ & $.69\mathsmaller{\pm .02}$\\
\midrule
Pika & .18 & $.57\mathsmaller{\pm .01}$ & $.55\mathsmaller{\pm .02}$ & $.07\mathsmaller{\pm .01}$ & $.06\mathsmaller{\pm .01}$ & $.54\mathsmaller{\pm .02}$ & $.67\mathsmaller{\pm .02}$ \\
Hailuo & .14 & $.63\mathsmaller{\pm .01}$ & $.62\mathsmaller{\pm .02}$ & $.07\mathsmaller{\pm .01}$ & $.08\mathsmaller{\pm .01}$ & $.55\mathsmaller{\pm .01}$ & $.70\mathsmaller{\pm .02}$ \\
\bottomrule
\end{tabular}
\end{table*}

The results demonstrate that all metric scores derived from human-annotated causal systems closely align with those obtained from automated causal systems. This indicates that the automatically generated causal systems effectively capture scenario-specific features and critical variables while establishing valid rules. Notably, the N/A ratio in observational data increased across all models compared to results from automated causal systems. Concurrently, model performance on rule consistency metrics exhibited degradation. These observations suggest that video generation models face slightly bigger challenges in interpreting human-annotated causal systems, likely due to increased complexity and ambiguity in manually defined causal relationships.

\begin{table}[htbp]
\centering
\caption{The ratio of N/A variables, correct variables and incorrect variables for text consistency on human-sourced causal systems.}
\label{tab:human_nan_ratio}
\small
\begin{tabular}{lccc}
\toprule
\textbf{Name} & \textbf{N/A ratio} & \textbf{correct ratio} & \textbf{incorrect ratio} \\
\midrule
CogVideoX1.5-5B  & .12 & .51 & .37 \\
\midrule
Pika  & .17 & .48 & .35 \\
Hailuo  & .13 & .54 & .33 \\
\bottomrule
\end{tabular}
\end{table}

\begin{table*}[htbp]
\centering
\caption{Metrics for rule consistency on human-sourced causal systems by applying threshold for each rule.}
\label{tab:human_metric_level_3_threshold}
\begin{tabular}{l *{8}{c}}  
\toprule
\multirow{2}{*}{\textbf{Name}} & 
\multicolumn{4}{c}{$s_3^{\mathrm{truth, threshold}}$} & 
\multicolumn{4}{c}{$s_3^{\mathrm{observe, threshold}}$} \\
\cmidrule(lr){2-5} \cmidrule(lr){6-9}
& 0.65 & 0.75 & 0.85 & 0.95 & 0.65 & 0.75 & 0.85 & 0.95 \\
\midrule
CogVideoX1.5-5B  & $.23 \mathsmaller{\pm .04}$ & $.14 \mathsmaller{\pm .03}$ & $.03 \mathsmaller{\pm .02}$ & $.03 \mathsmaller{\pm .02}$ & $.56 \mathsmaller{\pm .04}$ & $.47 \mathsmaller{\pm .05}$ & $.29 \mathsmaller{\pm .04}$ & $.11 \mathsmaller{\pm .03}$ \\
\midrule
Pika  & $.19 \mathsmaller{\pm .04}$ & $.09 \mathsmaller{\pm .03}$ & $.05 \mathsmaller{\pm .02}$ & $.05 \mathsmaller{\pm .02}$ & $.45 \mathsmaller{\pm .04}$ & $.38 \mathsmaller{\pm .05}$ & $.30 \mathsmaller{\pm .04}$ & $.20 \mathsmaller{\pm .04}$ \\
Hailuo & $.22 \mathsmaller{\pm .04}$ & $.12 \mathsmaller{\pm .03}$ & $.03 \mathsmaller{\pm .01}$ & $.01 \mathsmaller{\pm .01}$ & $.56 \mathsmaller{\pm .04}$ & $.42 \mathsmaller{\pm .04}$ & $.29 \mathsmaller{\pm .05}$ & $.16 \mathsmaller{\pm .04}$ \\
\bottomrule
\end{tabular}
\end{table*}

\section{Case study on benchmark results}
\label{app:example_analysis}
\subsection{About the ``degenerative'' rules}
\label{app:analyze_gen_stable}
Since our metric 2 only focus on the stability but not the correctness, we are worried that the lower (better, stabler) metric 2 combined with the poorer metric 1 and metric 3 (low accuracy) actually implies that the model learns shortcut on common scenario. In many cases, models ignore the changes in $\mathbf{X}$ but directly generate the most common results. We support our concern through some case studies.

In the scenario about ``A burning candle is placed with (wind and rain).'', a key outcome is whether the candle remains lit or is extinguished by these environmental factors. However, we found that most of the VGMs consistently generate a candle that continues to burn, without accounting for these influences. For Gen-3 Alpha, in three test cases of this scenario, the expected outcome---an extinguished candle---occurred 11, 10, and 10 times, respectively. However, the actual results were only 2, 0, and 3 instances where the candle was extinguished. This makes the ``candle extinguished'' result appear almost as a constant ``False''. Similar phenomenon can be found about the outcome ``whether the pencil mark has been removed'' in the scenario ``Rubber eraser rubs off (pencil) marks on paper.''. Similarly, the statement ``the water color is uniform'' is always false after ``Dropping dye into the water'' regardless of ``whether the water is stirred sufficiently''.

\subsection{About sample-based score}
\label{app:example_score_based_analysis}
Here we demonstrate how the sample-based scores provide a more detailed analysis of model behavior by an example. Taking the model CogVideoX1.5-5B and the scenario ``\textit{A hand squeezes a sponge.}'' as the example, one of the generated causal system is:

``hand squeeze sponge $\land$ sponge is wet $\to$ water is squeezed out''. 

\begin{figure*}[h!]
    \centering
    \begin{subfigure}{0.99\linewidth}
\includegraphics[width=1\linewidth]{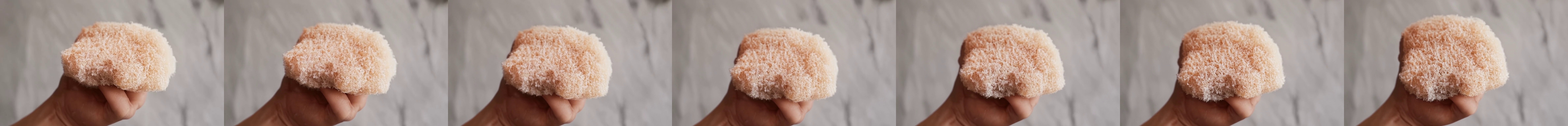}
    \caption{No squeeze, not wet, no water squeezed out.}
    \end{subfigure}

    \begin{subfigure}{0.99\linewidth}
\includegraphics[width=1\linewidth]{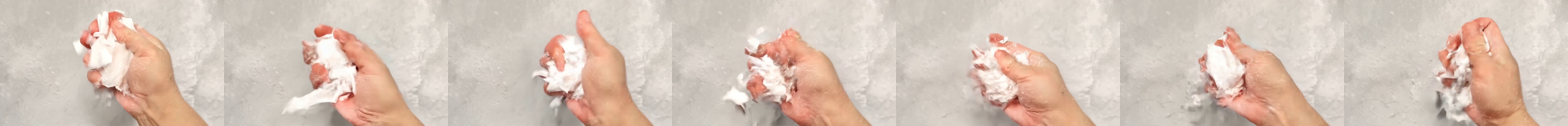}
    \caption{Squeeze, not wet, no water squeezed out.}
    \end{subfigure}

    \begin{subfigure}{0.99\linewidth}
\includegraphics[width=1\linewidth]{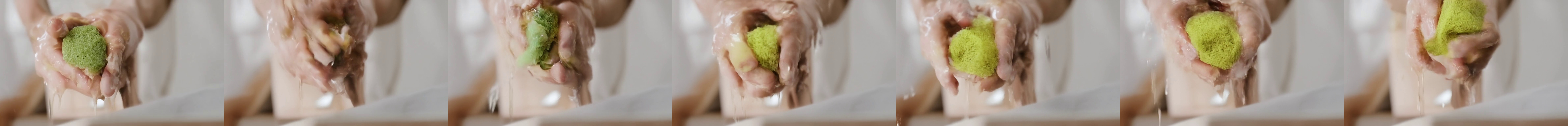}
    \caption{Squeeze, wet (deeper color in first several frames), water squeezed out.}
    \end{subfigure}
    
    \caption{Good examples with rule consistency score $1.0$.}
    \label{fig:sponge_example_good}
\end{figure*}

\begin{figure*}[h!]
    \centering
    \begin{subfigure}{0.99\linewidth}
    \includegraphics[width=1\linewidth]{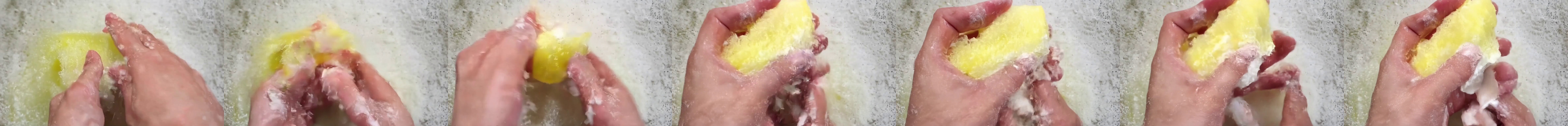}
    \caption{Squeeze, wet, no water squeezed out.}
    \end{subfigure}
    
    \begin{subfigure}{0.99\linewidth}
    \includegraphics[width=1\linewidth]{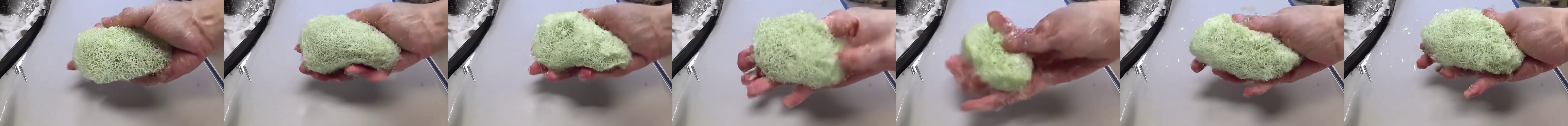}
    \caption{Squeeze, not wet, water squeezed out (water droplets appear in the last two images).}
    \end{subfigure}
    \caption{Bad examples with rule consistency score $0.0$.}
    \label{fig:sponge_example_bad}
\end{figure*}

By checking the scores of the generated videos, we observe that some videos have a metric 3 score (rule consistency) of $1.0$ (full score), indicating that these videos comply with all rules. We show these videos are shown in Figure~\ref{fig:sponge_example_good}, corresponding to some successful generation. As comparison, some of generation have much lower metric 3 score and are shown in Figure~\ref{fig:sponge_example_bad}. Intuitively, we can see the gap in generated causal content between them.
In this way, we can select some better samples which could be used to further finetune the model to achieve better causal alignment in this scenario.

% \subsection{Score alignment with human intuition}
% \label{app:score_alignment_human}
% \textcolor{red}{TODO}

\section{Discussion about LLM prompt enhancement technique}
\label{app:prompt_enhance}
Sora~\citep{sora_is_here} inherits a technique from Dall-E~\citep{betker2023improving} called prompt enhancement, where the model doesn't directly rely on the provided text prompt for generation. Instead, it first uses a pre-trained LLM to expand the prompt, adding missing elements such as environmental details and turning abstract concepts into more intuitive descriptions. Some models have already integrated this functionality into their latest VGM versions. 

We indeed observed that this technique slightly improved the model's ability to correctly understand causal rules. However, when scenarios became slightly more complex, either the LLM's expansion did not address the relevant parts, or even if the LLM did provide an expansion, the VGM still failed to generate reasonable results. We believe that, this technique is not the ultimate solution to creating a world simulator. On one hand, it supplements the VGM's shortcomings by leveraging the LLM's capabilities, but it doesn’t address the VGM’s core strengths. On the other hand, prompt enhancement cannot capture every detail because vision is much more complicated and informative than text, and once a scenario goes beyond the scope of the prompt, the VGM will struggle to respond appropriately.

To faithfully reflect the performance of the VGMs themselves, we disabled the prompt enhancement option for all closed-source models (where possible). Specifically, for Gen-3 and Hailuo, we turned off this feature. For Kling and Pika, however, we couldn’t find any official description on whether this technique was used.

\end{document}